\documentclass[twoside]{article}

\usepackage[accepted]{aistats2024}
\usepackage{eqnarray,amsmath}
\usepackage[utf8]{inputenc} 
\usepackage[T1]{fontenc}    
\usepackage{booktabs}       
\usepackage{amsfonts}       
\usepackage{nicefrac}       
\usepackage{microtype}      

\usepackage{caption}
\usepackage{subcaption}

\usepackage{epsf}
\usepackage{epsfig}
\usepackage{fancyhdr}
\usepackage{graphics}
\usepackage{graphicx}
\usepackage{psfrag}

\usepackage{url}
\usepackage[colorlinks,linkcolor=black,citecolor=blue, pagebackref=true]{hyperref}
\renewcommand*{\backrefalt}[4]{%
    \ifcase #1 \footnotesize{(Not cited.)}%
    \or        \footnotesize{(Cited on page~#2.)}%
    \else      \footnotesize{(Cited on pages~#2.)}%
    \fi}

\usepackage{color}

\usepackage{amsthm}
\usepackage{amsmath}
\usepackage{amssymb,bbm}
\usepackage{caption}
\usepackage{algorithmic}
\usepackage{algorithm}
\usepackage{textcomp}
\usepackage{siunitx}
\usepackage{wrapfig}
\usepackage{algorithmic}
\usepackage{algorithm}
\usepackage{multirow}
\usepackage{multicol}

\usepackage{comment}

\newtheorem{assumption}{Assumption}

\newtheorem{lemma}{Lemma}

\newtheorem{theorem}{Theorem}
\newtheorem{proposition}{Proposition}

\newcommand{\bbP}{\mathbb{P}}

\newcommand{\dint}{\mathrm{d}}

\newcommand{\blue}[1]{\textcolor{blue}{#1}}

\def\ie{i.e.,~}
\def\eg{e.g.,~}

\def\iid{i.i.d.~}
\def\wrt{w.r.t.~}
\def\resp{resp.~}

\newcommand{\nn}{\nonumber} 
\newcommand{\R}{\mathbb{R}} 

\newcommand{\Indi}{\mathbb{I}} 
\newcommand{\Ns}{\mathbb{N}} 

\newcommand{\cF}{\mathcal{F}}

\newcommand{\cS}{\mathcal{S}}

\newcommand{\cX}{\mathcal{X}}

\newcommand{\cY}{\mathcal{Y}}

\newcommand{\pin}{\pi_{i}^{n}}
\newcommand{\pizeroj}{\pi_{j}^{0}}
\newcommand{\dcijn}{\Delta c_{ij}^{n}}
\newcommand{\dgijn}{\Delta \Gamma_{ij}^{n}}
\newcommand{\daijn}{\Delta a_{ij}^{n}}
\newcommand{\dbijn}{\Delta b_{ij}^{n}}
\newcommand{\dvijn}{\Delta \nu_{ij}^{n}}
\newcommand{\norm}[1]{\|#1\|}

\newcommand{\cin}{c_i^n}
\newcommand{\gin}{\Gamma_i^n}
\newcommand{\ain}{a_i^n}
\newcommand{\bin}{b_i^n}
\newcommand{\vin}{\nu_i^n}

\newcommand{\cj}{c_j^0}
\newcommand{\gj}{\Gamma_j^0}
\newcommand{\aj}{a_j^0}
\newcommand{\bj}{b_j^0}
\newcommand{\vj}{\nu_j^0}

\newcommand{\cjp}{c_{j'}^0}
\newcommand{\gjp}{\Gamma_{j'}^0}
\newcommand{\ajp}{a_{j'}^0}
\newcommand{\bjp}{b_{j'}^0}
\newcommand{\vjp}{\nu_{j'}^0}

\newcommand{\zerod}{\mathbf{0}_d}
\newcommand{\ktilde}{\widetilde{k}}
\newcommand{\Dtilde}{\widetilde{D}}

\newcommand{\brj}{\bar{r}(|\mathcal{A}_j|)}
\newcommand{\trj}{\widetilde{r}(|\mathcal{A}_j|)}
\newcommand{\trjp}{\widetilde{r}(|\mathcal{A}_{j'}|)}

\newcommand{\brrj}{\bar{r}_j}
\newcommand{\trrj}{\widetilde{r}_j}

\newcommand{\brone}{\bar{r}(|\mathcal{A}_1|)}

\newcommand{\dcione}{\Delta c_{i1}^{n}}
\newcommand{\dgione}{\Delta \Gamma_{i1}^{n}}
\newcommand{\daione}{\Delta a_{i1}^{n}}
\newcommand{\dbione}{\Delta b_{i1}^{n}}
\newcommand{\dvione}{\Delta \nu_{i1}^{n}}

\newcommand{\trone}{\widetilde{r}(|\mathcal{A}_1|)}
\newcommand{\brjn}{\bar{r}(|\mathcal{A}^n_j|)}

\newcommand{\trs}{\widetilde{r}(|\mathcal{A}_{j^*}|)}

\DeclareMathOperator*{\argmax}{arg\,max}

\begin{document}

%

%

\twocolumn[

\aistatstitle{Towards Convergence Rates for Parameter Estimation in Gaussian-gated Mixture of Experts}

\aistatsauthor{Huy Nguyen$^{\diamond,\star}$ \And TrungTin Nguyen$^{\circ,\dagger,\star}$ \And  Khai Nguyen$^\diamond$ \And Nhat Ho$^\diamond$ }

\aistatsaddress{ Department of Statistics and Data Sciences, The University of Texas at Austin$^\diamond$ \\ School of Mathematics and Physics, The University of Queensland$^{\circ}$\\
Univ. Grenoble Alpes, Inria, CNRS, Grenoble INP, LJK, 38000 Grenoble, France$^{\dagger}$} ]

\begin{abstract}
  Originally introduced as a neural network for ensemble learning, mixture of experts (MoE) has recently become a fundamental building block of highly successful modern deep neural networks for heterogeneous data analysis in several applications of machine learning and statistics. Despite its popularity in practice, a satisfactory level of theoretical understanding of the MoE model is far from complete. To shed new light on this problem, we provide a convergence analysis for maximum likelihood estimation (MLE) in the Gaussian-gated MoE model. The main challenge of that analysis comes from the inclusion of covariates in the Gaussian gating functions and expert networks, which leads to their intrinsic interaction via some partial differential equations with respect to their parameters. We tackle these issues by designing novel Voronoi loss functions among parameters to accurately capture the heterogeneity of parameter estimation rates. Our findings reveal that the MLE has distinct behaviors under two complement settings of location parameters of the Gaussian gating functions, namely when all these parameters are non-zero versus when at least one among them vanishes. Notably, these behaviors can be characterized by the solvability of two different systems of polynomial equations. Finally, we conduct a simulation study to empirically verify our theoretical results.
\end{abstract}

\section{INTRODUCTION}
\label{sec:introduction}
Mixture of experts (MoE) \cite{jacobs_adaptive_1991,jordan_hierarchical_1994} is a popular statistical machine learning model where experts are either regression functions or classifiers, while the input-dependent weights (also called gating functions) softly partition the input space into different regions and define which regions each expert is responsible for (see  \cite{yuksel_twenty_2012,masoudnia_mixture_2014,fedus_review_2022} for further details).
In regression analysis with heterogeneous data, softmax-gated MoE \cite{jacobs_adaptive_1991,jordan_hierarchical_1994} and {\it Gaussian-gated MoE} (GMoE)~\cite{xu_alternative_1995} models are the most popular choices.
One of the main drawbacks of the softmax-gated MoE models is the difficulty of applying an expectation-maximization (EM) algorithm \cite{dempster_maximum_1977}, which requires an internal iterative numerical optimization procedure, \eg Newton-Raphson algorithm, to update the softmax parameters in the maximization step. On the other hand, parameters of the GMoE models can be updated analytically, which helps reduce the computational complexity of the estimation routine.
For those reasons, GMoE has become a fundamental component of modern deep neural networks in various fields, including speech recognition \cite{fritsch_adaptively_1996,you_speechmoe2_2022}, computer vision \cite{Lathuiliere2017,puigcerver_scalable_2021}, natural language processing \cite{shazeer_outrageously_2017,fedus_switch_2022,mustafa_multimodal_2022,do_hyperrouter_2023,pham2024competesmoe}, medical images \cite{han2024fusemoe}, robot dynamics \cite{{sato2000line,moody_fast_1989}}, remote sensing \cite{deleforge_high-dimensional_2015,kugler_fast_2022,forbes_mixture_2022,forbes_summary_2022}, and econometrics \cite{norets_adaptive_2021,norets_adaptive_2017,diani_multivariate_2022}.
However, there is a paucity of work aiming at theoretically understanding the density estimation and parameter estimation in the GMoE models, which has remained poorly understood in the literature to the best of our knowledge.

\noindent
\textbf{Related literature.} In the GMoE setting, early classical research focused on identifiability issues \cite{jiang_identifiability_1999} and parameter estimation in the exact-fitted setting, assuming the true number of components $k_0$ is known \cite{jiang_hierarchical_1999}. 
For most applications, it is a too strong presumption as the true number of components is seldom known. 
To deal with this problem, there are three common practical approaches. The first approach is based on model selection, most importantly the Bayesian information criterion from asymptotic theory \cite{forbes_summary_2022,chamroukhi_regularized_2019,khalili_new_2010} and the slope heuristic \cite{baudry_slope_2012,birge_minimal_2007} in a non-asymptotic framework \cite{nguyen_non_asymptotic_2022,nguyen_non_asymptotic_2023,nguyen_non_asymptotic_lasso_2023,nguyen_model_2022}. In particular, the bias term can be substantially reduced with a sufficiently large model collection \wrt the number of mixture components $k$ by well-studied universal approximations theorems \cite{nguyen_approximations_2021,mendes_convergence_2012,jiang_hierarchical_1999_AISTATS}.
However, since we have to search for the optimal $k$ over all possible values, this approach is computationally expensive.
%
The second approach is to design a tractable Bayesian nonparametric GMoE model. For example, \cite{nguyen_bayesian_2023} avoided any commitment to an arbitrary $k$ with posterior consistency guarantee thanks to the merge-truncate-merge post-processing in \cite{guha_posterior_2021}.
However, this approach still depends on a tuning parameter, which prevents the direct application of this approach to real data sets.
The last approach is to use prior knowledge to over-specify the true model, i.e. specifying more mixture components than necessary,
where most existing work is limited to its particular case, including mixture models \cite{ho_convergence_2016,ho_strong_2016,ho_singularity_2019,guha_posterior_2021,manole_refined_2022} and mixture of experts \cite{ho_convergence_2022,nguyen2023demystifying,nguyen2023general,nguyen2024statistical,nguyen2024temperature}. 
%
It is worth noting that the convergence behavior of parameter estimations in the GMoE model has remained an open question, which we aim to answer in this paper.
Before going into further details, we first formally introduce an affine instance of the GMoE model. This is a simplified but standard setting where we use linear functions for Gaussian mean experts.  

\noindent
\textbf{GMoE setting.}
%
GMoE models are used to capture the non-linear and heterogeneous relationship between the response $Y \in \cY \subset \R$ and the set of covariates $X \in \cX \subset   \R^d$, $d \in \Ns$.
%
In the affine GMoE model, the response $Y$ is approximated by  a $k_0$ local affine:
\begin{align}\label{eq_local_affine}
    Y = \sum_{j=1}^{k_0} \Indi \left(Z = j\right) [(a^0_j)^{\top}X + b^0_j + e^0_j].
\end{align}
Here $\Indi$ is an indicator function and $Z$ is a latent variable that captures a cluster relationship, such that $Z=j$ if $Y$ comes from cluster $j \in [k_0]:=\{1,2,\ldots,k_0\}$.
Vectors $a^0_j \in \R^d$ and scalars $b^0_j \in \R$ define cluster-specific affine transformations. In addition, $e^0_j$ are error terms that capture both the reconstruction error (due to the local affine approximations) and the observation noise in $\R$.
Let $\cF_d : = \left\{f(\cdot|\psi,\Sigma):\psi\in\mathbb{R}^d,\Sigma\in\mathcal{S}^+_{d}\right\}$ be the family of $d$-dimensional Gaussian density functions with mean $\psi$ and positive-definite covariance matrix $\Sigma$, where $\cS_d^{+}$ indicates the set of all symmetric positive-definite matrices on $\R^{d \times d}$.
Following the usual assumption that $e^0_j$ is a zero-mean Gaussian variable with variance $\nu^0_j \in \R_{+}$, it follows that
\begin{align*} 
	p\left(Y| X,Z = j\right) = f_{\mathcal{D}}\left(Y|(a^0_j)^{\top}X +b^0_j,\nu^0_j\right), \quad f_{\mathcal{D}}\in\mathcal{F}_{1}.
\end{align*}
To enforce the affine transformations to be local, $X$ is defined as a mixture of $k_0$ Gaussian components:
\begin{align}\label{eq_marginal_forward}
	p\left(X| Z = j\right) = f_{\mathcal{L}}\left(X|c^0_j,\Gamma^0_j\right),~
    p\left(Z=j\right) = \pi^0_j,
\end{align} where $f_{\mathcal{L}}\in\mathcal{F}_{d}$. Here, we refer to $f_{\mathcal{D}}$ and $f_{\mathcal{L}}$ as the data density and the local density, respectively. Additionally, $\pi^0_j > 0$ are called mixing proportions (or weights), satisfying $ \sum_{j=1}^{k_0} \pi^0_j  = 1$.
Via the law of total probability, we obtain the GMoE model of order $k_{0}$ whose joint density function $p_{G_0}(X,Y)$ is given by:
\begin{align}\label{eq_true_joint_GLLiM}
	%
    \sum_{j=1}^{k_0}\pi^0_jf_{\mathcal{L}}(X|c^0_j,\Gamma^0_j)\cdot f_{\mathcal{D}}(Y|(a^0_j)^{\top}X+b^0_j,\nu^0_j).
\end{align}
Here, $G_{0} := \sum_{j=1}^{k_0} \pi^0_j \delta_{(c^0_j,\Gamma^0_j,a^0_j,b^0_j,\nu^0_j)}$ denotes a true but unknown probability mixing measure, where $\delta$ is the Dirac measure and for $j \in[k_0]$, $(c^0_j,\Gamma^0_j,a^0_j,b^0_j,\nu^0_j)\in\Theta\subset \mathbb{R}^d\times\mathcal{S}^{+}_{d}\times\mathbb{R}^d\times\mathbb{R}\times\mathbb{R}_+$ are called components of $G_0$.
We assume that $\left\{\left(X_{i},Y_{i}\right)\right\}_{i \in [n]}$ are \iid~samples of random variable $\left(X,Y\right)$, coming from the GMoE model of order $k_{0}$.
To facilitate our theoretical guarantee, we assume that $\Theta$ is compact and $\cX$ is bounded.

\noindent
\textbf{Maximum likelihood estimation.} We propose a general theoretical framework for analyzing the statistical performance of {\it maximum likelihood estimation} (MLE) for parameters under the setting of the GMoE model. 
Since the true order $k_0$ is generally unknown in practice, it is necessary to over-specify the number of components of mixing measures to at most $k$, where $k> k_0$. In particular, we consider
\begin{align}
\label{eq_MLE_formulation}
\widehat{G}_n\in\argmax_{G\in\mathcal{O}_{k}(\Theta)}\sum_{i=1}^{n}\log(p_{G}(X_i, Y_i)),
\end{align}
where $\mathcal{O}_{k}(\Theta):=\{G=\sum_{i=1}^{k'}\pi_i\delta_{(c_i,\Gamma_i,a_i,b_i,\nu_i)}:1\leq k'\leq k, ~\sum_{i=1}^{k}\pi_i=1,~(c_i,\Gamma_i,a_i,b_i,\nu_i)\in\Theta\}$ denotes the set of all mixing measures with at most $k$ components.
\textbf{Theoretical challenges.} 
For the purpose of deriving parameter estimation rates in the GMoE model, we first use the Taylor expansion to decompose the term $p_{\widehat{G}_n}(X,Y)-p_{G_0}(X,Y)$ into a linear combination of elements which belong to a linearly independent set and associate with coefficients involving the discrepancies between parameter estimations and true parameters. By doing so, when the density estimation $p_{\widehat{G}_n}$ converges to the true density $p_{G_0}$, those parameter discrepancies also go to zero and we then obtain our desired parameter estimation rates. Nevertheless, the density decomposition is challenging due to a number of linearly dependent derivative terms in the Taylor expansion. In particular, we find out two interactions among the parameters of either function  $f_{\mathcal{D}}$ or  $f_{\mathcal{L}}$ via the following partial differential equations (PDEs):
\begin{align}
    \label{eq_interior_interaction}
    \dfrac{\partial^2 f_{\mathcal{D}}}{\partial b^2}=2\dfrac{\partial f_{\mathcal{D}}}{\partial \nu},\quad \dfrac{\partial^2f_{\mathcal{L}}}{\partial c~\partial c^{\top}}=2\dfrac{\partial f_{\mathcal{L}}}{\partial \Gamma}.
\end{align}
We refer to those interactions as \emph{interior interactions} since each of them involves either parameters $b,\nu$ of function $f_{\mathcal{D}}$ or parameters $c,\Gamma$ of function $f_{\mathcal{L}}$. 
Furthermore, we also figure out an interaction between the parameters of functions $f_{\mathcal{D}}$ and $f_{\mathcal{L}}$. 
More specifically, let us denote  $F(X,Y|\theta):=f_{\mathcal{L}}(X|c,\Gamma)f_{\mathcal{D}}(Y|a^{\top}X+b,\nu)$ where $\theta:=(c,\Gamma,a,b,\nu)$. Then, by taking the derivatives of $F$ with respect to its parameters as follows:
\begin{align*}
    \dfrac{\partial^2F}{\partial c~\partial b}(X,Y|\theta^0_j)&=\Gamma^{-1}(X-c^0_j)\cdot f_{\mathcal{L}}\cdot\frac{\partial f_{\mathcal{D}}}{\partial b};\\
    \dfrac{\partial F}{\partial a}(X,Y|\theta^0_j)&=X\cdot f_{\mathcal{L}}\cdot\frac{\partial f_{\mathcal{D}}}{\partial b},
\end{align*}
it can be seen that the following PDE holds true when the location parameter of $f_{\mathcal{L}}$ vanishes, i.e. $c^0_j=0$:
\begin{align}
    \label{eq:PDE}
    \dfrac{\partial^2F}{\partial c~\partial b}(X,Y|\theta^0_j)=\Gamma^{-1}\cdot\dfrac{\partial F}{\partial a}(X,Y|\theta^0_j).
\end{align}
We refer to the interaction among parameters $c,b,a$ in equation~\eqref{eq:PDE} as the \emph{exterior interaction}. Back to the density decomposition, it is necessary to aggregate linearly dependent derivative terms in equations~\eqref{eq_interior_interaction} and \eqref{eq:PDE} by taking the summation of their associated coefficients. As a result, we achieve our desired linear combination of linearly independent terms. However, the structure of associated coefficients in that combination becomes complex owing to the previous aggregation. Thus, when those coefficients converge to zero, we have to cope with two complex systems of polynomial equations given in equations~\eqref{eq:system_r_bar} and \eqref{eq:new_system}. 

\textbf{Overall contributions.} 
In this paper, we characterize the convergence behavior of maximum likelihood estimation in the GMoE model. Firstly, we demonstrate that the density estimation $p_{\widehat{G}_n}$ converges to the true density $p_{G_0}$ under the Total Variation distance $V$ at the parametric rate $V(p_{\widehat{G}_n},p_{G_0})=\mathcal{O}(n^{-1/2})$. Regarding the parameter estimation problem, given the above challenge discussion, we consider two complement settings of the location parameters $c^0_1,c^0_2,\ldots,c^0_{k_0}$ based on the validity of the PDE in equation~\eqref{eq:PDE} as follows (see also Table~\ref{table:parameter_rates}): 

\textbf{1. Type I setting:} \emph{all the values of $c^0_1,c^0_2,\ldots,c^0_{k_0}$ are different from zero}. Since the PDE~\eqref{eq:PDE} does not hold under this setting, we have to deal with only the interior interactions in equation~\eqref{eq_interior_interaction}. Thus, we propose a novel Voronoi loss function $\overline{D}(G,G_0)$ defined in equation~\eqref{eq_D_bar_formulation} to capture those interactions, and then establish the Total Variation lower bound $\overline{D}(\widehat{G}_n,G_0)\lesssim V(p_{\widehat{G}_n},p_{G_0})=\mathcal{O}(n^{-1/2})$. This result together with the formulation of $\overline{D}(\widehat{G}_n,G_0)$ indicate that exact-fitted parameters $\cj,\gj,\aj,\bj,\vj$, which are approximated by exactly one component, share the same estimation rate of order $\mathcal{O}(n^{-1/2})$. By contrast, the rates for estimating over-fitted parameters $\cj,\gj,\bj,\vj$, which are fitted by at least two components, depend on the solvability of the system of polynomial equations \eqref{eq:system_r_bar} and become no faster than $\mathcal{O}(n^{-1/4})$. These slow rates are due to the interior interactions among those parameters in equation~\eqref{eq_interior_interaction}.
As over-fitted parameters $\aj$ are not involved in those interactions, their estimation rates keep unchanged of order $\mathcal{O}(n^{-1/4})$.

\textbf{2. Type II setting:} \emph{at least one among the values of $c^0_1,c^0_2,\ldots,c^0_{k_0}$ is equal to zero}. Without loss of generality, we assume that $c^0_1,c^0_2,\ldots,c^0_{\ktilde}$ equal zero, where $1\leq \ktilde\leq k_0$, while other $c_j$'s are non-zero. Since the PDE~\eqref{eq:PDE} holds true under this setting, we have to confront both interior and exterior interactions among parameters. For that purpose, we construct another novel Voronoi loss function $\Dtilde(G,G_0)$ in equation~\eqref{eq_D_tilde_formulation} to handle those interactions, and then derive the Total Variation lower bound $\Dtilde(\widehat{G}_n,G_0)\lesssim V(p_{\widehat{G}_n},p_{G_0})=\mathcal{O}(n^{-1/2})$. Due to the occurrence of both interior and exterior interactions, the rates for estimating over-fitted parameters $\cj,\gj,\aj,\bj,\vj$ are now determined by the solvability of both systems of polynomial equations \eqref{eq:system_r_bar} and \eqref{eq:new_system}. Meanwhile, the estimation rates for their exact-fitted counterparts remain the same of order $\mathcal{O}(n^{-1/2})$.

\begin{table*}[!ht]
\centering
\begin{tabular}{ | m{3.5em} | m{6em}| c | c |c|c|c|c|} 
\hline
\multirow{2}{4em}{\textbf{Setting}} & \multirow{2}{4em}{\textbf{Exact-fitted} $c^0_j,\Gamma^0_j,a^0_j,b^0_j,\nu^0_j$} &  \multicolumn{2}{c|}{\textbf{Over-fitted} $a^0_j$} &  \multicolumn{2}{c|}{\textbf{Over-fitted} $c^0_j,b^0_j$} & \multicolumn{2}{c|}{\textbf{Over-fitted} $\Gamma^0_j,\nu^0_j$} \\ \cline{3-8}
& & $j\in[\ktilde]$ & $j\in[k_0]\setminus[\ktilde]$ & $j\in[\ktilde]$ & $j\in[k_0]\setminus[\ktilde]$ & $j\in[\ktilde]$ & $j\in[k_0]\setminus[\ktilde]$\\
\hline 
Type I & $\mathcal{O}(n^{-1/2})$ & \multicolumn{2}{c|}{$\mathcal{O}(n^{-1/4})$} & \multicolumn{2}{c|}{$\mathcal{O}(n^{-1/2\brrj})$} & \multicolumn{2}{c|}{$\mathcal{O}(n^{-1/\brrj})$}\\
\hline
Type II & $\mathcal{O}(n^{-1/2})$ & $\mathcal{O}(n^{-1/\trrj})$ & $\mathcal{O}(n^{-1/4})$ & $\mathcal{O}(n^{-1/2\trrj})$ & $\mathcal{O}(n^{-1/2\brrj})$ & $\mathcal{O}(n^{-1/\trrj})$ & $\mathcal{O}(n^{-1/\brrj})$\\
\hline
\end{tabular}
\caption{Summary of parameter estimation rates in the GMoE model under the Type I and Type II settings. 
Recall that the cardinality of Voronoi cells $\mathcal{A}_j$ (see Section~\ref{sec_preliminaries}) generated by true components $(c^0_j,\Gamma^0_j,a^0_j,b^0_j,\nu^0_j)$  indicates the number of components fitting them. When $|\mathcal{A}_j|=1$, we call them  exact-fitted parameters, but when $|\mathcal{A}_j|>1$, they are referred to as over-fitted parameters. Additionally, the notations $\brrj:=\brj$ and $\trrj:=\trj$ stand for the solvability of two polynomial equation systems \eqref{eq:system_r_bar} and \eqref{eq:new_system}, respectively. For example, if $|\mathcal{A}_j|=2$, then we have $\brrj=\trrj=4$. Meanwhile, we get $\brrj=\trrj=6$ if $|\mathcal{A}_j|=3$.
}
\label{table:parameter_rates}
\end{table*}

\textbf{Practical implication.} 
In practice, the parameters specific to each mixing component may carry useful information about the heterogeneity of the underlying (latent) subpopulations. 
Since in reality there is a tendency to ``over-fit'' the mixture generously by adding many more mixing components, our theory warns against this because, as we have shown, the convergence rate via standard methods such as MLE for subpopulation-specific parameters deteriorates rapidly with the number of redundant components.
Hopefully, the theoretical results will suggest practical ways to identify benign scenarios and impose helpful constraints when GMoE models have favourable convergence rates, and detect pathological scenarios that practitioners would do well to avoid.
In particular, practitioners can consistently estimate the true number of components based on our important threshold on the convergence rates of the MLE using the merge-truncate-merge procedure \cite{guha_posterior_2021} or Group-Sort-Fuse \cite{manole_estimating_2021}. 

\paragraph{Paper organization.} 
The rest of this paper proceeds as follows. In Section~\ref{sec_preliminaries}, we begin with providing some background on the identifiability of the GMoE model and the rate for estimating the joint density function under that model.
Next, in Section~\ref{sec_rate_estimating_parameters}, we establish the convergence rates of parameter estimation under both Type I and Type II settings, which are then empirically verified by simulation studies in Section~\ref{sec_experiments}.
%
%
Finally, we conclude the paper in Section~\ref{sec_conclusion} and defer proofs of all theoretical results to the supplementary material.

\paragraph{Notation.} Throughout the paper, 
$\{1, 2, \ldots, n\}$ is abbreviated as $[n]$ for any $n \in \Ns$.
Given any two sequences of positive real numbers $\{a_n\}_{n = 1}^{\infty}$ and $\{b_n\}_{n = 1}^{\infty}$, we write $a_n = \mathcal{O}(b_n)$ or $a_{n} \lesssim b_{n}$  to indicate that there exists a constant $C > 0$ such that $a_n \leq C b_n$ for all $ n\geq 1$. Next, for any vector $v\in\mathbb{R}^d$, we denote $|v|:=v_1+v_2+\ldots+v_d$, whereas $\|v\|_{p}$ stands for its $p$-norm with a note that $\|v\|$ implicitly indicates the $2$-norm unless stating otherwise. By abuse of notation, we also denote by $\|A\|$ the Frobenius norm of any matrix $A\in\mathbb{R}^{d\times d}$. Additionally, the notation $|S|$ represents for the cardinality of any set $S$.
Finally, given two probability density functions $p,q$ with respect to the Lebesgue measure $\mu$, we define $V(p,q) := \frac 1 2 \int |p-q| d\mu$ as their Total Variation distance, while $h^2(p,q) := \frac 1 2 \int (\sqrt p - \sqrt q)^2 d\mu$ denotes the squared Hellinger distance between them.

\section{PRELIMINARIES}
\label{sec_preliminaries}
In this section, we first verify the identifiability of the GMoE model, and then establish the density estimation rate under that model. Lastly, we introduce a notion of a Voronoi cells, which will be used to build Voronoi loss functions in Section~\ref{sec_rate_estimating_parameters}. 

Firstly, we demonstrate in the following proposition that the GMoE model is identifiable:
\begin{proposition}[Identifiability of the GMoE model]
    \label{prop_identifiability}
    Let $G$ and $G'$ be two mixing measures in $\mathcal{O}_{k}(\Theta)$. If the equation $p_{G}(X,Y)=p_{G'}(X,Y)$ holds true for almost surely $(X,Y)\in\mathcal{X}\times\mathcal{Y}$, then we obtain that $G\equiv G'$.
\end{proposition}
The proof of Proposition~\ref{prop_identifiability} is deferred to Appendix~\ref{appendix_identifiability}. Given the above result, we know that two mixing measures $G$ and $G'$ are equivalent if and only if they share the same joint density function.

Next, we characterize the convergence rate of the joint density estimation $p_{\widehat{G}_n}$ to its true counterpart $p_{G_0}$ under the Total Variation distance.

\begin{proposition}[Joint Density Estimation Rate]
    \label{prop_density_rate}
    With the MLE $\widehat{G}_n$ defined in equation~\eqref{eq_MLE_formulation}, the following bound indicates that the density estimation $p_{\widehat{G}_n}$ converges to the true density $p_{G_0}$ under the Total Variation distance at the parametric rate of order $\mathcal{O}(n^{-1/2})$ (up to a logarithmic term):
    \begin{align*}
        \mathbb{P}(V(p_{\widehat{G}_n},p_{G_0})>C_1\sqrt{\log(n)/n})\lesssim n^{-C_2},
    \end{align*}
    where $C_1$ and $C_2$ are universal constants.
\end{proposition}
The proof of Proposition~\ref{prop_density_rate} can be found in Appendix~\ref{appendix_density_rate}. This result is a key ingredient to study the parameter estimation problem in the GMoE model in subsequent sections. In particular, if we are able to establish the lower bound of the Total Variation distance in terms of some loss function $D$ between two mixing measures, i.e., $V(p_{G},p_{G_0})\gtrsim D(G,G_0)$ for any mixing measure $G\in\mathcal{O}_{k}(\Theta)$, then the MLE $\widehat{G}_n$ also converges to the true mixing measure $G_0$ at the parametric rate of $\mathcal{O}(n^{-1/2})$. Based on this result, we then achieve the parameter estimation rates through the formulation of the loss function $D(\widehat{G}_n,G_0)$. For that purpose, let us introduce a notion of Voronoi cells which are essential to construct Voronoi loss functions later in  Section~\ref{sec_rate_estimating_parameters}.

\textbf{Voronoi cells}. In general, true parameters which are fitted by exactly one component should enjoy faster estimation rates than those approximated by more than one component. Therefore, in order to capture the convergence behavior of parameter estimations accurately, we define $k_0$ different index sets called Voronoi cells to control the number of fitted components approaching each of the $k_0$ true components.
More formally, for any $G\in\mathcal{O}_{k}(\Theta)$, the Voronoi cell $\mathcal{A}_j:=\mathcal{A}_j(G)$ generated by $\theta^0_j:=(\cj,\gj,\aj,\bj,\vj)$ is defined as 
\begin{align}\label{eq_Voronoi_cell}
    \mathcal{A}_j
    &:=\{i\in[k]:\|\theta_i-\theta^0_j\|\leq\|\theta_i-\theta^0_{\ell}\|,\ \forall \ell\neq j\},
\end{align}
for any $j\in[k_0]$, where $\theta_i:=(c_i,\Gamma_i,a_i,b_i,\nu_i)$. An illustration of Voronoi cells is given in Appendix~\ref{appendix_Voronoi}. Notably, the cardinality of each Voronoi cell $\mathcal{A}_j$ is exactly the number of fitted components approximating the true component $\theta^0_j$.


\section{PARAMETER ESTIMATION RATES} \label{sec_rate_estimating_parameters}
In this section, we conduct the convergence analysis for parameter estimation in the GMoE model under the Type I and Type II settings in Section~\ref{sec_type_I} and Section~\ref{sec_type_II}, respectively. Then, we sketch the proof for main results in both settings in Section~\ref{sec_proof_sketch}. 

\subsection{Type I Setting}\label{sec_type_I}

Let us recall that under this setting, all the values of $c^0_1,c^0_2,\ldots,c^0_{k_0}$ are non-zero. Although the exterior interaction between the parameters of two functions $f_{\mathcal{L}}$ and $f_{\mathcal{D}}$ mentioned in equation~\eqref{eq:PDE} does not hold in this scenario, we encounter two interior interactions among parameters $b,\nu$ and $c,\Gamma$ via the following PDEs: 
\begin{align}
    \label{eq:interior_interaction}
    \dfrac{\partial^2 f_{\mathcal{D}}}{\partial b^2}=2\dfrac{\partial f_{\mathcal{D}}}{\partial \nu},\quad \dfrac{\partial^2f_{\mathcal{L}}}{\partial c~\partial c^{\top}}=2\dfrac{\partial f_{\mathcal{L}}}{\partial \Gamma}.
\end{align}
\textbf{System of polynomial equations.} To precisely characterize the estimation rates for those parameters, we need to consider the solvability of a system of polynomial equations which was previously studied in \cite{ho_convergence_2016}. In particular, for each $m\geq 2$, let $\Bar{r}(m)$ be the smallest positive integer $r$ such that the system:
\begin{align}
\label{eq:system_r_bar}
\sum_{l=1}^{m}\sum_{\substack{n_1,n_2\in\mathbb{N}:\\n_1+2n_2=s}}\dfrac{p^2_{l}~q^{n_1}_{1l}~q^{n_2}_{2l}}{n_1!~n_2!}=0, \quad s=1,2,\ldots,r,
\end{align}
does not have any non-trivial solutions for the unknown variables $\{p_{l},q_{1l},q_{2l}\}_{l=1}^{m}$. Here, a solution is called non-trivial if all the values of $p_{l}$ are different from zero, whereas at least one among $q_{1l}$ is non-zero. The following lemma gives us the values of $\Bar{r}(m)$ at some specific points $m$.
\begin{lemma}[Proposition 2.1, \cite{ho_convergence_2016}]
    \label{lemma_r_bar}
    When $m=2$, we have that $\Bar{r}(m)=4$, while for $m=3$, we get $\Bar{r}(m)=6$. If $m\geq 4$, then $\Bar{r}(m)\geq 7$.
\end{lemma}
Proof of Lemma~\ref{lemma_r_bar} is in \cite{ho_convergence_2016}. Now, we are ready to introduce a Voronoi loss function used for this setting. 

\textbf{Voronoi loss function.} For simplicity, we denote $\Delta c_{ij}:=c_i-\cj$, $\Delta \Gamma_{ij}:=\Gamma_i-\gj$, $\Delta a_{ij}:=a_i-\aj$, $\Delta b_{ij}:=b_i-\bj$, $\Delta \nu_{ij}:=\nu_i-\vj$ and $\Bar{r}_j:=\brj$. Additionally, we also define mappings $K_{ij}:\mathbb{N}^5\to\mathbb{R}$ such that $K_{ij}(\kappa_1,\kappa_2,\kappa_3,\kappa_4,\kappa_5):=\norm{\Delta c_{ij}}^{\kappa_1}+\norm{\Delta\Gamma_{ij}}^{\kappa_2}+\norm{\Delta a_{ij}}^{\kappa_3}+|\Delta b_{ij}|^{\kappa_4}+|\Delta \nu_{ij}|^{\kappa_5}$,
for any $j\in[k_0]$ and $i\in\mathcal{A}_j$. Then, the Voronoi loss function $\overline{D}(G,G_0)$ of interest in this setting is given by:
\begin{align}
    &\overline{D}(G,G_0):=\sum_{\substack{j:|\mathcal{A}_j|>1,\\ i\in\mathcal{A}_j }}\pi_iK_{ij}\Big(\bar{r}_j,\frac{\bar{r}_j}{2},2,\bar{r}_j,\frac{\bar{r}_j}{2}\Big)~+\nonumber\\
    \label{eq_D_bar_formulation}
    &\sum_{\substack{j:|\mathcal{A}_j|=1,\\ i\in\mathcal{A}_j }}\pi_i{K_{ij}(1,1,1,1,1)} + \sum_{j=1}^{k_0}\left|\sum_{i\in\mathcal{A}_j}\pi_i-\pizeroj\right|.
\end{align}
Given this loss function, we capture parameter estimation rates in the GMoE model in the following theorem.
\begin{theorem}
\label{theorem:type_1_setting}
Under the Type I setting, the Total Variation lower bound $V(p_{G},p_{G_0})\gtrsim \overline{D}(G,G_0)$ holds for any $G\in\mathcal{O}_{k}(\Theta)$, which implies that there exists a universal constant $C_3>0$ depending on $G_0$ and $\Theta$ satisfying
\begin{align*}
    \bbP(\overline{D}(\widehat{G}_n,G_0)>C_3\sqrt{\log (n)/n})\lesssim n^{-C_4},
\end{align*}
where $C_4>0$ is a constant that depends only on $\Theta$.
\end{theorem}
Proof of Theorem~\ref{theorem:type_1_setting} is in Appendix~\ref{appendix:type_1_setting}. It follows from Theorem~\ref{theorem:type_1_setting} that the discrepancy $\overline{D}(\widehat{G}_n,G_0)$ vanishes at a rate of order $\mathcal{O}(n^{-1/2})$ up to a logarithmic constant, which leads to following observations: \textbf{(i)} True parameters $c^0_j,\Gamma^0_j,a^0_j,b^0_j,\nu^0_j$, which are fitted by exactly one component, share the same estimation rate of order $\mathcal{O}(n^{-1/2})$; \textbf{(ii)} By contrast, the rates for estimating parameters fitted by more than one element are significantly slower. In particular, the estimation rates for $\cj,\bj$ are of order $\mathcal{O}(n^{-1/2\brjn})$, whereas those for $\gj,\vj$ are of order $\mathcal{O}(n^{-1/\brjn})$ in which $\mathcal{A}^n_j:=\mathcal{A}_j(\widehat{G}_n)$. For instance, if we have $|\mathcal{A}^n_j|=3$, then Lemma~\ref{lemma_r_bar} indicates that the previous two rates become $\mathcal{O}(n^{-1/12})$ and $\mathcal{O}(n^{-1/6})$, respectively. These slow rates are owing to the interior interactions among those parameters in equation~\eqref{eq:interior_interaction}. Meanwhile, $\aj$ admits a much faster rate of order $\mathcal{O}(n^{-1/4})$ as it does not interact with other parameters.

\subsection{Type II Setting}\label{sec_type_II}
Next, we consider the Type II setting, namely when at least one among $c^0_1,c^0_2,\ldots,c^0_{k_0}$ is equal to vector $\zerod$. Without loss of generality, we assume that $c^0_1=c^0_2=\ldots=c^0_{\ktilde}=\zerod$, while $c^0_{\ktilde+1},c^0_{\ktilde+2},\ldots,c^0_{k_0}$ are different from $\zerod$. Under this setting, we encounter not only the two interior interactions in equation~\eqref{eq:interior_interaction} but also the exterior interaction expressed by the following PDE:
\begin{align}
    \label{eq:exterior_interaction}
    \dfrac{\partial^2F}{\partial c~\partial b}(X,Y|\theta^0_j)=\Gamma^{-1}\cdot\dfrac{\partial F}{\partial a}(X,Y|\theta^0_j),
\end{align}
where $F(X,Y|\theta):=f_{\mathcal{L}}(X|c,\Gamma)f_{\mathcal{D}}(Y|a^{\top}X+b,\nu)$ and $\theta:=(c,\Gamma,a,b,\nu)$. This phenomenon poses a lot of challenges in the parameter estimation problem. Therefore, we will only present the results when $d=1$ for simplicity, while those for the setting $d>1$ can be argued in a similar fashion but with more complex notations. 

\textbf{System of polynomial equations.} Due to the emergence of the exterior interaction, we need to control the solvability of a totally new system of polynomial equations, which is given by  
\begin{align}
\label{eq:new_system}
\sum_{l=1}^{m}\sum_{\alpha\in\mathcal{J}_{\ell_1,\ell_2}}\dfrac{p_l^2~q_{1l}^{\alpha_1}~q_{2l}^{\alpha_2}~q_{3l}^{\alpha_3}~q_{4l}^{\alpha_4}~q_{5l}^{\alpha_5}}{\alpha_1!~\alpha_2!~\alpha_3!~\alpha_4!~\alpha_5!}=0,
\end{align}
for all $\ell_1,\ell_2\geq 0$ satisfying $1\leq \ell_1+\ell_2\leq r$, where $\mathcal{J}_{\ell_1,\ell_2}:=\{\alpha=(\alpha_l)_{l=1}^5\in\mathbb{N}^5:\alpha_1+2\alpha_2+\alpha_3=\ell_1,\ \alpha_3+\alpha_4+2\alpha_5=\ell_2\}$. Now, we define $\widetilde{r}(m)$ as the smallest natural number $r$ such that the system in equation~\eqref{eq:new_system} does not have any non-trivial solutions for the unknown variables $\{p_{l},q_{1l},q_{2l},q_{3l},q_{4l},q_{5l}\}_{l=1}^m$, namely, all of $p_{l}$ are non-zero, whereas at least one among $q_{4l}$ is different from zero. The following lemma establishes a connection between $\widetilde{r}(m)$ and $\Bar{r}(m)$ as well as provides the values of $\widetilde{r}(m)$ given some specific choices of $m$.
\begin{lemma}
    \label{lemma_r_tilde}
    In general, we have $\widetilde{r}(m)\leq \Bar{r}(m)$ for all $m\in\mathbb{N}$. Furthermore, the equality occurs when $m=2$ and $m=3$, meaning that $\widetilde{r}(2)=4$ and $\widetilde{r}(3)=6$. 
\end{lemma}
Proof of Lemma~\ref{lemma_r_tilde} is in Appendix~\ref{appendix_r_tilde}. Next, we introduce a Voronoi loss function tailored to this setting. 

\textbf{Voronoi loss function.} Firstly, let us reformulate the mappings $K_{ij}$ defined in Section~\ref{sec_type_I} for $d=1$ as $K_{ij}(\kappa_1,\kappa_2,\kappa_3,\kappa_4,\kappa_5):=|\Delta c_{ij}|^{\kappa_1}+|\Delta\Gamma_{ij}|^{\kappa_2}+|\Delta a_{ij}|^{\kappa_3}+|\Delta b_{ij}|^{\kappa_4}+|\Delta \nu_{ij}|^{\kappa_5}$. In addition, we denote $\trrj:=\trj$ and $\bar{r}_j:=\brj$, for any $j\in[k_0]$. Then, the Voronoi loss of interest $\Dtilde(G,G_0)$ is defined as follows:
\begin{align}
\label{eq_D_tilde_formulation}
&\Dtilde(G,G_0):=\sum_{\substack{j\in[\ktilde]:|\mathcal{A}_j|>1,\\ i\in\mathcal{A}_j}}\pi_iK_{ij}\Big(\trrj,\frac{\trrj}{2},\frac{\trrj}{2},\trrj,\frac{\trrj}{2}\Big)\nonumber\\
&+\sum_{\substack{j\in[k_0]\setminus[\ktilde]:|\mathcal{A}_j|>1,\\ i\in\mathcal{A}_j}}\pi_iK_{ij}\Big(\bar{r}_j,\frac{\bar{r}_j}{2},2,\bar{r}_j,\frac{\bar{r}_j}{2}\Big)\nonumber\\
&+\sum_{\substack{j\in[k_0]:|\mathcal{A}_j|=1,\\ i\in\mathcal{A}_j}}\pi_iK_{ij}(1,1,1,1,1)+\sum_{j=1}^{k_0}\left|\sum_{i\in\mathcal{A}_j}\pi_i-\pizeroj\right|.
\end{align}
Given the above loss function, we derive the rates for estimating parameters under the Type II setting in the following theorem.
\begin{theorem}
\label{theorem:type_2_setting}
Under the Type II setting, the Total Variation lower bound $V(p_{G},p_{G_0})\gtrsim \Dtilde(G,G_0)$ holds for any $G\in\mathcal{O}_{k}(\Theta)$, which indicates that we can find a constant $C_5>0$ depending on $G_0,\Theta$ such that
\begin{align*}
    \bbP(\Dtilde(\widehat{G}_n,G_0)>C_5\sqrt{\log (n)/n})\lesssim n^{-C_6},
\end{align*}
where $C_6>0$ is a constant that depends only on $\Theta$.
\end{theorem}
Proof of Theorem~\ref{theorem:type_2_setting} is in Appendix~\ref{appendix:type_2_setting}. Similar to Theorem~\ref{theorem:type_1_setting}, the Voronoi loss $\widetilde{D}(\widehat{G}_n,G_0)$ also converges to zero at a rate of order $\mathcal{O}(n^{-1/2})$ (up to a logarithmic term) under the Type II setting. Moreover, true parameters $c^0_j,\Gamma^0_j,a^0_j,b^0_j,\nu^0_j$ enjoy the same estimation rates as their counterparts in Section~\ref{sec_type_I} for any $j\in[k_0]:|\mathcal{A}^n_j|=1$ and $j\in[k_0]\setminus[\ktilde]:|\mathcal{A}^n_j|>1$. However, the difference in the convergence behavior occurs when $j\in[\ktilde]:|\mathcal{A}^n_j|>1$. In particular, the rates for estimating parameters $\aj$ now drop substantially to $\mathcal{O}(n^{-1/\trj})$ in comparison with $\mathcal{O}(n^{-1/4})$ under the Type I settings. This phenomenon happens due to the interaction of $\aj$ with parameters $\cj,\bj$ via the PDE in equation~\eqref{eq:exterior_interaction}.
\subsection{Proof Sketch}\label{sec_proof_sketch}
Since arguments used for the proof of Theorem~\ref{theorem:type_1_setting} are included in that of Theorem~\ref{theorem:type_2_setting}, we will present the former proof sketch implicitly inside the latter. In particular, we focus on establishing the bound $\inf_{G\in\mathcal{O}_{k}(\Theta)}V(p_{G},p_{G_0})/\Dtilde(G,G_0)>0$ under the Type II setting when $d=1$. For that purpose, we will respectively demonstrate its local and global versions by contradiction as follows:

\textbf{Local bound:} We wish to prove that $$\lim_{\varepsilon>0}\inf_{G\in\mathcal{O}_{k}(\Theta),\Dtilde(G,G_0)\leq\varepsilon}V(p_{G},p_{G_0})/\Dtilde(G,G_0)>0.$$ Assume that this bound does not hold, then we can find a sequence $G_n=\sum_{i=1}^{k_n}\pin\delta_{\theta^n_i}\in\mathcal{O}_{k}(\Theta)$, where $\theta^n_i:=(\cin,\gin,\ain,\bin,\vin)$, such that $V(p_{G_n},p_{G_0})/\Dtilde(G_n,G_0)$ and $\Dtilde(G_n,G_0)$ both vanish as $n\to\infty$. Now, we decompose $\Xi_n:=p_{G_n}(X,Y)-p_{G_0}(X,Y)$ as
\begin{align*}
    \Xi_n&=\sum_{j=1}^{k_0}\sum_{i\in\mathcal{A}_j}\pin[F(X,Y|\theta^n_i)-F(X,Y|\theta^0_j)] \nn \\&+\sum_{j=1}^{k_0}\left(\sum_{i\in\mathcal{A}_j}\pin-\pizeroj\right)F(X,Y|\theta^0_j),
\end{align*}
where $\theta^0_j:=(\cj,\gj,\aj,\bj,\vj)$. Let us denote $h_1(X,a,b)=a^{\top}X+b$ for any $a\in\mathbb{R}^d,b\in\mathbb{R}$. Then, for $i\in\mathcal{A}_{j}$ and $i'\in\mathcal{A}_{j'}$ where $j\in[\ktilde]$ and $j'\in[k_0]\setminus[\ktilde]$, we invoke the Taylor expansion up to some orders $r_{1j}$ and $r_{2j'}$ (we will choose later) for $F(X,Y|\theta^n_{i})$ and $F(X,Y|\theta^n_{i'})$, respectively, as follows:
\begin{align*}
    &F(X,Y|\theta^n_{i}) - F(X,Y|\theta^0_{j})\\
    &=\sum_{\ell_1+\ell_2=1}^{2r_{1j}}Q^{n}_{\ell_1,\ell_2}(j)\cdot X^{\ell_1}f_{\mathcal{L}}(X|\cj,\gj)\nn\\
    &\quad \times \frac{\partial^{\ell_2}f_{\mathcal{D}}}{\partial h_1^{\ell_2}}(Y|\aj X+\bj,\vj)+R_{1ij}(X,Y),\\
    &F(X,Y|\theta^n_{i'}) - F(X,Y|\theta^0_{j'})\\&=R_{2i'j'}(X,Y)
    +\sum_{\alpha_3=0}^{r_{2j'}}\sum_{\tau_1+\tau_2=0}^{2(r_{2j'}-\alpha_3)}T^{n}_{\alpha_3,\tau_1,\tau_2}(j')\cdot X^{\alpha_3} \nn\\
    &\quad \times \frac{\partial^{\tau_1}f_{\mathcal{L}}}{\partial c^{\tau_1}}(X|c^0_{j'},\Gamma^0_{j'})\frac{\partial^{\alpha_3+\tau_2}f_{\mathcal{D}}}{\partial h_1^{\alpha_3+\tau_2}}(Y|a^0_{j'}X+b^0_{j'},\nu^0_{j'}).
\end{align*}
Here $R_{1ij}(X,Y)$ and $R_{2i'j'}(X,Y)$ are Taylor remainders such that their ratios to $\Dtilde(G_n,G_0)$ vanishes as $n\to\infty$. Thus, we can treat $\Xi_n/\Dtilde(G_n,G_0)$ as a linear combination of linearly independent terms $$X^{\ell_1}\cdot f_{\mathcal{L}}(X|\cj,\gj)\cdot\frac{\partial^{\ell_2}f_{\mathcal{D}}}{\partial h_1^{\ell_2}}(Y|\aj X+\bj,\vj),$$ $$X^{\alpha_3}\cdot\frac{\partial^{\tau_1}f_{\mathcal{L}}}{\partial c^{\tau_1}}(X|c^0_{j'},\Gamma^0_{j'})\cdot\frac{\partial^{\alpha_3+\tau_2}f_{\mathcal{D}}}{\partial h_1^{\alpha_3+\tau_2}}(Y|a^0_{j'}X+b^0_{j'},\nu^0_{j'})$$ associated with coefficients $Q^{n}_{\ell_1,\ell_2}(j)$ and $T^{n}_{\alpha_3,\tau_1,\tau_2}(j')$, respectively.
Moreover, it follows from Fatou's lemma that $\Xi_n/\Dtilde(G_n,G_0)$ approaches zero when $n\to\infty$. Consequently, all the coefficients in the representation of $\Xi_n/\Dtilde(G_n,G_0)$, i.e. $Q^{n}_{\ell_1,\ell_2}(j)/\Dtilde(G_n,G_0)$ and $T^{n}_{\alpha_3,\tau_1,\tau_2}(j')/\Dtilde(G_n,G_0)$, go to zero as $n\to\infty$. Therefore, in order to point out a contradiction, we need to choose the values of $r_{1j}$ and $r_{2j'}$ such that at least one among these coefficients does not vanish. As a result, we achieve the aforementioned local bound. Now, we will show how to determine such values of $r_{1j}$ and $r_{2j'}$. It is worth noting that if we set $\ktilde=0$, then Type II settings reduces to Type I settings and we only need to deal with $r_{2j'}$ as follows:

\textbf{Type I setting:} We will specify an appropriate of $r_{2j'}$ during proving by contradiction that not all the coefficients $T^{n}_{\alpha_3,\tau_1,\tau_2}(j')/\Dtilde(G_n,G_0)$ tend to zero. Assume that these coefficients all vanish, then we extract some useful limits among them for our arguments and end up with the following system of polynomial equations:
\begin{align*}
    \sum_{i'\in\mathcal{A}_{j'}}\sum_{n_1+2n_2=s}\frac{p^2_{l}~q^{n_1}_{1l}~q^{n_2}_{2l}}{n_1!~n_2!}=0, \quad s=1,2,\ldots,r_{2j'}.
\end{align*}
By construction, this system must have at least one non-trivial solution. Thus, to contradict this condition, we set $r_{2j'}=\Bar{r}(|\mathcal{A}_{j'}|)$, which makes the above system has no non-trivial solutions.

\textbf{Type II setting:} When $\ktilde>0$, i.e. there exist some zero-valued parameter $c_j$, we will keep $r_{2j'}=\Bar{r}(|\mathcal{A}_{j'}|)$ for all $j'\in[k_0]\setminus[\ktilde]$ and find the desired values of $r_{1j}$ for $j\in[\ktilde]$ by showing by contradiction that not all the coefficients $Q^{n}_{\ell_1,\ell_2}(j)/\Dtilde(G_n,G_0)$ go to zero. En route to pointing out a contradiction to the hypothesis, we come across a more complex system of polynomial equations than its counterpart in the previous setting, specifically
\begin{align*}
    \sum_{i\in\mathcal{A}_j}\sum_{\alpha\in\mathcal{J}_{\ell_1,\ell_2}}\dfrac{p_i^2~q_{1i}^{\alpha_1}~q_{2i}^{\alpha_2}~q_{3i}^{\alpha_3}~q_{4i}^{\alpha_4}~q_{5i}^{\alpha_5}}{\alpha_1!~\alpha_2!~\alpha_3!~\alpha_4!~\alpha_5!}=0,
\end{align*}
for all $\ell_1,\ell_2\geq 0$ such that $1\leq \ell_1+\ell_2\leq r_{1j}$, where $\mathcal{J}_{\ell_1,\ell_2}:=\{\alpha=(\alpha_i)_{i=1}^5\in\mathbb{N}^5:\alpha_1+2\alpha_2+\alpha_3=\ell_1,\ \alpha_3+\alpha_4+2\alpha_5=\ell_2\}$. Since this system necessarily has a non-trivial solution, we choose $r_{1j}=\trj$ so that it admits only trivial solutions, which contradicts the previous claim. Consequently, we can find a constant $\varepsilon'>0$ such that $$\inf_{G\in\mathcal{O}_{k}(\Theta),\Dtilde(G,G_0)\leq\varepsilon'}V(p_{G},p_{G_0})/\Dtilde(G,G_0)>0.$$

\textbf{Global bound:} Thus, to complete the proof, it is sufficient to demonstrate the global bound $$\inf_{G\in\mathcal{O}_{k}(\Theta),\Dtilde(G,G_0)>\varepsilon'}V(p_{G},p_{G_0})/\Dtilde(G,G_0)>0.$$ If this bound did not hold, there would be a mixing measure $G'\in\mathcal{O}_{k}(\Theta)$ that satisfies $p_{G'}(X,Y)=p_{G_0}(X,Y)$ for almost surely $(X,Y)$, which leads to $G'\equiv G_0$ by Proposition~\ref{prop_identifiability}. As a result, we obtain $\Dtilde(G',G_0)=0$, which contradicts the constraint that $\Dtilde(G',G_0)>\varepsilon'$. Hence, the proof sketch is completed.
\begin{figure*}[!ht]
    \centering
    \begin{subfigure}{.45\textwidth}
        \centering
        \includegraphics[scale = .48]{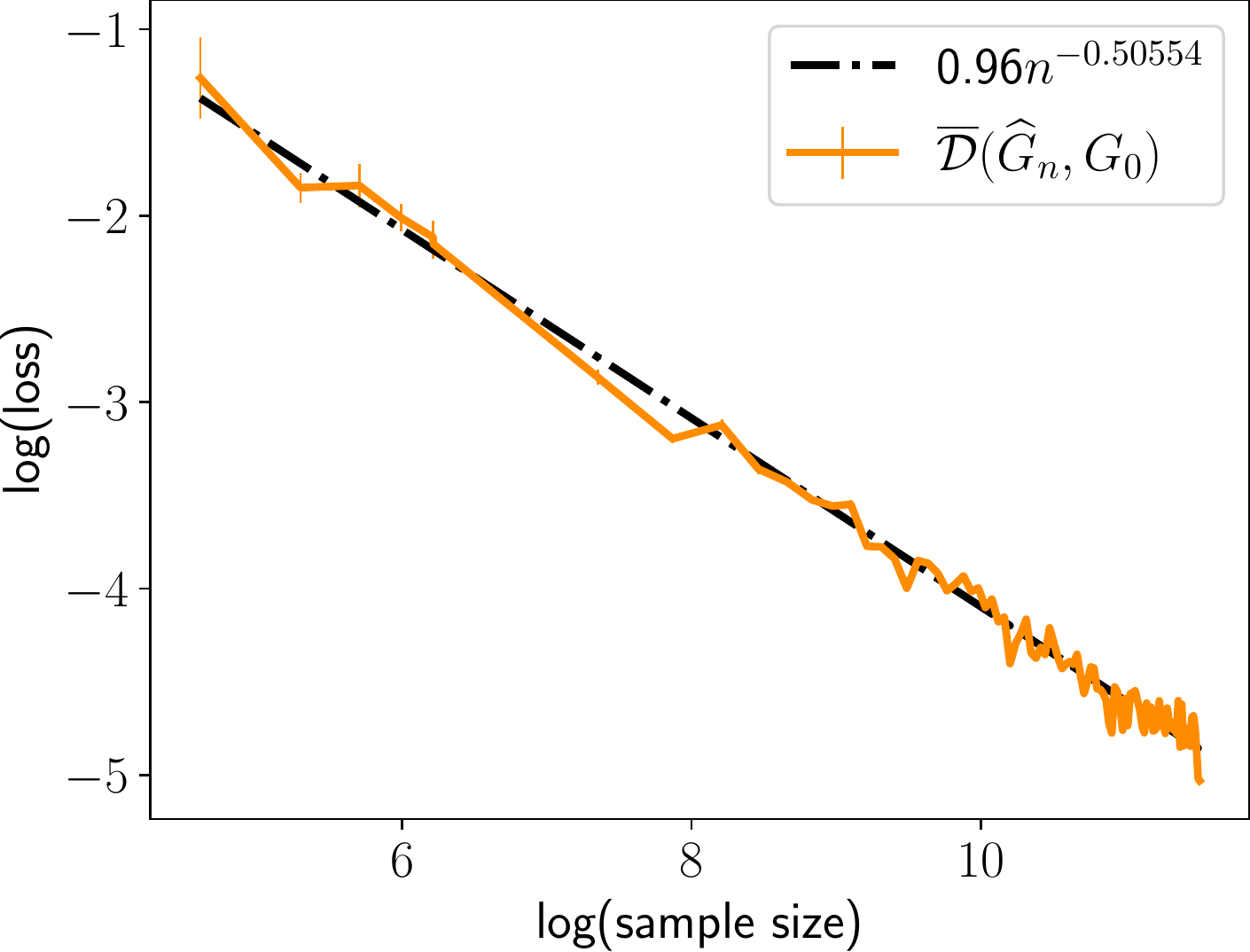}
        \caption{Model I, $k = 4$}	
        \label{subfig_model_type_I_k4}
    \end{subfigure}
    \hspace{0.6cm}
    \begin{subfigure}{.45\textwidth}
        \centering
        \includegraphics[scale = .48]{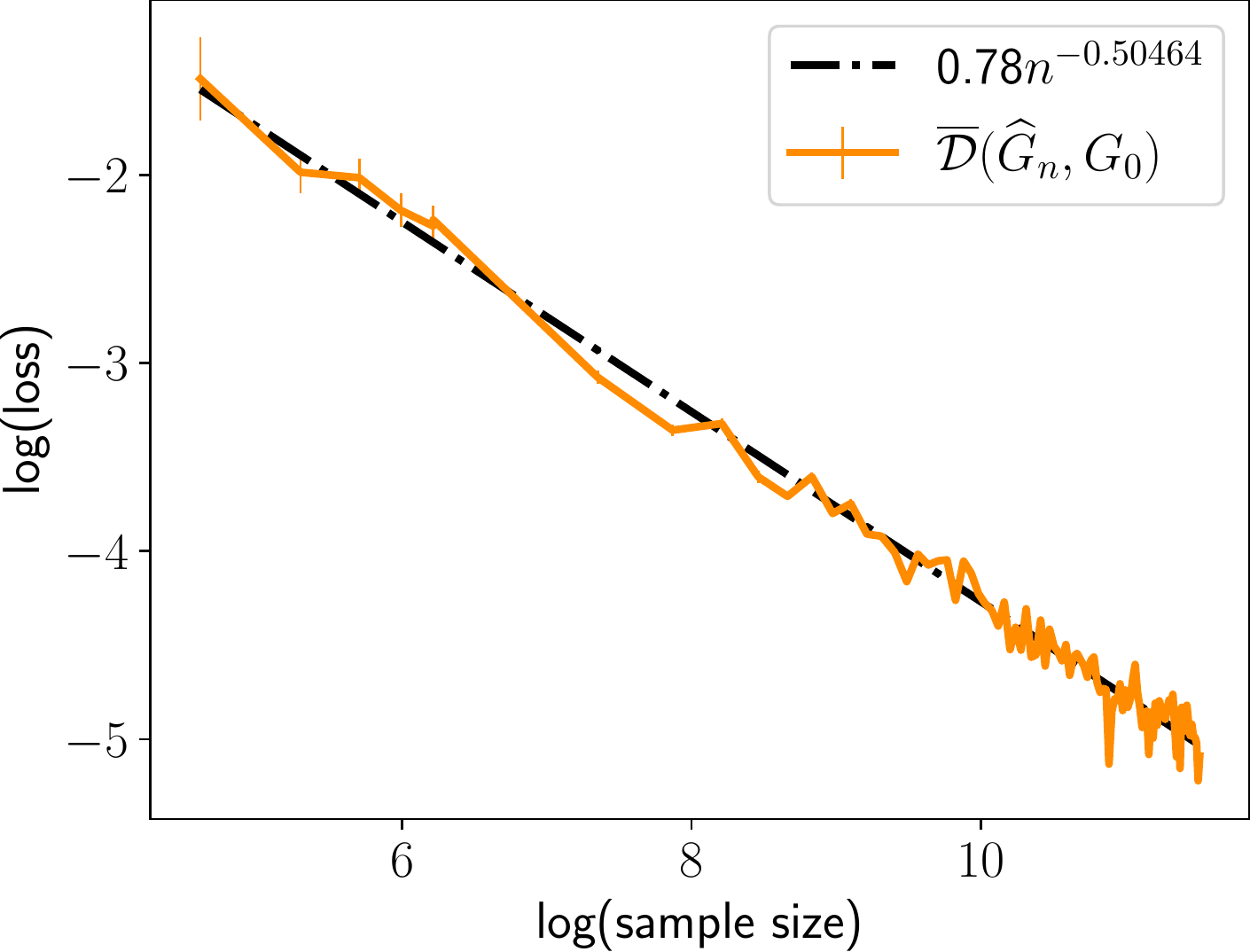}
        \caption{Model I, $k = 5$}	
        \label{subfig_model_type_I_k5}
    \end{subfigure}%
    \vspace{0.1cm}
    \begin{subfigure}{.45\textwidth}
        \centering
        \includegraphics[scale = .48]{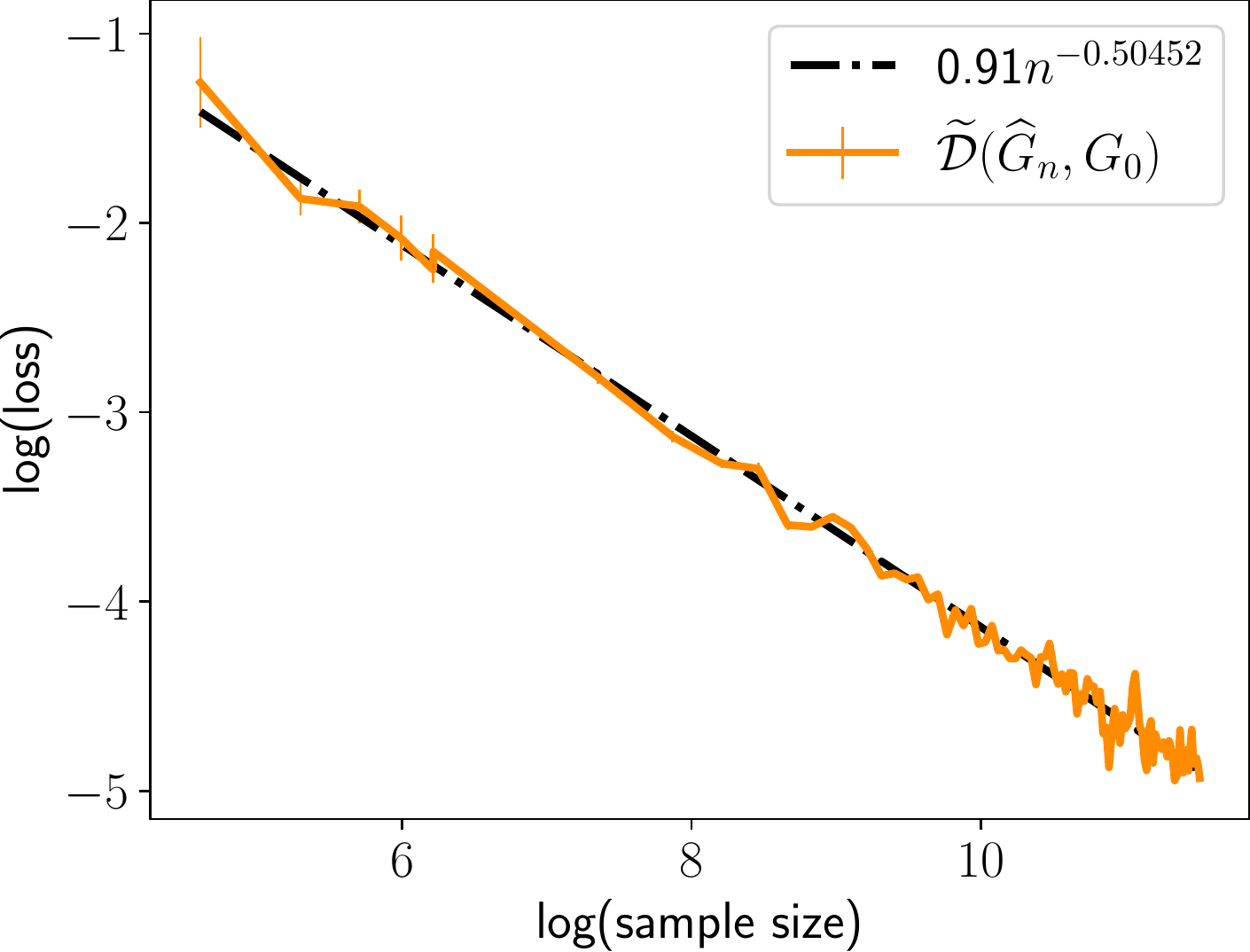}
        \caption{Model II, $k = 4$}	
        \label{subfig_model_type_II_k4}
    \end{subfigure}%
    \hspace{0.6cm}
    \begin{subfigure}{.45\textwidth}
        \centering
        \includegraphics[scale = .48]{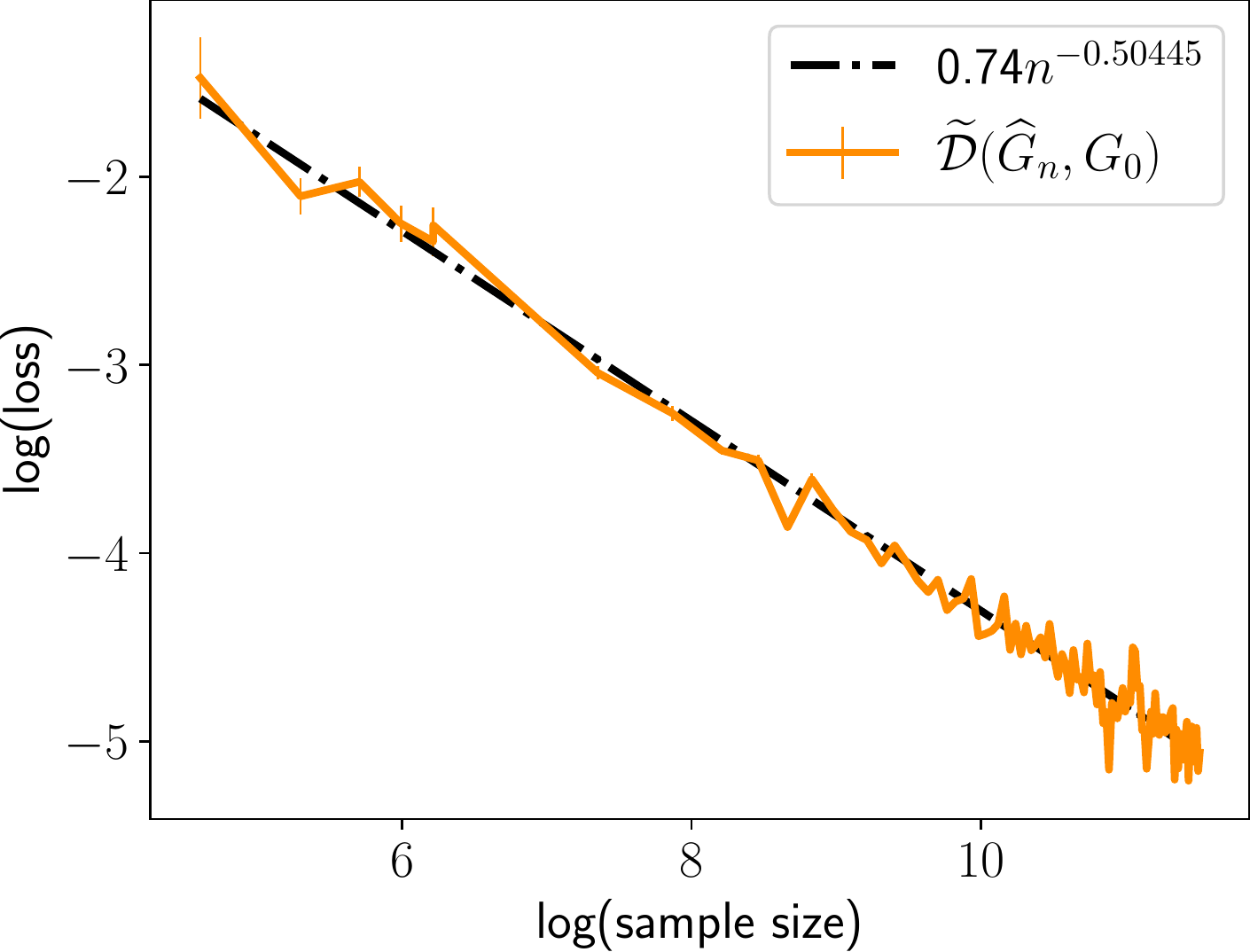}
        \caption{Model II, $k = 5$}	
        \label{subfig_model_type_II_k5}
    \end{subfigure}%
    \caption{Log-log scaled plots of the empirical mean of discrepancies $\overline{D}(\widehat{G}_n,G_0)$ and $\widetilde{D}(\widehat{G}_n,G_0)$   and $G_0$ (orange lines with error bars) and least-squares fitted linear regression (black dash-dotted lines) when $d=1$ and $k_0 = 3$. \label{fig_plot_model_type_1d}}
\end{figure*}

\section{EXPERIMENTS}
\label{sec_experiments}
In this section, we empirically validate the convergence rates of parameter estimation in four GMoE models which satisfy the assumptions of Type I and Type II settings, respectively, when $k_0 = 3$.
Note that for simplicity, we only perform a simulation study to illustrate the convergence rates of Theorems~\ref{theorem:type_1_setting} and \ref{theorem:type_2_setting} for the GMoE model when $X$ lies in one- and two-dimensional space with unknown location and scale parameters.
All code to reproduce our simulation study is publicly available\footnote{\url{https://github.com/Trung-TinNGUYEN/CRPE-GMoE}} and all simulations below were
performed in Python 3.9.13 on a standard Unix machine.

{\bf Numerical schemes.} 
In Model I, we set $G_0$ as follows: 
\begin{align*} 
    &\sum_{j=1}^{3} \pi^0_j \delta_{(c^0_j,\Gamma^0_j,a^0_j,b^0_j,\nu^0_j)} =0.3\delta_{(-0.1,\hspace{.05cm} 0.04,\hspace{.05cm} 0.40,\hspace{.05cm} 0.34,\hspace{.05cm} 0.01)} \nonumber\\
    &+0.4\delta_{(0.1,\hspace{.05cm} 0.02,\hspace{.05cm} -0.71,\hspace{.05cm} -0.33,\hspace{.05cm} 0.03)} + 0.3\delta_{(0.5,\hspace{.05cm} 0.01,\hspace{.05cm} 0,\hspace{.05cm} 0.2,\hspace{.05cm} 0.02)}.
\end{align*}
For Model II, we consider the same setting as in Model I but with $c^0_1 = 0$ and $b^0_1 = 0.3$.
To demonstrate the claim that the empirical convergence rates of parameter estimation under the Type I (Model III) and Type II (Model IV) settings also hold in higher dimensions, we conduct a numerical simulation for $d = 2$ and $k_0 = 3$.
In Model III, we set $G_0 $ as   
\begin{align*} 
    &\sum_{j=1}^{3} \pi^0_j \delta_{(c^0_j,\Gamma^0_j,a^0_j,b^0_j,\nu^0_j)} =
    0.3\delta_{(-0.1\cdot\mathbf{1}_d,\hspace{.02cm}  0.04\cdot\mathbf{I}_d,\hspace{.02cm} 0.4\cdot\mathbf{1}_d,\hspace{.02cm} 0.34,\hspace{.02cm} 0.01)}\nonumber\\ 
    & +0.4\delta_{(0.1\cdot\mathbf{1}_d,\hspace{.02cm} 0.02\cdot\mathbf{I}_d,\hspace{.02cm} -0.71\cdot\mathbf{1}_d,\hspace{.02cm} -0.33,\hspace{.02cm} 0.03)} \nonumber
    \\ 
    &
    + 0.3\delta_{(0.5\cdot\mathbf{1}_d,\hspace{.02cm} 0.01\cdot\mathbf{I }_d,\hspace{.02cm} \mathbf{0}_d,\hspace{.02cm} 0.2,\hspace{.02cm} 0.02)},
\end{align*}
where $\mathbf{1}_d=(1,1)$, $\mathbf{0}_d=(0,0)$ and $\mathbf{I}_d$ is the identity matrix of size $d$. In Model IV, we consider the same setting of $G_0$ as in Model III but with $c_1^0=\mathbf{0}_d$ and $b_1^0=0.3$.

\begin{figure*}[!ht]
    \centering
    \begin{subfigure}{.45\textwidth}
        \centering
        \includegraphics[scale = .48]{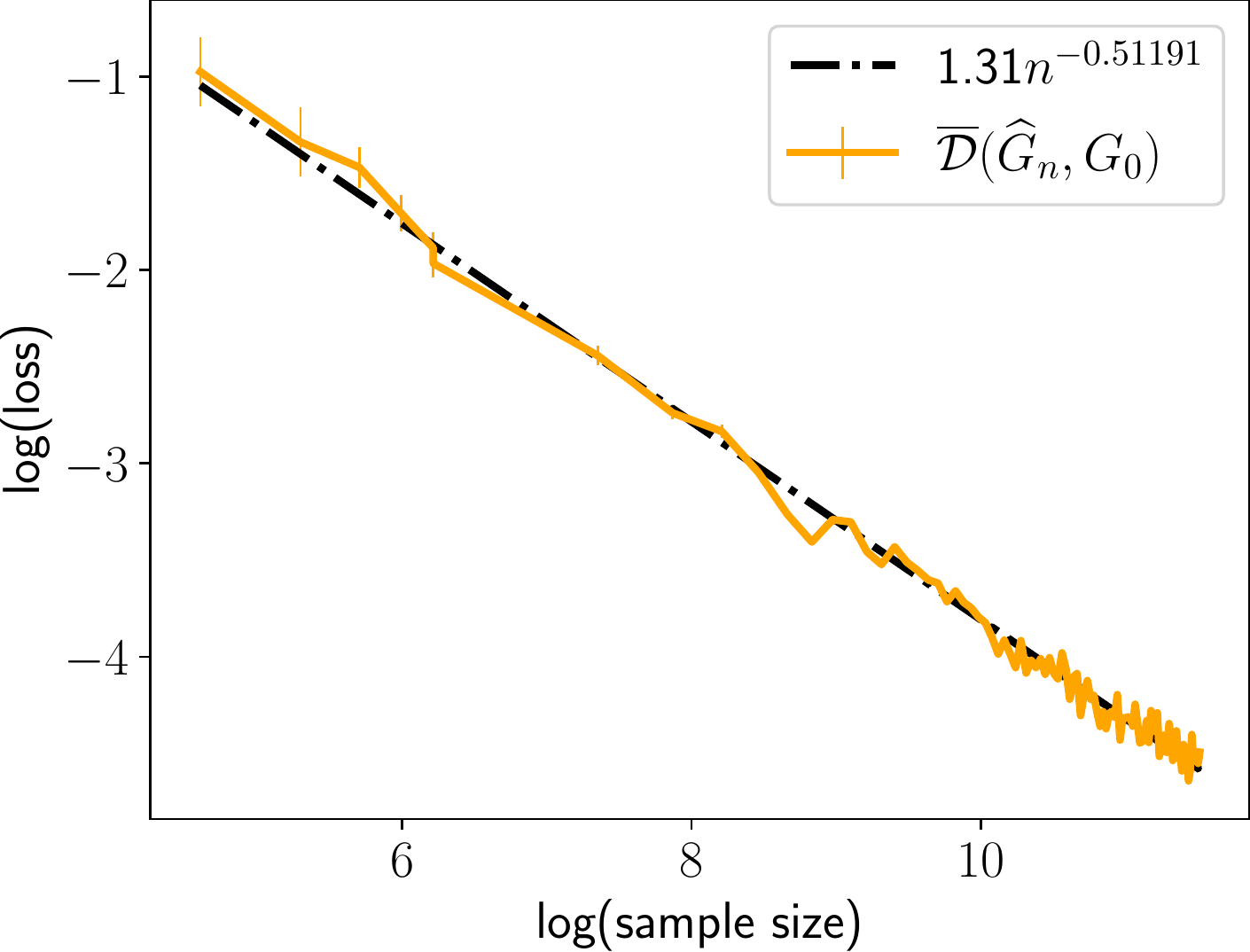}
        \caption{Model III, $k = 4$}	
        \label{subfig_model_type_III_k4}
    \end{subfigure}
    \hspace{0.6cm}
    \begin{subfigure}{.45\textwidth}
        \centering
        \includegraphics[scale = .48]{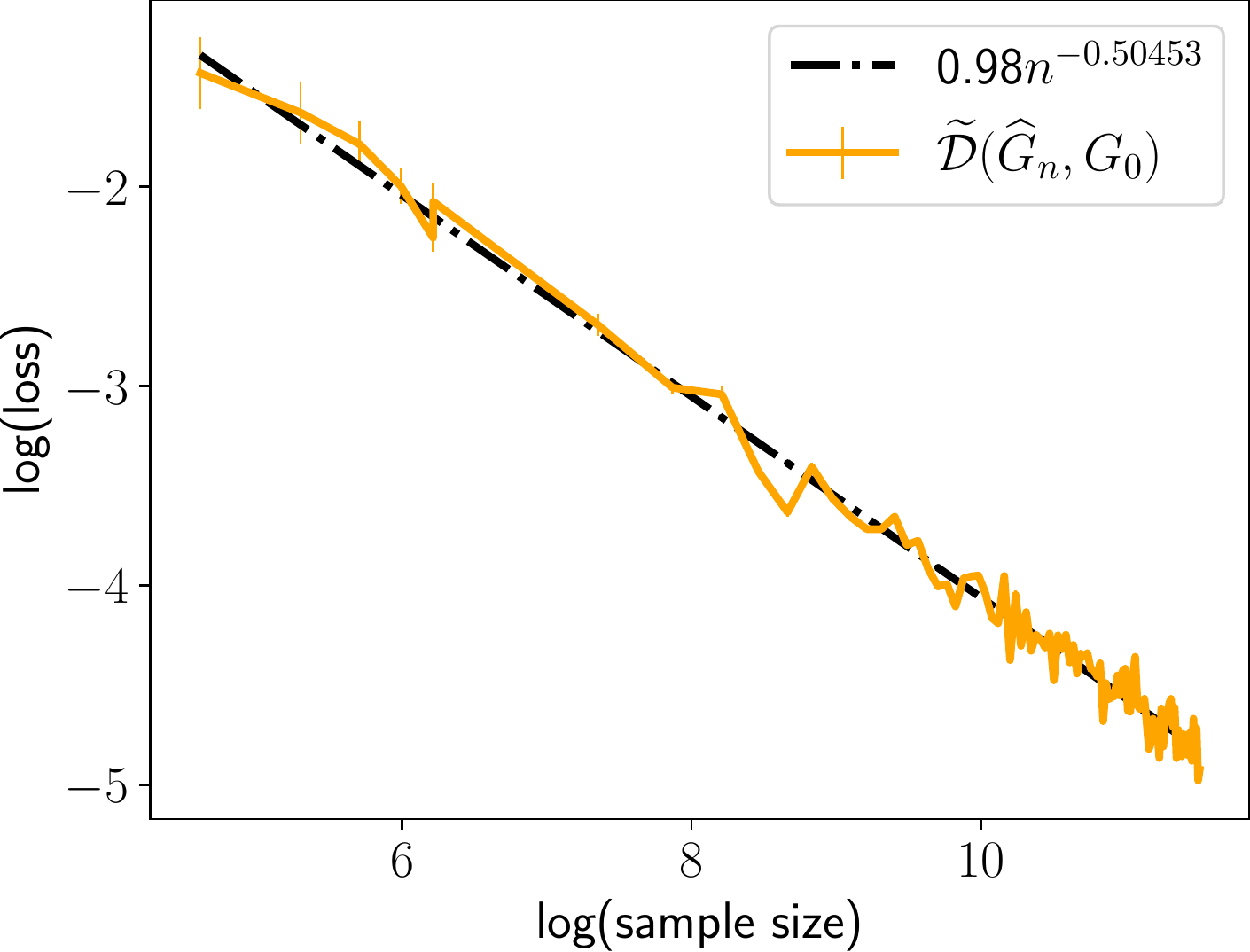}
        \caption{Model III, $k = 5$}	
        \label{subfig_model_type_III_k5}
    \end{subfigure}%
    \vspace{0.1cm}
    \begin{subfigure}{.45\textwidth}
        \centering
        \includegraphics[scale = .48]{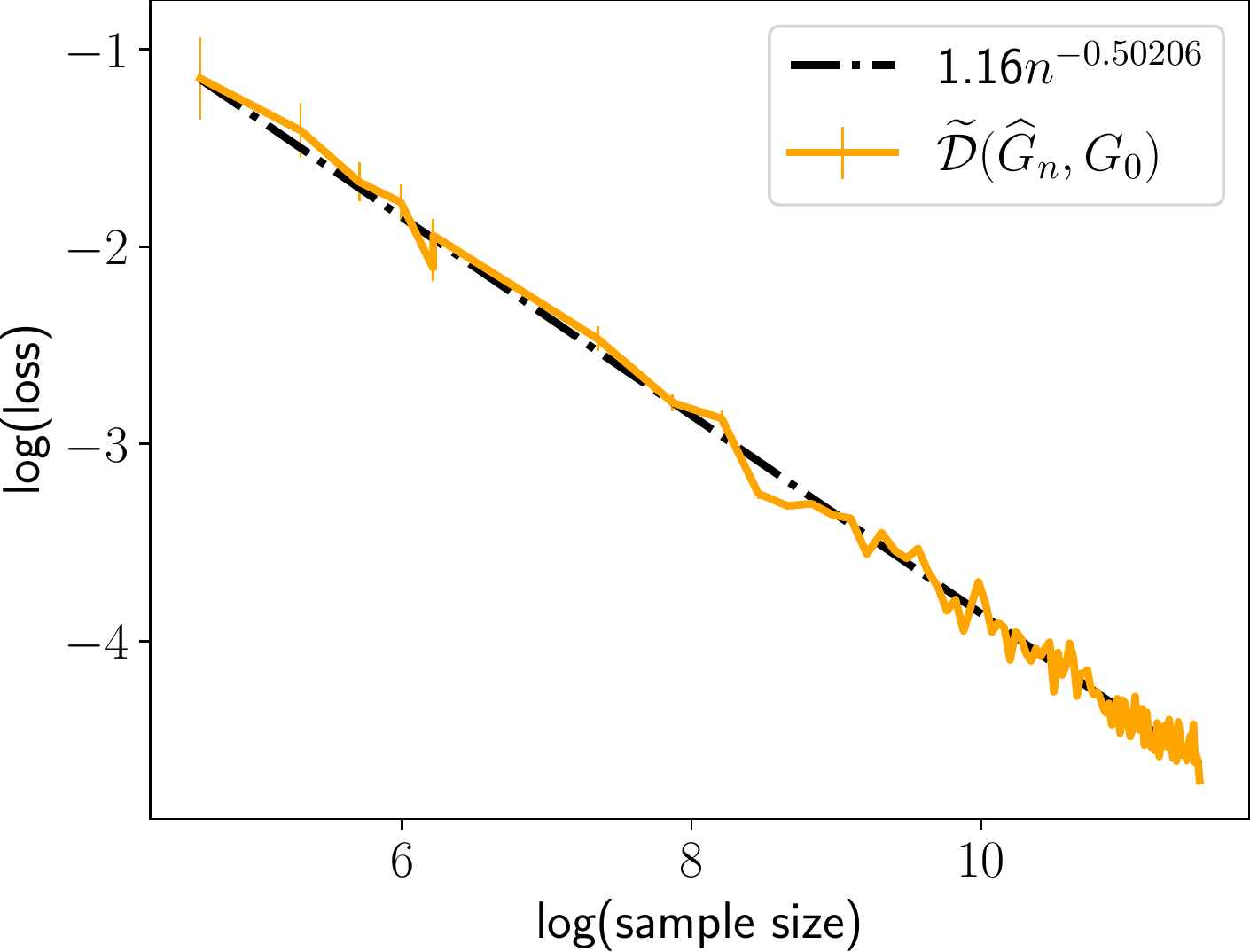}
        \caption{Model IV, $k = 4$}	
        \label{subfig_model_type_IV_k4}
    \end{subfigure}%
    \hspace{0.6cm}
    \begin{subfigure}{.45\textwidth}
        \centering
        \includegraphics[scale = .48]{figures/plot_model4_K5_n0_0_n100000_rep20.pdf}
        \caption{Model IV, $k = 5$}	
        \label{subfig_model_type_IV_k5}
    \end{subfigure}%
    \caption{Log-log scaled plots of the empirical mean of discrepancies $\overline{D}(\widehat{G}_n,G_0)$ and $\widetilde{D}(\widehat{G}_n,G_0)$ (orange lines with error bars) and least-squares fitted linear regression (black dash-dotted lines) when $d=2$ and $k_0 = 3$. \label{fig_plot_model_type_2d}}
    \vspace{-0.6cm}
\end{figure*}
{\bf Numerical details.} In accordance with the hierarchical GMoE setting of \eqref{eq_marginal_forward}, we generate 20 samples $(X_i,Y_i)_{i\in[n]}$ of size $n$ for each setting, given $100$ different choices of sample size $n$ between $10^2$ and $10^5$.
Then, we compute the MLE $\widehat{G}_n$ \wrt a number of components $k$ for each sample. For both of these settings, we choose $k \in \left\{k_0+1, k_0+2\right\}$ with corresponding $\overline{r}, \widetilde{r}\in \{4,6\}$ using Lemmas \ref{lemma_r_bar} and \ref{lemma_r_tilde}.
Here we implement the MLE using the EM algorithm for GMoE. This is a simplification of a general hybrid GMoE-EM from~\cite[Section 5]{deleforge_high-dimensional_2015}.
We choose the convergence criteria $\epsilon = 10^{-5}$ and $2000$ maximum EM iterations. Our goal is to illustrate the theoretical properties of the estimator $\widehat{G}_n$. Therefore, we have initialized the EM algorithm in a favourable way. 
More specifically, we first randomly partitioned the set $\{1,\ldots,k\}$ into $k_0$ index sets $J_1,\ldots,J_{k_0}$, each containing at least one point, for any given $k$ and $k_0$ and for each replication.
Finally, we sampled $c^0_j$ (\resp $\Gamma^0_j,a^0_j,b^0_j,\nu^0_j$) from a unique Gaussian distribution centered on $c^0_t$ (\resp $\Gamma^0_t,a^0_t,b^0_t,\nu^0_t$), with vanishing covariance so that $j \in J_t$.

{\bf Empirical convergence rates.} The empirical mean of discrepancies $\overline{D}$ and $\widetilde{D}$ between $\widehat{G}_n$ and $G_0$, and the choice of $k$ for Models I-II are reported in Figure~\ref{fig_plot_model_type_1d}.
It can be observed from Figure~\ref{fig_plot_model_type_1d} that those average discrepancies vanish at a rate of order $\mathcal{O}(n^{-1/2})$, which matches the results of Theorems~\ref{theorem:type_1_setting} and~\ref{theorem:type_2_setting},
where the only theoretical assumption that can be violated is the global convergence of the MLE. Note that the use of the joint density function allows the GMoE to be linked to a hierarchical mixture model, which guarantees global convergence for parameter estimation for arbitrary dimensions, see recent advances, \eg~\cite{kwon_minimax_2021,kwon_em_2020,kwon_global_2019}. We can therefore guarantee that the rates in Theorems~\ref{theorem:type_1_setting} and~\ref{theorem:type_2_setting} also hold in higher dimensions.
Indeed, it can be observed from Figure \ref{fig_plot_model_type_2d} that the average discrepancies $\overline{D}(\widehat{G}_n,G_0)$ and $\widetilde{D}(\widehat{G}_n,G_0)$ also approach zero at the rate of order $\mathcal{O}(n^{-1/2})$ for $d=2$, confirming the empirical behaviour of Theorems 1 and 2 under the high dimensional settings.


\section{CONCLUSION}\label{sec_conclusion}
In this paper, we conduct a convergence analysis for density estimation and parameter estimation in the Gaussian-gated mixture of experts (GMoE) under two complement settings of location parameters of the gating function. We demonstrate that the density estimation rate remains parametric on the sample size under both settings. On the other hand, due to several challenges induced by the interior and exterior interactions among parameters arising in those settings, we have to solve two complex systems of polynomial equations and then propose two corresponding novel Voronoi loss functions among parameters. We show that these Voronoi losses are able to capture the dependence of parameter estimation rates on the number of fitted components, which are more accurate than those characterized by the generalized Wasserstein loss used in previous works. We believe that our current techniques can be extended to the GMoE model with general experts in \cite{ho_convergence_2022} and to the hierarchical MoE for exponential family models in \cite{jiang_hierarchical_1999}. In addition, understanding the convergence behavior of least squares estimation under the deterministic MoE model \cite{nguyen2024squares} with Gaussian gate is also a potential direction. However, we leave such non-trivial developments for future work.

\section*{Acknowledgements}
NH acknowledges support from the NSF IFML 2019844 and the NSF AI Institute for Foundations of Machine Learning.

\bibliographystyle{abbrv}
\bibliography{aistats_references}

\clearpage
\onecolumn
\appendix


In this supplementary material, we first include an illustration of Voronoi cells in Appendix~\ref{appendix_Voronoi} to help the readers understand this concept better. 
Then, we provide the proof of Theorem~\ref{theorem:type_1_setting} and Theorem~\ref{theorem:type_2_setting} in Appendix~\ref{sec_appendix_Main_Results}. Finally, proofs for the remaining results are presented in Appendix~\ref{sec_appendix_Auxiliary_Results}. 

\section{ILLUSTRATION OF VORONOI CELLS}\label{appendix_Voronoi}
\begin{figure*}[!ht]
    \centering
     \centering
        \includegraphics[scale = .2]{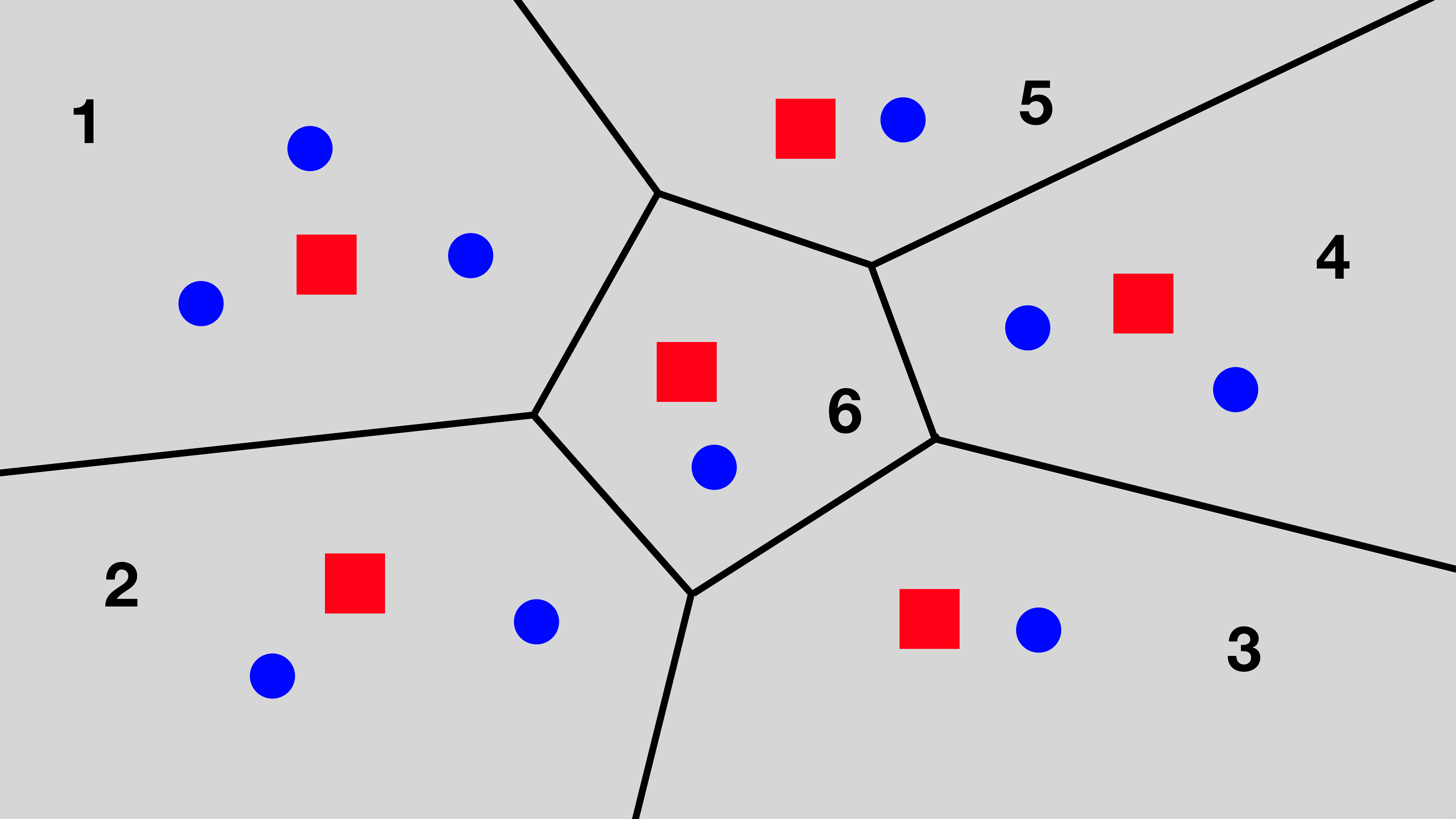}
    \caption{Illustration of Voronoi cells defined in equation~\eqref{eq_new_Voronoi_cell} when $k_0 = 6$ and $k = 10$. In this figure, red squares represent for true components (i.e. components of $G_0$), while blue circles indicate fitted components (i.e. components of $G$). By definition, each Voronoi cell is generated by one true component, and its cardinality is exactly the number of corresponding fitted components. For example, the square in cell 4 is approximated by two rounds, which means that the cardinality of cell 4 is two.
    \label{fig_Voronoi}}
\end{figure*}

In this appendix, we aim to illustrate the Voronoi cells defined in Section~\ref{sec_preliminaries}. For that purpose, let us recall the definition of that concept here. In particular, for any mixing measure $G\in\mathcal{O}_{k}(\Theta)$, the Voronoi cell $\mathcal{A}_j:=\mathcal{A}_j(G)$ generated by a true component $\theta^0_j:=(\cj,\gj,\aj,\bj,\vj)$ of $G_0$ is given by 
\begin{align}
\label{eq_new_Voronoi_cell}
    \mathcal{A}_j
    :=\{i\in[k]:\|\theta_i-\theta^0_j\|\leq\|\theta_i-\theta^0_{\ell}\|,\ \forall \ell\neq j\},
\end{align}
for any $j\in[k_0]$, where $\theta_i:=(c_i,\Gamma_i,a_i,b_i,\nu_i)$ is a component of $G$. Now, we provide an illustration of the above Voronoi cells under the setting when $k_0=6$ and $k=10$ in Figure~\ref{fig_Voronoi}.

\textbf{Connection to Theorem~\ref{theorem:type_1_setting}.} Under the Type I setting, parameters of the true components $(\cj,\gj,\aj,\bj,\vj)$ in cells 3, 5 and 6, which are fitted by one component, enjoy a parametric estimation rate of order $\mathcal{O}(n^{-1/2})$. Next, the rates for estimating parameters $c^0_1,b^0_1$ of the true component in cell 1, which are approximated by three components, stand at order $\mathcal{O}(n^{-1/2\bar{r}(3)})=\mathcal{O}(n^{-1/12})$, while those for $\Gamma^0_1,\nu^0_1$ are of order $\mathcal{O}(n^{-1/\bar{r}(3)})=\mathcal{O}(n^{-1/6})$. Meanwhile, the estimation rate for $a^0_1$ is independent of the cardinality of its corresponding Voronoi cell and remains stable at order $\mathcal{O}(n^{-1/4})$.

\textbf{Connection to Theorem~\ref{theorem:type_2_setting}.} Parameter estimation rates under the Type II setting share the same behavior as those in Theorem~\ref{theorem:type_1_setting} except for the rates of estimating $\aj$. More specifically, if $c^0_1=0$, the estimation rate for $a^0_1$ now depends on the cardinality of cell 1, and experiences a drop to order $\mathcal{O}(n^{-1/\widetilde{r}(3)})=\mathcal{O}(n^{-1/6})$.

\section{PROOF OF MAIN RESULTS}\label{sec_appendix_Main_Results}
Before going to the proofs for Theorems~\ref{theorem:type_1_setting} and \ref{theorem:type_2_setting} in  Appendices~\ref{appendix:type_1_setting} and \ref{appendix:type_2_setting}, respectively, let us define some necessary notations used throughout this appendix. Firstly, for any vector $v\in\mathbb{R}^d$, either $v_i$ or $v^{(i)}$ represents the $i$-th entry of $v$, while the sum of its entries is abbreviated as $|v|:=v_1+v_2+\ldots+v_d$. Next, for any vector $p\in\mathbb{N}^d$, we denote $v^p:=v_{1}^{p_{1}}v_{2}^{p_{2}}\ldots v_{d}^{p_{d}}$ and $p!:=p_1!p_2!\ldots p_d!$. Additionally, we sometimes use the notation $h_1$ and $h_2$ to denote the expert functions considered in this work. In particular, we define $h_1(X,a,b)=a^{\top}X+b$ as the mean expert function for any $X\in\mathcal{X}\subset\mathbb{R}^d$, $a\in\mathbb{R}^d$ and $b\in\mathbb{R}$, whereas $h_2(X,\nu)=\nu$ stands for the variance expert function for any $\nu\in\mathbb{R}_+$. Finally, since parameters in the proofs for Theorem~\ref{theorem:type_1_setting} and Theorem~\ref{theorem:type_2_setting} belong to various high-dimensional spaces, we summarize their domains in Table~\ref{table:thm1} and Table~\ref{table:thm2}, respectively, to help readers keep track of them.
\begin{table}[h!]
\centering
\begin{tabular}{ |c|c|c|c|c|c|c|c|c|c|c|c|c|c|c|c| } 
\hline
& $c$ &  $\Gamma$ & $a$ & $b$ & $\nu$ & $\tau_1$ & $\tau_2$ & $\alpha_1$ &  $\alpha_2$ & $\alpha_3$ & $\alpha_4$ & $\alpha_5$ & $\ell_1$ & $\ell_2$ \\
\hline 
Thm 1 & $\mathbb{R}^d$ & $\mathcal{S}_d^+$ & $\mathbb{R}^d$ & $\mathbb{R}$ & $\mathbb{R}_+$ & $\mathbb{N}^d$ & $\mathbb{N}$ & $\mathbb{N}^d$ & $\mathbb{N}^{d\times d}$ & $\mathbb{N}^d$ & $\mathbb{N}$ & $\mathbb{N}^d$ & N/A & N/A \\  
\hline
\end{tabular}
\caption{Domains for parameters used in the proof of Theorem~\ref{theorem:type_1_setting}}
\label{table:thm1}
\end{table}

\begin{table}[h!]
\centering
\begin{tabular}{ |c|c|c|c|c|c|c|c|c|c|c|c|c|c|c|c| } 
\hline
& $c$ &  $\Gamma$ & $a$ & $b$ & $\nu$ & $\tau_1$ & $\tau_2$ & $\alpha_1$ &  $\alpha_2$ & $\alpha_3$ & $\alpha_4$ & $\alpha_5$ & $\ell_1$ & $\ell_2$ \\
\hline 
Thm 2 & $\mathbb{R}$ & $\mathbb{R}_+$ & $\mathbb{R}$ & $\mathbb{R}$ & $\mathbb{R}_+$ & $\mathbb{N}$ & $\mathbb{N}$ & $\mathbb{N}$ & $\mathbb{N}$ & $\mathbb{N}$ & $\mathbb{N}$ & $\mathbb{N}$ & $\mathbb{N}$ & $\mathbb{N}$ \\  
\hline
\end{tabular}
\caption{Domains for parameters used in the proof of Theorem~\ref{theorem:type_2_setting}}
\label{table:thm2}
\end{table}

\subsection{Proof of Theorem~\ref{theorem:type_1_setting}}
\label{appendix:type_1_setting}
Our goal is to show the following inequality:
\begin{align}
    \label{eq:inverse_inequality}
    \inf_{G\in\mathcal{O}_{k,\beta}(\Theta)}{V(p_{G},p_{G_0})}/{\overline{D}(G,G_0)}>0,
\end{align}
which implies the desired Total Variation lower bound $V(p_{\widehat{G}_n},p_{G_0})\gtrsim \overline{D}(\widehat{G}_n,G_0)$. Given this bound, the joint density estimation rate in Proposition~\ref{prop_density_rate} then leads to the convergence rate of the MLE $\widehat{G}_n$ to $G_0$ under the loss $\overline{D}$ as follows:
\begin{align*}
    \bbP(\overline{D}(\widehat{G}_n,G_0)>C_3\sqrt{\log (n)/n})\lesssim n^{-C_4},
\end{align*}
for some universal constants $C_3$ and $C_4$. Note that the infimum in equation~\eqref{eq:inverse_inequality} is subject to all the mixing measures in the set $\mathcal{O}_{k,\beta}(\Theta):=\{G=\sum_{i=1}^{k'}\pi_i\delta_{(c_i,\Gamma_i,a_i,b_i,\nu_i)}:1\leq k'\leq k, ~\sum_{i=1}^{k}\pi_i=1,\ \pi_i\geq\beta,\  (c_i,\Gamma_i,a_i,b_i,\nu_i)\in\Theta\}$, for some positive constant $\beta$.
Now, we divide the proof of inequality~\eqref{eq:inverse_inequality} into two parts which we refer to as local bound and global bound.

\textbf{Local bound}: Firstly, we will prove the local version of inequality~\eqref{eq:inverse_inequality}:
\begin{align}
    \label{eq:local_version}
    \lim_{\varepsilon\to 0}\inf_{\substack{G\in\mathcal{O}_{k,\beta}(\Theta)\\ \overline{D}(G,G_0)\leq\varepsilon}}{V(p_{G},p_{G_0})}/{\overline{D}(G,G_0)}>0.
\end{align}
Assume by contrary that the claim in equation~\eqref{eq:local_version} does not hold. Then, there exists a sequence of mixing measures $G_n=\sum_{i=1}^{k_n}\pin\delta_{(c^n_i,\Gamma^n_i, a^n_i,b^n_i,\nu^n_i)}\in\mathcal{O}_{k,\beta}(\Theta)$ such that $\overline{D}(G_n,G_0)\to 0$ and ${V(p_{G_n},p_{G_0})}/{\overline{D}(G_n,G_0)}\to 0$ as $n\to\infty$. Moreover, since $k_n\leq k$ for all $n\in\mathbb{N}$, we can replace $(G_n)$ by its subsequence that admits a fixed number of atoms $k_n=k'\leq k$. Additionally, $\mathcal{A}_j=\mathcal{A}_j^n$ does not change with $n$ for all $j\in[k_0]$. 

\textbf{Step 1 - Taylor expansion for density decomposition:} Now, we consider the quantity
\begin{align*}
    &p_{G_n}(X,Y)-p_{G_0}(X,Y)\\
    =&\sum_{j:|\mathcal{A}_j|>1}\sum_{i\in\mathcal{A}_j}\pin[f_{\mathcal{L}}(X| \cin,\gin)f_{\mathcal{D}}(Y| (\ain)^{\top}X+\bin,\vin)-f_{\mathcal{L}}(X| \cj,\gj)f_{\mathcal{D}}(Y| (\aj)^{\top}X+\bj,\vj)]\\
    +&\sum_{j:|\mathcal{A}_j|=1}\sum_{i\in\mathcal{A}_j}\pin[f_{\mathcal{L}}(X| \cin,\gin)f_{\mathcal{D}}(Y| (\ain)^{\top}X+\bin,\vin)-f_{\mathcal{L}}(X| \cj,\gj)f_{\mathcal{D}}(Y| (\aj)^{\top}X+\bj,\vj)]\\
    +&\sum_{j=1}^{k_0}\left(\sum_{i\in\mathcal{A}_j}\pin-\pizeroj\right)f_{\mathcal{L}}(X| \cj,\gj)f_{\mathcal{D}}(Y| (\aj)^{\top}X+\bj,\vj)\\
    :&=A_n+B_n+E_n.
\end{align*}
For each $j\in[k_0]:|\mathcal{A}_j|>1$, we perform a Taylor expansion up to the $\brj$-th order, and then rewrite $A_n$ with a note that $\alpha=(\alpha_1,\alpha_2,\alpha_3,\alpha_4,\alpha_5)\in\mathbb{N}^{d}\times\mathbb{N}^{d\times d}\times\mathbb{N}^d\times\mathbb{N}\times\mathbb{N}$ as follows:
\begin{align*}
    A_n&=\sum_{j:|\mathcal{A}_j|>1}\sum_{i\in\mathcal{A}_j}\pin\sum_{|\alpha|=1}^{\brj}\frac{1}{\alpha!}(\dcijn)^{\alpha_1}(\dgijn)^{\alpha_2}(\daijn)^{\alpha_3}(\dbijn)^{\alpha_4}(\dvijn)^{\alpha_5}\\
    &\hspace{2cm} \times\dfrac{\partial^{|\alpha_1|+|\alpha_2|}f_{\mathcal{L}}}{\partial c^{\alpha_1}\partial \Gamma^{\alpha_2}}(X| \cj,\gj)\cdot\dfrac{\partial^{|\alpha_3|+\alpha_4+\alpha_5}f_{\mathcal{D}}}{\partial a^{\alpha_3}\partial b^{\alpha_4}\partial \nu^{\alpha_5}}(Y| (\aj)^{\top}X+\bj,\vj) + R_1(X,Y)\\
\end{align*}  
where $R_1(X,Y)$ is a remainder term such that $R_1(X,Y)/\overline{D}(G_n,G_0)\to 0$ as $n\to\infty$, which is due to the uniform Holder continuity of a location-scale Gaussian family. Since $f_{\mathcal{L}}$ $d$-dimensional Gaussian density functions, we have the following partial differential equation (PDE):
\begin{align*}
    \dfrac{\partial^{|\alpha_1|+|\alpha_2|}f_{\mathcal{L}}}{\partial c^{\alpha_1}\partial \Gamma^{\alpha_2}}(X| \cj,\gj)=\dfrac{1}{2^{|\alpha_2|}}\cdot\dfrac{\partial^{|\alpha_1|+2|\alpha_2|}f_{\mathcal{L}}}{\partial c^{\tau(\alpha_1,\alpha_2)}}(X| \cj,\gj),
\end{align*}
where $\tau(\alpha_1,\alpha_2):=\Big(\alpha_1^{(v)}+\sum_{u=1}^d(\alpha_2^{(uv)}+\alpha_2^{(vu)})\Big)_{v=1}^d=\Big(\alpha_1^{(v)}+2\sum_{u=1}^d\alpha_2^{(uv)}\Big)_{v=1}^d\in\mathbb{N}^d$.
Similarly, as $f_{\mathcal{L}}$ is an univariate Gaussian density function, then
\begin{align*}
    \dfrac{\partial^{|\alpha_3|+\alpha_4+\alpha_5}f_{\mathcal{D}}}{\partial a^{\alpha_3}\partial b^{\alpha_4}\partial \nu^{\alpha_5}}(Y| (\aj)^{\top}X+\bj,\vj)&=\dfrac{X^{\alpha_3}}{2^{\alpha_5}}\cdot\dfrac{\partial^{|\alpha_3|+\alpha_4+2\alpha_5}f_{\mathcal{D}}}{\partial h_1^{|\alpha_3|+\alpha_4+2\alpha_5}}(Y| (\aj)^{\top}X+\bj,\vj),
\end{align*}
where $h_1(X,a,b)=a^{\top}X+b$ is the mean expert function. Combine these results together, $A_n$ can be represented as follows:
\begin{align*}    A_n&=\sum_{j:|\mathcal{A}_j|>1}\sum_{i\in\mathcal{A}_j}\pin\sum_{|\alpha|=1}^{\brj}\frac{1}{\alpha!}(\dcijn)^{\alpha_1}(\dgijn)^{\alpha_2}(\daijn)^{\alpha_3}(\dbijn)^{\alpha_4}(\dvijn)^{\alpha_5}\\
    & \times\dfrac{1}{2^{|\alpha_2|}}\dfrac{\partial^{|\alpha_1|+2|\alpha_2|}f_{\mathcal{L}}}{\partial c^{\tau(\alpha_1,\alpha_2)}}(X| \cj,\gj)\cdot\dfrac{X^{\alpha_3}}{2^{\alpha_5}}\dfrac{\partial^{|\alpha_3|+\alpha_4+2\alpha_5}f_{\mathcal{D}}}{\partial h_1^{|\alpha_3|+\alpha_4+2\alpha_5}}(Y| (\aj)^{\top}X+\bj,\vj) + R_1(X,Y),\\
\end{align*}  
Let $\tau_1=\tau(\alpha_1,\alpha_2)\in\mathbb{N}^d$ and $\tau_2=\alpha_4+2\alpha_5\in\mathbb{N}$, we can rewrite $A_n$ as
\begin{align*}    A_n&=\sum_{j:|\mathcal{A}_j|>1}\sum_{|\alpha_3|=0}^{\brj}\sum_{|\tau_1|+\tau_2=0}^{2(\brj-|\alpha_3|)}\sum_{\substack{ \tau(\alpha_1,\alpha_2)=\tau_1\\\alpha_4+2\alpha_5=\tau_2}}\sum_{i\in\mathcal{A}_j}\dfrac{\pin}{2^{|\alpha_2|+\alpha_5}\alpha!}(\dcijn)^{\alpha_1}(\dgijn)^{\alpha_2}(\daijn)^{\alpha_3}\\
    &\times(\dbijn)^{\alpha_4}(\dvijn)^{\alpha_5}\cdot X^{\alpha_3}\dfrac{\partial^{|\tau_1|}f_{\mathcal{L}}}{\partial c^{\tau_1}}(X| \cj,\gj)\dfrac{\partial^{|\alpha_3|+\tau_2}f_{\mathcal{D}}}{\partial h_1^{|\alpha_3|+\tau_2}}(Y| (\aj)^{\top}X+\bj,\vj)+R_1(X,Y),
\end{align*}
Analogously, for each $j\in[k_0]:|\mathcal{A}_j|=1$, by means of Taylor expansion up to the first order, $B_n$ is rewritten as follows:
\begin{align}
    \label{eq:Bn_formulation}
    B_n&=\sum_{j:|\mathcal{A}_j|=1}\sum_{|\alpha_3|=0}^{1}\sum_{|\tau_1|+\tau_2=0}^{2(1-|\alpha_3|)}\sum_{\substack{ \tau(\alpha_1,\alpha_2)=\tau_1\\\alpha_4+2\alpha_5=\tau_2}}\sum_{i\in\mathcal{A}_j}\dfrac{\pin}{2^{|\alpha_2|+\alpha_5}\alpha!}(\dcijn)^{\alpha_1}(\dgijn)^{\alpha_2}(\daijn)^{\alpha_3}\nonumber\\
    &\times(\dbijn)^{\alpha_4}(\dvijn)^{\alpha_5}\cdot X^{\alpha_3}\dfrac{\partial^{|\tau_1|}f_{\mathcal{L}}}{\partial c^{\tau_1}}(X| \cj,\gj)\dfrac{\partial^{|\alpha_3|+\tau_2}f_{\mathcal{D}}}{\partial h_1^{|\alpha_3|+\tau_2}}(Y| (\aj)^{\top}X+\bj,\vj)+R_2(X,Y),
\end{align}
where $R_2(X,Y)$ is a remainder such that $R_2(X,Y)/\overline{D}(G_n,G_0)\to 0$ as $n\to\infty$.

It is worth noting that $A_n$, $B_n$ and $E_n$ can be treated as linear combinations of elements of the following set:
\begin{align}
    \mathcal{F}:=\Big\{X^{\alpha_3}\dfrac{\partial^{|\tau_1|}f_{\mathcal{L}}}{\partial c^{\tau_1}}(X| \cj,\gj)\dfrac{\partial^{|\alpha_3|+\tau_2}f_{\mathcal{D}}}{\partial h_1^{|\alpha_3|+\tau_2}}(Y| &(\aj)^{\top}X+\bj,\vj):~j\in[k_0],0\leq|\alpha_3|\leq\brj,\nonumber\\
    \label{eq:F_formulation}
    &~0\leq|\tau_1|+\tau_2\leq 2(\brj-|\alpha_3|)\Big\}.
\end{align}
Let $T^{n}_{\alpha_3,\tau_1,\tau_2}(j)$ be the coefficients of $$X^{\alpha_3}\dfrac{\partial^{|\tau_1|}f_{\mathcal{L}}}{\partial c^{\tau_1}}(X| \cj,\gj)\dfrac{\partial^{|\alpha_3|+\tau_2}f_{\mathcal{D}}}{\partial h_1^{|\alpha_3|+\tau_2}}(Y| (\aj)^{\top}X+\bj,\vj)$$ in the representations of $A_n$, $B_n$ and $E_n$. 

\textbf{Step 2 - Proof of non-vanishing coefficients by contradiction:} Assume that all the coefficients in the representations of $A_n/\overline{D}(G_n,G_0)$, $B_n/\overline{D}(G_n,G_0)$ and $E_n/\overline{D}(G_n,G_0)$ go to 0 as $n\to\infty$. Then, by taking the summation of the absolute values of coefficients in $E_n/\overline{D}(G_n,G_0)$, which are  $|T_{\zerod,\zerod,0}(j)|/\overline{D}(G_n,G_0)$ for all $j\in[k_0]$, we get that
\begin{align}
    \label{eq:vanish_En}
    \dfrac{1}{\overline{D}(G_n,G_0)}\cdot\sum_{j=1}^{k_0}\left|\sum_{i\in\mathcal{A}_j}\pin-\pizeroj\right|\to 0.
\end{align}
Subsequently, from the formulation of $B_n$ in equation~\eqref{eq:Bn_formulation}, we have
\begin{align*}
    \dfrac{1}{\overline{D}(G_n,G_0)}\cdot\sum_{j:|\mathcal{A}_j|=1}\sum_{i\in\mathcal{A}_j}\pin(\norm{\dcijn}_1+\norm{\dgijn}_1+\norm{\daijn}_1+|\dbijn|+|\dvijn|)\to 0.
\end{align*}
It follows from the topological equivalence of $1$-norm and $2$-norm that
\begin{align}
    \label{eq:vanish_Bn}
    \dfrac{1}{\overline{D}(G_n,G_0)}\cdot\sum_{j:|\mathcal{A}_j|=1}\sum_{i\in\mathcal{A}_j}\pin(\norm{\dcijn}+\norm{\dgijn}+\norm{\daijn}+|\dbijn|+|\dvijn|)\to 0.
\end{align}
Next, from the formulation of $A_n$, by combining all terms of the form $|T_{\alpha_3,\zerod,0}(j)|/\overline{D}(G_n,G_0)$ where $j\in[k_0]:|\mathcal{A}_j|>1$ and $\alpha_3\in\{2e_1,2e_2,\ldots,2e_d\}$ with $e_u:=(0,\ldots,0,\underbrace{1}_{\textit{u-th}},0,\ldots,0)$ being a one-hot vector in $\mathbb{R}^d$ for all $u\in[d]$, we obtain that
\begin{align}
    \label{eq:vanish_alpha3_0_0}
    \dfrac{1}{\overline{D}(G_n,G_0)}\cdot\sum_{j:|\mathcal{A}_j|>1}\sum_{i\in\mathcal{A}_j}\pin\norm{\daijn}^2\to 0.
\end{align}
Putting the results in  equations~\eqref{eq:vanish_En}, \eqref{eq:vanish_Bn} and \eqref{eq:vanish_alpha3_0_0} together with the formulation of $\overline{D}(G_n,G_0)$ in equation~\eqref{eq_D_bar_formulation}, we deduce that
\begin{align*} \dfrac{\sum_{j:|\mathcal{A}_j|>1}\sum_{i\in\mathcal{A}_j}\pin(\norm{\dcijn}^{\brj}+\norm{\dgijn}^{\brj/2}+|\dbijn|^{\brj}+|\dvijn|^{\brj/2})}{\overline{D}(G_n,G_0)}\to 1.
\end{align*}
As a result, we can find an index $j^*\in[k_0]$ such that $|\mathcal{A}_j|>1$ and
\begin{align}
    \label{eq:mimic_i_bar}
    \dfrac{\sum_{i\in\mathcal{A}_{j^*}}\pin(\norm{\Delta c_{ij^*}^n}^{\bar{r}(\mathcal{A}_{j^*})}+\norm{\Delta\Gamma_{ij^*}^n}^{\bar{r}(\mathcal{A}_{j^*})/2}+|\Delta b_{ij^*}^n|^{\bar{r}(\mathcal{A}_{j^*})}+|\Delta \nu_{ij^*}^{n}|^{\bar{r}(\mathcal{A}_{j^*})/2})}{\overline{D}(G_n,G_0)}\not\to 0.
\end{align}
Without loss of generality (WLOG), we may assume that $j^*=1$. Now, we divide our arguments into two main cases as follows:

\textbf{Case 1}:~ $\dfrac{1}{\overline{D}(G_n,G_0)}\cdot\sum_{i\in\mathcal{A}_1}\pin(\norm{\dcione}^{\brone}+\norm{\dgione}^{\brone/2})\not\to 0.$

Here, we continue to split this case into two possibilities:

\textbf{Case 1.1}:~ $\dfrac{1}{\overline{D}(G_n,G_0)}\cdot\sum_{i\in\mathcal{A}_1}\pin\Big(\norm{\dcione}^{\brone}+\norm{((\dgione)^{(uu)})_{u=1}^d}^{\brone/2}\Big)\not\to 0.$

In this case, it must hold for some index $u^*\in[d]$ that
\begin{align}
    \label{eq:denominatior_not_vanish}
    \dfrac{1}{\overline{D}(G_n,G_0)}\cdot\sum_{i\in\mathcal{A}_1}\pin\Big(|(\dcione)^{(u^*)}|^{\brone}+|(\dgione)^{(u^*u^*)}|^{\brone/2}\Big)\not\to 0.
\end{align}
WLOG, we assume that $u^*=1$ throughout case 1.1. In the representation of $A_n$, we consider the following coefficient: 
\begin{align}
    \label{eq:coefficient_0_tau_0}
    T_{\zerod,\tau_1,0}(1)=\sum_{i\in\mathcal{A}_1}\sum_{\substack{\alpha_1,\alpha_2:\\ \tau(\alpha_1,\alpha_2)=\tau_1}}\frac{\pin}{2^{|\alpha_2|}\alpha_1!\alpha_2!}(\dcione)^{\alpha_1}(\dgione)^{\alpha_2},
\end{align}
where $\tau_1\in\mathbb{N}^d$ such that $\tau_1^{(u)}=0$ for all $u=2,\ldots,d$. Thus, the constraint $\tau(\alpha_1,\alpha_2)=\tau_1$ holds if and only if $\alpha_1^{(u)}=\alpha_2^{(u1)}=\alpha_2^{(1v)}=\alpha_2^{(uv)}=0$ for all $u,v=2,\ldots,d$. Therefore, by assumption, we have
\begin{align}
    \label{eq:numerator_vanish}
    \dfrac{T_{\zerod,\tau_1,0}(1)}{\overline{D}(G_n,G_0)}=\frac{1}{\overline{D}(G_n,G_0)}\cdot\sum_{i\in\mathcal{A}_1}\sum_{\substack{\alpha_1^{(1)}+2\alpha_2^{(11)}=\tau_1^{(1)}}}\frac{\pin}{2^{\alpha_2^{(11)}}\alpha_1^{(1)}!~\alpha_2^{(11)}!}(\dcione)^{\alpha_1^{(1)}}(\dgione)^{\alpha_2^{(11)}}\to 0.
\end{align}
Collect results in  equations~\eqref{eq:denominatior_not_vanish} and \eqref{eq:numerator_vanish}, we obtain that
\begin{align}
    \label{eq:ratio_vanish}
    \dfrac{\sum_{i\in\mathcal{A}_1}\sum_{\substack{\alpha_1^{(1)}+2\alpha_2^{(11)}=\tau_1^{(1)}}}\dfrac{\pin}{2^{\alpha_2^{(11)}}\alpha_1^{(1)}!~\alpha_2^{(11)}!}(\dcione)^{\alpha_1^{(1)}}(\dgione)^{\alpha_2^{(11)}}}{\sum_{i\in\mathcal{A}_1}\pin\Big(|(\dcione)^{(1)}|^{\brone}+|(\dgione)^{(11)}|^{\brone/2}\Big)}\to 0.
\end{align}
Next, we define $\overline{M}_n=\max\{|(\dcione)^{(1)}|,|(\dgione)^{(11)}|^{1/2}:i\in\mathcal{A}_1\}$ and $\overline{\pi}_n=\max_{i\in\mathcal{A}_1}\pin$. For any $i\in\mathcal{A}_1$, it is clear that the sequence of positive real numbers $(\pin/\overline{\pi}_n)$ is bounded, therefore, we can replace it by its subsequence that admits a non-negative limit denoted by $p_i^2=\lim_{n\to\infty}\pin/\overline{\pi}_n$. In addition, let us denote $(\dcione)^{(1)}/\overline{M}_n\to \eta_i$ and $(\dgione)^{(11)}/2\overline{M}_n^2\to \gamma_i$. From the formulation of $\mathcal{O}_{k,\beta}(\Theta)$, since $\pin\geq \beta$, the real numbers $p_i$ will not vanish, and at least one of them is equal to 1. Analogously, at least one of the $\eta_i$ and $\gamma_i$ is equal to either 1 or $-1$. 

Note that $\sum_{i\in\mathcal{A}_1}\pin\Big(|(\dcione)^{(1)}|^{\brone}+|(\dgione)^{(11)}|^{\brone/2}\Big)/(\overline{\pi}_n\overline{M}_n^{\tau_1^{(1)}})\not\to 0$ for all $\tau_1^{(1)}\in[\brone]$. Thus, we are able to divide both the numerator and the denominator in equation~\eqref{eq:ratio_vanish} by $\overline{\pi}_n\overline{M}_n^{\tau_1^{(1)}}$ and let $n\to\infty$ in order to achieve the following system of polynomial equations:
\begin{align*}
    \sum_{i\in\mathcal{A}_1}\sum_{\alpha_1^{(1)}+2\alpha_2^{(11)}=\tau_1^{(1)}}\dfrac{p_i^2\eta_i^{\alpha_1^{(1)}}\gamma_i^{\alpha_2^{(11)}}}{\alpha_1^{(1)}!~\alpha_2^{(11)}!}=0, \quad   \tau_1^{(1)}\in[\brone].
\end{align*}
However, by the definition of $\brone$, the above system cannot admit any non-trivial solutions, which is a contradiction. Thus, case 1.1 cannot happen.

\textbf{Case 1.2}:~ $\dfrac{1}{\overline{D}(G_n,G_0)}\cdot\sum_{i\in\mathcal{A}_1}\pin\Big(\norm{((\dgione)^{(uv)})_{1\leq u\neq v\leq d}}^{\brone/2}\Big)\not\to 0.$

In this case, it must hold for some indices $u^*\neq v^*$ that 
\begin{align*}
    \dfrac{1}{\overline{D}(G_n,G_0)}\cdot\sum_{i\in\mathcal{A}_1}\pin|(\dgione)^{(u^*v^*)}|^{\brone/2}\not\to 0.
\end{align*}
Recall that $|\mathcal{A}_1|>1$, or equivalently, $|\mathcal{A}_1|\geq 2$, we have that $\brone\geq 4$. Therefore, the above equation leads to
\begin{align}
     \label{eq:gamma_not_vanish}
    \dfrac{1}{\overline{D}(G_n,G_0)}\cdot\sum_{i\in\mathcal{A}_1}\pin|(\dgione)^{(u^*v^*)}|^{2}\not\to 0.
\end{align}
WLOG, we assume that $u^*=1$ and $v^*=2$ throughout case 1.2. We continue to consider the coefficient $T_{\zerod,\tau_1,0}$ in equation~\eqref{eq:coefficient_0_tau_0} with $\tau_1=(2,2,0,\ldots,0)\in\mathbb{N}^d$. By assumption, we have ${T_{\zerod,\tau_1,0}}/{\overline{D}(G_n,G_0)}\to 0$, which together with equation~\eqref{eq:gamma_not_vanish} imply that
\begin{align}
    \label{eq:case_12_ratio_vanish}
   \dfrac{\sum_{i\in\mathcal{A}_1}\sum_{\substack{\alpha_1,\alpha_2:\\ \tau(\alpha_1,\alpha_2)=\tau_1}}\dfrac{\pin}{2^{|\alpha_2|}\alpha_1!\alpha_2!}(\dcione)^{\alpha_1}(\dgione)^{\alpha_2}}{\sum_{i\in\mathcal{A}_1}\pin|(\dgione)^{(12)}|^{2}}\to 0.
\end{align}
Similarly, by combining the fact that case 1.1 does not hold and the result in equation~\eqref{eq:gamma_not_vanish}, we get
\begin{align*}
    \dfrac{\sum_{i\in\mathcal{A}_1}\pin\Big(\norm{\dcione}^{\brone}+\norm{((\dgione)^{(uu)})_{u=1}^d}^{\brone/2}\Big)}{\sum_{i\in\mathcal{A}_1}\pin|(\dgione)^{(12)}|^{2}}\to 0.
\end{align*}
Since $\brone\geq 4$, the above limit indicates that any terms in equation~\eqref{eq:case_12_ratio_vanish} with $\alpha_1^{(u)}>0$ and $\alpha_2^{(uu)}>0$ for $u\in\{1,2\}$ will vanish. Consequently, we deduce from equation~\eqref{eq:case_12_ratio_vanish} that
\begin{align*}
    1=\dfrac{\sum_{i\in\mathcal{A}_1}\pin|(\dgione)^{(12)}|^{2}}{\sum_{i\in\mathcal{A}_1}\pin|(\dgione)^{(12)}|^{2}}\to 0,
\end{align*}
which is a contradiction. Thus, case 1.2 cannot happen.

\textbf{Case 2}:~ $\dfrac{1}{\overline{D}(G_n,G_0)}\cdot\sum_{i\in\mathcal{A}_1}\pin(|\dbione|^{\brone}+|\dvione|^{\brone/2})\not\to 0.$

In this case, we consider the coefficient $T_{\zerod,\zerod,0}(1)$ in the formulation of $A_n$. By assumption, 
\begin{align*}
    \dfrac{T_{\zerod,\zerod,0}(1)}{\overline{D}(G_n,G_0)}=\dfrac{1}{\overline{D}(G_n,G_0)}\cdot\sum_{i\in\mathcal{A}_1}\pin\sum_{\substack{\alpha_4,\alpha_5:\\ \alpha_4+2\alpha_5=\tau_2}}\dfrac{(\dbione)^{\alpha_4}(\dvione)^{\alpha_5}}{2^{\alpha_5}\alpha_4!\alpha_5!}\to 0.
\end{align*}
Consequently, we obtain that
\begin{align*}
    \dfrac{\sum_{i\in\mathcal{A}_1}\pin\sum_{\substack{\alpha_4,\alpha_5:\\ \alpha_4+2\alpha_5=\tau_2}}\dfrac{(\dbione)^{\alpha_4}(\dvione)^{\alpha_5}}{2^{\alpha_5}\alpha_4!\alpha_5!}}{\sum_{i\in\mathcal{A}_1}\pin(|\dbione|^{\brone}+|\dvione|^{\brone/2})}\to 0.
\end{align*}
By employing the same arguments for showing that the equation~\eqref{eq:ratio_vanish} does not hold in case 1.1, we obtain that the above limit does not hold, either. Thus, case 2 cannot happen.

From the above results of the two main cases, we conclude that not all the coefficients in the representations of $A_n/\overline{D}(G_n,G_0)$, $B_n/\overline{D}(G_n,G_0)$ and $E_n/\overline{D}(G_n,G_0)$ vanish as $n\to\infty$.

\textbf{Step 3 - Application of Fatou's lemma:} Subsequently, we denote by $m_n$ the maximum of the absolute values of the coefficients in the representations of $A_n/\overline{D}(G_n,G_0)$, $B_n/\overline{D}(G_n,G_0)$ and $E_n/\overline{D}(G_n,G_0)$, that is,
\begin{align*}
    m_n:=\max_{(\alpha_3,\tau_1,\tau_2,j)\in\mathcal{S}}|T^{n}_{\alpha_3,\tau_1,\tau_2}(j)|/\overline{D}(G_n,G_0),
\end{align*}
where the constraint set $\mathcal{S}$ is defined as
\begin{align*}
    \mathcal{S}:=\Big\{(\alpha_3,\tau_1,\tau_2,j)\in\mathbb{N}^d\times\mathbb{N}^d\times\mathbb{N}\times[k_0]:0\leq |\alpha_3|\leq \brj,0\leq |\tau_1|,\tau_2\leq 2(\brj-|\alpha_3|)\Big\}.
\end{align*}
Additionally, we define $T^{n}_{\alpha_3,\tau_1,\tau_2}(j)/m_n\to\xi_{\alpha_3,\tau_1,\tau_2}(j)$ as $n\to\infty$ for all $(\alpha_3,\tau_1,\tau_2,j)\in\mathcal{S}$. Since not all the coefficients in the representations of $A_n/\overline{D}(G_n,G_0)$, $B_n/\overline{D}(G_n,G_0)$ and $E_n/\overline{D}(G_n,G_0)$ vanish as $n\to\infty$, at least one among $\xi_{\alpha_3,\tau_1,\tau_2}(j)$ is different from zero and $m_n\not\to 0$. Then, by applying the Fatou's lemma, we get that
\begin{align*}
    0=\lim_{n\to\infty}\dfrac{1}{m_n}\cdot\dfrac{2V(p_{G_n},p_{G})}{\overline{D}(G_n,G_0)}\geq\int\liminf_{n\to\infty}\dfrac{1}{m_n}\cdot\dfrac{|p_{G_n}(X,Y)-p_{G}(X,Y)|}{\overline{D}(G_n,G_0)}~\dint(X,Y)\geq 0.
\end{align*}
Moreover, by definition, we have
\begin{align*}
    \dfrac{1}{m_n}\cdot&\dfrac{p_{G_n}(X,Y)-p_{G}(X,Y)}{\overline{D}(G_n,G_0)}\\
    &\to \sum_{(\alpha_3,\tau_1,\tau_2,j)\in\mathcal{S}}\xi_{\alpha_3,\tau_1,\tau_2}(j)X^{\alpha_3}\dfrac{\partial^{|\tau_1|}f_{\mathcal{L}}}{\partial c^{\tau_1}}(X| \cj,\gj)\dfrac{\partial^{|\alpha_3|+\tau_2}f_{\mathcal{D}}}{\partial h_1^{|\alpha_3|+\tau_2}}(Y| (\aj)^{\top}X+\bj,\vj).
\end{align*}
As a consequence, we achieve that
\begin{align*}
    \sum_{(\alpha_3,\tau_1,\tau_2,j)\in\mathcal{S}}\xi_{\alpha_3,\tau_1,\tau_2}(j)X^{\alpha_3}\dfrac{\partial^{|\tau_1|}f_{\mathcal{L}}}{\partial c^{\tau_1}}(X| \cj,\gj)\dfrac{\partial^{|\alpha_3|+\tau_2}f_{\mathcal{D}}}{\partial h_1^{|\alpha_3|+\tau_2}}(Y| (\aj)^{\top}X+\bj,\vj)=0,
\end{align*}
for almost surely $(X,Y)$. Since elements of the set $\mathcal{F}$ defined in equation~\eqref{eq:F_formulation} are linearly independent (proof of this claim is deferred to the end of this proof), the above equation implies that $\xi_{\alpha_3,\tau_1,\tau_2}(j)=0$ for all $(\alpha_3,\tau_1,\tau_2,j)\in\mathcal{S}$, which contradicts the fact that at least one among $\xi_{\alpha_3,\tau_1,\tau_2}(j)$ is different from zero. Hence, we reach the conclusion in equation~\eqref{eq:local_version}, which indicates that there exists some $\varepsilon_0>0$ such that
\begin{align*}
    \inf_{\substack{G\in\mathcal{O}_{k,\beta}(\Theta)\\ \overline{D}(G,G_0)\leq\varepsilon_0}}{V(p_{G},p_{G_0})}/{\overline{D}(G,G_0)}>0.
\end{align*}
\textbf{Global bound}: Given the above result, in order to achieve the inequality in equation~\eqref{eq:inverse_inequality}, we only need to prove its following global version:
\begin{align*}
    \inf_{\substack{G\in\mathcal{O}_{k,\beta}(\Theta)\\ \overline{D}(G,G_0)>\varepsilon_0}}{V(p_{G},p_{G_0})}/{\overline{D}(G,G_0)}>0.
\end{align*}
Assume by contrary that the above claim is not true. Then, there exists a sequence $G^{\prime}_n\in\mathcal{O}_{k,\beta}(\Theta)$ such that ${V(p_{G^{\prime}_n},p_{G_0})}/{\overline{D}(G^{\prime}_n,G_0)}\to 0$ and $\overline{D}(G^{\prime}_n,G_0)>\varepsilon_0$ for all $n\in\mathbb{N}$. Since the set $\Theta$ is compact, we can replace $G^{\prime}_n$ by its subsequence that converges to some mixing measure $G^{\prime}\in\mathcal{O}_{k,\beta}(\Theta)$. Consequently, we deduce that $\overline{D}(G^{\prime},G_0)=\lim_{n\to\infty}\overline{D}(G^{\prime}_n,G_0)\geq\varepsilon_0$. This result together with the fact that ${V(p_{G^{\prime}_n},p_{G_0})}/{\overline{D}(G^{\prime}_n,G_0)}\to 0$ lead to the limit $V(p_{G^{\prime}_n},p_{G_0})\to 0$ as $n\to\infty$. Again, by applying the Fatou's lemma, we obtain that
\begin{align*}
    0=\lim_{n\to\infty}2V(p_{G^{\prime}_n},p_{G_0})&\geq\int\liminf_{n\to\infty}|p_{G^{\prime}_n}(X,Y)-p_{G_0}(X,Y)|\dint(X,Y)\\
    &=\int|p_{G^{\prime}}(X,Y)-p_{G_0}(X,Y)|\dint(X,Y)\geq 0.
\end{align*}
As a consequence, we have that $p_{G^{\prime}}(X,Y)=p_{G_0}(X,Y)$ for almost surely $(X,Y)$. Due to the identifiability of the model, this equality leads to $G^{\prime}\equiv G_0$, which contradicts the bound $\overline{D}(G^{\prime},G_0)\geq\varepsilon_0>0$. Hence, we achieve the conclusion in equation~\eqref{eq:inverse_inequality}.

\textbf{Linear independence of elements in $\mathcal{F}$}: For completion, we will demonstrate elements of the set $\mathcal{F}$ defined in equation~\eqref{eq:F_formulation} are linearly independent by definition. In particular, assume that there exist real numbers $\xi_{\alpha_3,\tau_1,\tau_2}(j)$, where $(\alpha_3,\tau_1,\tau_2,j)\in\mathcal{S}$, such that the following equation holds for almost surely $(X,Y)$:
\begin{align*}
    \sum_{(\alpha_3,\tau_1,\tau_2,j)\in\mathcal{S}}\xi_{\alpha_3,\tau_1,\tau_2}(j)X^{\alpha_3}\dfrac{\partial^{|\tau_1|}f_{\mathcal{L}}}{\partial c^{\tau_1}}(X| \cj,\gj)\dfrac{\partial^{|\alpha_3|+\tau_2}f_{\mathcal{D}}}{\partial h_1^{|\alpha_3|+\tau_2}}(Y| (\aj)^{\top}X+\bj,\vj)=0.
\end{align*}
Now, we rewrite the above equation as follows:
\begin{align}
\label{eq:independent}    \sum_{j=1}^{k_0}\sum_{\omega=0}^{2\brj}\Big(\sum_{|\alpha_3|+\tau_2=\omega}\sum_{|\tau_1|=0}^{2(\brj-\alpha_3)-\tau_2}&\xi_{\alpha_3,\tau_1,\tau_2}(j)X^{\alpha_3}\cdot\dfrac{\partial^{|\tau_1|}f_{\mathcal{L}}}{\partial c^{\tau_1}}(X| \cj,\gj)\Big)\nonumber\\
&\times\dfrac{\partial^{\omega}f_{\mathcal{D}}}{\partial h_1^{\omega}}(Y| (\aj)^{\top}X+\bj,\vj)=0,
\end{align}
for almost surely $(X,Y)$. As $(\aj,\bj,\vj)$ for $j\in[k_0]$ are $k_0$ distinct tuples, we deduce that $((\aj)^{\top}X+\bj,\vj)$ for $j\in[k_0]$ are also $k_0$ distinct tuples for almost surely $X$. Thus, for almost surely $X$, one has $\dfrac{\partial^{\omega}f_{\mathcal{D}}}{\partial h_1^{\omega}}(Y| (\aj)^{\top}X+\bj,\vj)$ for $j\in[k_0]$ and $0\leq\omega\leq 2\brj$ are linearly independent with respect to $Y$. Given that result, the equation~\eqref{eq:independent} indicates that for almost surely $X$,
\begin{align*}
    \sum_{|\alpha_3|+\tau_2=\omega}\sum_{|\tau_1|=0}^{2(\brj-\omega)}\xi_{\alpha_3,\tau_1,\tau_2}(j)X^{\alpha_3}\cdot\dfrac{\partial^{|\tau_1|}f_{\mathcal{L}}}{\partial c^{\tau_1}}(X| \cj,\gj)=0,
\end{align*}
for all $j\in[k_0]$ and $0\leq\omega\leq 2\brj$. Note that for each $j\in[k_0]$ and $0\leq \omega\leq \brj$, the left hand side of the above equation can be viewed as a high-dimensional polynomial of two random vectors $X$ and $X-\cj$ ($\cj\neq \zerod$) in $\mathcal{X}$, which is a compact set in $\mathbb{R}^d$. As a result, the above equation holds when $\xi_{\alpha_3,\tau_1,\tau_2}(j)=0$ for all $j\in[k_0]$, $0\leq\omega\leq 2\brj$, $|\alpha_3|+\tau_2=\omega$ and $|\tau_1|\leq 2(\brj-\alpha_3)-\tau_2$. This is equivalent to $\xi_{\alpha_3,\tau_1,\tau_2}(j)=0$ for all $(\alpha_3,\tau_1,\tau_2,j)\in\mathcal{S}$.

Hence, we conclude that the elements of $\mathcal{F}$ are linearly independent.

\subsection{Proof of Theorem~\ref{theorem:type_2_setting}}
\label{appendix:type_2_setting}
In order to reach the conclusion in Theorem~\ref{theorem:type_2_setting}, we only need to demonstrate the following inequality:
\begin{align}
    \label{eq:inverse_inequality_2}
    \inf_{G\in\mathcal{O}_{k,\beta}(\Theta)}V(p_{G},p_{G_0})/\Dtilde(G,G_0)>0.
\end{align}
In this proof, we will only prove the following local version of inequality~\eqref{eq:inverse_inequality_2} while the global version can be argued in the same fashion as in Appendix~\eqref{appendix:type_1_setting}:
\begin{align}
\label{eq:local_version_2}
\lim_{\varepsilon\to0}\inf_{\substack{G\in\mathcal{O}_{k,\beta}(\Theta)\\ \Dtilde(G,G_0)\leq\varepsilon}}V(p_{G},p_{G_0})/\Dtilde(G,G_0)>0.
\end{align}
Assume that the claim in equation~\eqref{eq:local_version_2} is not true. This indicates that we can find a sequence of mixing measures $G_n=\sum_{i=1}^{k_n}\pi_i^n\delta_{(c_i,\Gamma_i^n,a_i^n,b_i^n,\nu_i^n)}\in\mathcal{O}_{k,\beta}(\Theta)$ that satisfies: $\Dtilde(G_n,G_0)\to 0$ and $V(p_{G_n},p_{G_0})/\Dtilde(G_n,G_0)\to0$ as $n\to\infty$. Additionally, since $k_n\leq k$ for all $n\in\mathbb{N}$, we are able to replace $(G_n)$ by its subsequence which admits a fixed number of atoms $k_n\leq k'\leq k$ and $\mathcal{A}_j=\mathcal{A}_j^n$ is independent of $n$ for all $j\in[k_0]$. 

\textbf{Step 1 - Taylor expansion for density decomposition:} Next, we take into account the quantity
\begin{align*}
    &p_{G_n}(X,Y)-p_{G_0}(X,Y)\\
    &=\sum_{j:|\mathcal{A}_j|>1}\sum_{i\in\mathcal{A}_j}\pin[f_{\mathcal{L}}(X| \cin,\gin)f_{\mathcal{D}}(Y| \ain X+\bin,\vin)-f_{\mathcal{L}}(X| \cj,\gj)f_{\mathcal{D}}(Y| \aj X+\bj,\vj)]\\
    &+\sum_{j:|\mathcal{A}_j|=1}\sum_{i\in\mathcal{A}_j}\pin[f_{\mathcal{L}}(X| \cin,\gin)f_{\mathcal{D}}(Y| \ain X+\bin,\vin)-f_{\mathcal{L}}(X| \cj,\gj)f_{\mathcal{D}}(Y| \aj X+\bj,\vj)]\\
    &+\sum_{j=1}^{k_0}\left(\sum_{i\in\mathcal{A}_j}\pin-\pizeroj\right)f_{\mathcal{L}}(X| \cj,\gj)f_{\mathcal{D}}(Y| \aj X+\bj,\vj)\\
    &:=A_n+B_n+E_n.
\end{align*}
For each $j\in[k_0]:|\mathcal{A}_j|>1$, by means of Taylor expansion up to the $\trj$-th order, $A_n$ can be rewritten as follows with a note that $\alpha=(\alpha_1,\alpha_2,\alpha_3,\alpha_4,\alpha_5)\in\mathbb{N}^{5}$:
\begin{align*}
    A_n&=\sum_{j:|\mathcal{A}_j|>1}\sum_{i\in\mathcal{A}_j}\pin\sum_{|\alpha|=1}^{\trj}\frac{1}{\alpha!}(\dcijn)^{\alpha_1}(\dgijn)^{\alpha_2}(\daijn)^{\alpha_3}(\dbijn)^{\alpha_4}(\dvijn)^{\alpha_5}\\
    &\qquad \times\dfrac{\partial^{\alpha_1+\alpha_2}f_{\mathcal{L}}}{\partial c^{\alpha_1}\partial \Gamma^{\alpha_2}}(X| \cj,\gj)\cdot\dfrac{\partial^{\alpha_3+\alpha_4+\alpha_5}f}{\partial a^{\alpha_3}\partial b^{\alpha_4}\partial \nu^{\alpha_5}}(Y| \aj X+\bj,\vj) + R_1(X,Y)\\
    &=\sum_{j:|\mathcal{A}_j|>1}\sum_{i\in\mathcal{A}_j}\pin\sum_{|\alpha|=1}^{\trj}\frac{1}{\alpha!}(\dcijn)^{\alpha_1}(\dgijn)^{\alpha_2}(\daijn)^{\alpha_3}(\dbijn)^{\alpha_4}(\dvijn)^{\alpha_5}\\
    &\qquad\times\dfrac{1}{2^{\alpha_2}}\dfrac{\partial^{\alpha_1+2\alpha_2}f_{\mathcal{L}}}{\partial c^{\alpha_1+2\alpha_2}}(X| \cj,\gj)\cdot\dfrac{X^{\alpha_3}}{2^{\alpha_5}}\dfrac{\partial^{\alpha_3+\alpha_4+2\alpha_5}f_{\mathcal{D}}}{\partial h_1^{\alpha_3+\alpha_4+2\alpha_5}}(Y| \aj X+\bj,\vj) + R_3(X,Y),
\end{align*}
where $R_3(X,Y)$ is Taylor remainder such that $R_3(X,Y)/\Dtilde(G_n,G_0)\to 0$. Since $\cj$ is equal to zero when $j\in[\ktilde]$ and different from zero otherwise, the formulation of $\dfrac{\partial^{\alpha_1+2\alpha_2}f_{\mathcal{L}}}{\partial c^{\alpha_1+2\alpha_2}}(X| \cj,\gj)$ will vary when $j\in[\ktilde]$ compared to $\ktilde+1\leq j\leq k_0$. Thus, we will consider these two cases of $j$ separately.

For $j\in[\ktilde]$, when $\alpha_1$ is an even integer, we have
\begin{align*}
    \dfrac{\partial^{\alpha_1+2\alpha_2} f_{\mathcal{L}}}{\partial c^{\alpha_1+2\alpha_2}}(X| \cj,\gj)=
    \begin{cases}
    \sum_{w=0}^{\alpha_1/2+\alpha_2}t_{2w, \alpha_{1} + 2\alpha_{2}}X^{2w},\quad j\in[\ktilde]\\ 
    \textbf{}\\
    \sum_{w=0}^{\alpha_1/2+\alpha_2}s_{2w, \alpha_{1} + 2\alpha_{2}}(X-\cj)^{2w},\quad \ktilde+1\leq j\leq k_0.
    \end{cases}
\end{align*}
On the other hand, when $\alpha_1$ is an odd integer, we get
\begin{align*}
    \dfrac{\partial^{\alpha_1+2\alpha_2} f_{\mathcal{L}}}{\partial c^{\alpha_1+2\alpha_2}}(X| \cj,\gj)=
    \begin{cases}
    \sum_{w=0}^{(\alpha_1-1)/2+\alpha_2}t_{2w+1, \alpha_{1} + 2\alpha_{2}}X^{2w+1},\quad j\in[\ktilde]\\ 
    \textbf{}\\
    \sum_{w=0}^{(\alpha_1-1)/2+\alpha_2}s_{2w+1, \alpha_{1} + 2\alpha_{2}}(X-\cj)^{2w+1},\quad \ktilde+1\leq j\leq k_0.
    \end{cases}
\end{align*}
By combining both cases, we rewrite $A_n$ as follows:
\begin{align}
    A_n&=\sum_{\substack{j:|\mathcal{A}_j|>1,\\ j\in[\ktilde]}}\sum_{i\in\mathcal{A}_j}\sum_{\ell_1+\ell_2=1}^{2\trj}~\sum_{\alpha\in\mathcal{I}_{\ell_1,\ell_2}}\dfrac{\pin}{2^{\alpha_2+\alpha_5}\alpha!}(\dcijn)^{\alpha_1}(\dgijn)^{\alpha_2}(\daijn)^{\alpha_3}\nonumber\\
    &\qquad\times (\dbijn)^{\alpha_4}(\dvijn)^{\alpha_5}~t_{\ell_1-\alpha_3,\alpha_1+2\alpha_2}X^{\ell_1}\dfrac{\partial^{\ell_2}f_{\mathcal{D}}}{\partial h_1^{\ell_2}}(Y| \aj X+\bj,\vj)f_{\mathcal{L}}(X| \cj,\gj)\nonumber\\
    +&\sum_{\substack{j:|\mathcal{A}_j|>1\\ \ktilde+1\leq j\leq k_0}}\sum_{i\in\mathcal{A}_j}\sum_{\alpha_3=0}^{\trj}\sum_{\tau_1+\tau_2=0}^{2(\trj-\alpha_3)}\sum_{\substack{ \alpha_1+2\alpha_2=\tau_1\\\alpha_4+2\alpha_5=\tau_2}}\dfrac{\pin}{2^{\alpha_2+\alpha_5}\alpha!}(\dcijn)^{\alpha_1}(\dgijn)^{\alpha_2}(\daijn)^{\alpha_3}\nonumber\\
    \label{eq:An_new_form}
    &\times(\dbijn)^{\alpha_4}(\dvijn)^{\alpha_5}\times X^{\alpha_3}\dfrac{\partial^{\tau_1}f_{\mathcal{L}}}{\partial c^{\tau_1}}(X| \cj,\gj)\dfrac{\partial^{\alpha_3+\tau_2}f_{\mathcal{D}}}{\partial h_1^{\alpha_3+\tau_2}}(Y| \aj X+\bj,\vj)+R_3(X,Y),
\end{align}
where for any $0\leq\ell_1\leq 2\brj$ and $0\leq \ell_2\leq 2\brj-\ell_1$, we define
\begin{align*}
    \mathcal{I}_{\ell_1,\ell_2}:=\Big\{\alpha=(\alpha_i)_{i=1}^{5}\in\mathbb{N}^5:&~\alpha_1+2\alpha_2+\alpha_3\geq\ell_1, ~\alpha_3+\alpha_4+2\alpha_5=\ell_2,\\
    &~1\leq \alpha_1+\alpha_2+\ldots+\alpha_5\leq\trj\Big\}.
\end{align*}
Regarding the formulation of $B_n$, for each $j\in[k_0]:|\mathcal{A}_j|=1$, we perform a Taylor expansion up to the first order and obtain that
\begin{align}
    \label{eq:Bn_new_form}
    B_n
    &=\sum_{\substack{j:|\mathcal{A}_j|=1,\\ j\in[\ktilde]}}\sum_{i\in\mathcal{A}_j}\sum_{\ell_1+\ell_2=1}^{2}~\sum_{\alpha\in\mathcal{I}_{\ell_1,\ell_2}}\dfrac{\pin}{2^{\alpha_2+\alpha_5}\alpha!}(\dcijn)^{\alpha_1}(\dgijn)^{\alpha_2}(\daijn)^{\alpha_3}\nonumber\\
    &\qquad\times (\dbijn)^{\alpha_4}(\dvijn)^{\alpha_5}~t_{\ell_1-\alpha_3,\alpha_1+2\alpha_2}X^{\ell_1}\dfrac{\partial^{\ell_2}f_{\mathcal{D}}}{\partial h_1^{\ell_2}}(Y| \aj X+\bj,\vj)f_{\mathcal{L}}(X| \cj,\gj)\nonumber\\
    &+\sum_{\substack{j:|\mathcal{A}_j|=1\\ \ktilde+1\leq j\leq k_0}}\sum_{i\in\mathcal{A}_j}\sum_{\alpha_3=0}^{2}\sum_{\tau_1+\tau_2=0}^{2(1-\alpha_3)}\sum_{\substack{\alpha_1,\alpha_2:\\ \alpha_1+2\alpha_2=\tau_1}}\sum_{\substack{\alpha_4,\alpha_5:\\ \alpha_4+2\alpha_5=\tau_2}}\dfrac{\pin}{2^{\alpha_2+\alpha_5}\alpha!}(\dcijn)^{\alpha_1}(\dgijn)^{\alpha_2}(\daijn)^{\alpha_3}\nonumber\\
    &\qquad\times(\dbijn)^{\alpha_4}(\dvijn)^{\alpha_5}\times X^{\alpha_3}\dfrac{\partial^{\tau_1}f_{\mathcal{L}}}{\partial c^{\tau_1}}(X| \cj,\gj)\dfrac{\partial^{\alpha_3+\tau_2}f_{\mathcal{D}}}{\partial h_1^{\alpha_3+\tau_2}}(Y| \aj X+\bj,\vj)+R_4(X,Y),
\end{align}
where $R_4(X,Y)$ is a Taylor remainder such that $R_4(X,Y)/\Dtilde(G_n,G_0)\to 0$ as $n\to\infty$.

From equations~\eqref{eq:An_new_form} and \eqref{eq:Bn_new_form}, we can treat $A_n/\Dtilde(G_n,G_0)$, $B_n/\Dtilde(G_n,G_0)$ and $E_n/\Dtilde(G_n,G_0)$ as linear combinations of elements of the following set:
\begin{align}
\label{eq:set_H}
\mathcal{H}:&=\Big\{X^{\ell_1}\dfrac{\partial^{\ell_2} f_{\mathcal{D}}}{\partial h_1^{\ell_2}}(Y| \aj X+\bj,\vj)f_{\mathcal{L}}(X| \cj,\gj):j\in[\ktilde],~0\leq\ell_1+\ell_2\leq 2\trj\Big\}\nonumber\\
&\cup\Big\{X^{\alpha_3}\dfrac{\partial^{\tau_1}f_{\mathcal{L}}}{\partial c^{\tau_1}}(X| \cj,\gj)\dfrac{\partial^{\alpha_3+\tau_2}f_{\mathcal{D}}}{\partial h_1^{\alpha_3+\tau_2}}(Y| \aj X+\bj,\vj):\ktilde+1\leq j\leq k_0,\ 0\leq\alpha_3\leq\trj,\nonumber\\ 
&\hspace{8cm}0\leq\tau_1+\tau_2\leq 2(\trj-\alpha_3)\Big\}.
\end{align}
For any $(j,\ell_1,\ell_2)\in\mathcal{Q}:=\{(j,\ell_1,\ell_2)\in\mathbb{N}^3:j\in[\ktilde],~0\leq\ell_1+\ell_2\leq 2\trj\}$, let $Q^{n}_{\ell_1,\ell_2}(j)$ be the coefficient of 
\begin{align*}
    X^{\ell_1}\dfrac{\partial^{\ell_2} f_{\mathcal{D}}}{\partial h_1^{\ell_2}}(Y| \aj X+\bj,\vj)f_{\mathcal{L}}(X| \cj,\gj)
\end{align*}
in the representations of $A_n$, $B_n$ and $E_n$. It follows from equations~\eqref{eq:An_new_form} and \eqref{eq:Bn_new_form} that $Q^{n}_{\ell_1,\ell_2}(j)$ is given by
\begin{align*}
    Q^{n}_{\ell_1,\ell_2}(j)=\begin{cases}
        \sum_{\alpha\in\mathcal{I}_{\ell_1,\ell_2}}\sum_{i\in\mathcal{A}_j}\dfrac{\pin\cdot t_{\ell_1-\alpha_3,\alpha_1+2\alpha_2}}{2^{\alpha_2+\alpha_5}\alpha!}(\dcijn)^{\alpha_1}(\dgijn)^{\alpha_2}(\daijn)^{\alpha_3}(\dbijn)^{\alpha_4}(\dvijn)^{\alpha_5}\\
        \hspace{8cm}(\ell_1,\ell_2)\neq (0,0),\\
        \textbf{}\\
        \sum_{i\in\mathcal{A}_j}\pin-\pizeroj,\hspace{5.5cm} (\ell_1,\ell_2)=(0,0).
    \end{cases}
\end{align*}
Meanwhile, we denote by $T^{n}_{\alpha_3,\tau_1,\tau_2}(j)$ the coefficient of 
\begin{align*}
    X^{\alpha_3}\dfrac{\partial^{\tau_1}f_{\mathcal{L}}}{\partial c^{\tau_1}}(X| \cj,\gj)\dfrac{\partial^{\alpha_3+\tau_2}f_{\mathcal{D}}}{\partial h_1^{\alpha_3+\tau_2}}(Y| \aj X+\bj,\vj),
\end{align*}
for all $(j,\alpha_3,\tau_1,\tau_2)\in\mathcal{T}:=\{(j,\alpha_3,\tau_1,\tau_2)\in\mathbb{N}^3:\ktilde+1\leq j\leq k_0,~0\leq\alpha_3\leq\trj,~0\leq\tau_1+\tau_2\leq 2(\trj-\alpha_3)\}$. 
Thus, $T^{n}_{\alpha_3,\tau_1,\tau_2}(j)$ is represented as
\begin{align*}
    T^{n}_{\alpha_3,\tau_1,\tau_2}(j)=\begin{cases}
        \sum_{\substack{ \alpha_1+2\alpha_2=\tau_1, \\ \alpha_4+2\alpha_5=\tau_2}}\sum_{i\in\mathcal{A}_j}\dfrac{\pin}{2^{\alpha_2+\alpha_5}\alpha!}(\dcijn)^{\alpha_1}(\dgijn)^{\alpha_2}(\daijn)^{\alpha_3}(\dbijn)^{\alpha_4}(\dvijn)^{\alpha_5}\\ 
        \hspace{6.4cm}(\alpha_3,\tau_1,\tau_2)\neq (0,0,0),\\
        \textbf{}\\ 
        \sum_{i\in\mathcal{A}_j}\pin-\pizeroj,\hspace{4cm} (\alpha_3,\tau_1,\tau_2)=(0,0,0).
    \end{cases}
\end{align*}
\textbf{Step 2 - Proof of non-vanishing coefficients by contradiction:} Assume by contrary that all the coefficients of elements in the set $\mathcal{H}$ in the representations of $A_n/\Dtilde(G_n,G_0)$, $B_n/\Dtilde(G_n,G_0)$ and $E_n/\Dtilde(G_n,G_0)$ vanish when $n$ tends to infinity. It is worth noting that for $(\ell_1,\ell_2)\neq (0,0)$, we have $\mathcal{I}_{\ell_1+1,\ell_2}\subseteq\mathcal{I}_{\ell_1,\ell_2}$ and
\begin{align*}
    \mathcal{J}_{\ell_1,\ell_2}:=\mathcal{I}_{\ell_1,\ell_2}\setminus\mathcal{I}_{\ell_1+1,\ell_2}=\Big\{(\alpha_1,\ldots,\alpha_5)\in\mathbb{N}^5:&~\alpha_1+2\alpha_2+\alpha_3=\ell_1, ~\alpha_3+\alpha_4+2\alpha_5=\ell_2,\\
    &\hspace{1cm} 1\leq \alpha_1+\alpha_2+\ldots+\alpha_5\leq\trj\Big\}.
\end{align*}
Since $Q^{n}_{\ell_1,\ell_2}(j)/\Dtilde(G_n,G_0)\to 0$ for all tuples $(j,\ell_1,\ell_2)\in\mathcal{Q}$, we achieve that
\begin{align*}
    &\dfrac{S^{n}_{\ell_1,\ell_2}(j)}{\Dtilde(G_n,G_0)}:=\dfrac{Q^{n}_{\ell_1,\ell_2}(j)-Q_{\ell_1+1,\ell_2}(j)}{\Dtilde(G_n,G_0)}\\
    &=\dfrac{\sum_{\alpha\in\mathcal{J}_{\ell_1,\ell_2}}\sum_{i\in\mathcal{A}_j}\dfrac{\pin\cdot t_{\ell_1-\alpha_3,\alpha_1+2\alpha_2}}{2^{\alpha_2+\alpha_5}\alpha!}(\dcijn)^{\alpha_1}(\dgijn)^{\alpha_2}(\daijn)^{\alpha_3}(\dbijn)^{\alpha_4}(\dvijn)^{\alpha_5}}{\Dtilde(G_n,G_0)}\\
    &=\dfrac{\sum_{\alpha\in\mathcal{J}_{\ell_1,\ell_2}}\sum_{i\in\mathcal{A}_j}\dfrac{\pin}{(\gj)^{\alpha_1+2\alpha_2}~2^{\alpha_2+\alpha_5}\alpha!}(\dcijn)^{\alpha_1}(\dgijn)^{\alpha_2}(\daijn)^{\alpha_3}(\dbijn)^{\alpha_4}(\dvijn)^{\alpha_5}}{\Dtilde(G_n,G_0)}\\
    &\to 0,
\end{align*}
where the third inequality follows from the fact that $t_{\ell_1-\alpha_3,\alpha_1+2\alpha_2}=t_{\alpha_1+2\alpha_2,\alpha_1+2\alpha_2}=(\gj)^{-(\alpha_1+2\alpha_2)}$.
Additionally, we also let $S_{0,0}(j):=Q_{0,0}(j)$ for all $j\in[\ktilde]$.

By assumption, $|S_{0,0}(j)|/\Dtilde(G_n,G_0)\to 0$ for all $j\in[\ktilde]$ and $|T_{0,0,0}(j)|/\Dtilde(G_n,G_0)\to0$ for all $\ktilde+1\leq j\leq k_0$ as $n\to\infty$. By taking the summation of all such terms, we get that
\begin{align}
    \label{eq:En_vanish}
    \frac{1}{\Dtilde(G_n,G_0)}\cdot\sum_{j=1}^{k_0}\left|\sum_{i\in\mathcal{A}_j}\pin-\pizeroj\right|\to 0.
\end{align}
Next, we consider indices $j\in[k_0]:|\mathcal{A}_j|=1$, i.e. those in the formulation of $B_n$. For $j\in[\ktilde]$, since $|S^{n}_{\ell_1,\ell_2}(j)|/\Dtilde(G_n,G_0)\to 0$ for all $(\ell_1,\ell_2)\in\{(1,0),(0,1),(1,1),(2,0),(0,2)\}$, we get that 
\begin{align}
    \label{eq:Bn_vanish_1}
    \dfrac{1}{\Dtilde(G_n,G_0)}\cdot \sum_{\substack{j:|\mathcal{A}_j|=1\\ j\in[\ktilde]}}\sum_{i\in\mathcal{A}_j}\pin\Big(|\dcijn|+|\dgijn|+|\daijn|+|\dbijn|+|\dvijn|\Big)\to 0.
\end{align}
Moreover, for $\ktilde+1\leq j\leq k_0$, as $|T^{n}_{\alpha_3,\tau_1,\tau_2}(j)|/\Dtilde(G_n,G_0)\to 0$ for all $(\alpha_3,\tau_1,\tau_2)\in\{(0,1,0),(0,2,0),(1,0,0),(0,0,1),(0,0,2)\}$, we deduce that
\begin{align}
     \label{eq:Bn_vanish_2}
    \dfrac{1}{\Dtilde(G_n,G_0)}\cdot \sum_{\substack{j:|\mathcal{A}_j|=1\\ \ktilde+1\leq j\leq k_0}}\sum_{i\in\mathcal{A}_j}\pin\Big(|\dcijn|+|\dgijn|+|\daijn|+|\dbijn|+|\dvijn|\Big)\to 0.
\end{align}
Let us denote 
\begin{align*}
    K^n_{ij}(\kappa_1,\kappa_2,\kappa_3,\kappa_4,\kappa_5):=|\dcijn|^{\kappa_1}+|\dgijn|^{\kappa_2}+|\daijn|^{\kappa_3}+|\dbijn|^{\kappa_4}+|\dvijn|^{\kappa_5}.
\end{align*}
Then, equations~\eqref{eq:Bn_vanish_1} and \eqref{eq:Bn_vanish_2} indicates that 
\begin{align}
    \label{eq:Bn_vanish}
    \dfrac{1}{\Dtilde(G_n,G_0)}\cdot \sum_{j:|\mathcal{A}_j|=1}\sum_{i\in\mathcal{A}_j}\pin K^n_{ij}(1,1,1,1,1)\to 0.
\end{align}
Additionally, since $|T_{2,0,0}(j)|/\Dtilde(G_n,G_0)\to 0$ for all $\ktilde+1\leq j\leq k_0$, we have that
\begin{align}
    \label{eq:an2_vanish}
    \dfrac{1}{\Dtilde(G_n,G_0)}\cdot \sum_{j:|\mathcal{A}_j|>1}\sum_{i\in\mathcal{A}_j}\pin|\daijn|^2\to0.
\end{align}
Putting the results in equations~\eqref{eq:En_vanish}, \eqref{eq:Bn_vanish} and \eqref{eq:an2_vanish} together with the formulation of $\Dtilde(G_n,G_0)$ in equation~\eqref{eq_D_tilde_formulation}, we obtain that
\begin{align}
    \dfrac{1}{\Dtilde(G_n,G_0)}&\Bigg[\sum_{\substack{j:|\mathcal{A}_j|>1\\ j\in[\ktilde]}}\sum_{i\in\mathcal{A}_j}\pin K^n_{ij}\Big(\trj,\frac{\trj}{2},\frac{\trj}{2},\trj,\frac{\trj}{2}\Big)\nonumber\\
    \label{eq:contradict_result}
    &+\sum_{\substack{j:|\mathcal{A}_j|>1\\ \ktilde+1\leq j\leq k_0}}\sum_{i\in\mathcal{A}_j}\pin K^n_{-3,ij}\Big(\trj,\frac{\trj}{2},\trj,\frac{\trj}{2}\Big)\Bigg]\to 1,
\end{align}
where $K^n_{-3,ij}(\kappa_1,\kappa_2,\kappa_3,\kappa_4,\kappa_5):=|\dcijn|^{\kappa_1}+|\dgijn|^{\kappa_2}+|\dbijn|^{\kappa_4}+|\dvijn|^{\kappa_5}$.
Now, we will divide our arguments into two main scenarios based on the above limit:

\textbf{Case 1}: $\dfrac{\sum_{\substack{j:|\mathcal{A}_j|>1\\ j\in[\ktilde]}}\sum_{i\in\mathcal{A}_j}\pin K^n_{ij}\Big(\trj,\frac{\trj}{2},\frac{\trj}{2},\trj,\frac{\trj}{2}\Big)}{\Dtilde(G_n,G_0)}\not\to 0$.

This assumption indicates that we can find an index $j^*\in[\ktilde]:|\mathcal{A}_{j^*}|>1$ such that 
\begin{align*}
    \dfrac{1}{\Dtilde(G_n,G_0)}\cdot\sum_{i\in\mathcal{A}_{j^*}}\pin K^n_{ij^*}\Big(\trs,\frac{\trs}{2},\frac{\trs}{2},\trs,\frac{\trs}{2}\Big)\not\to 0.
\end{align*}
WLOG, we may assume that $j^*=1$ throughout this case. Recall that $S^{n}_{\ell_1,\ell_2}(1)/\Dtilde(G_n,G_0)\to 0$ for all pairs $(\ell_1,\ell_2)$ such that $0\leq \ell_1+\ell_2\leq 2\trone$. Combine this result with the assumption of case 1, we obtain 
\begin{align*}
    \dfrac{S^{n}_{\ell_1,\ell_2}(1)}{D_1(G_n,G_0)}=\dfrac{S^{n}_{\ell_1,\ell_2}(1)}{\Dtilde(G_n,G_0)}\cdot\dfrac{\Dtilde(G_n,G_0)}{D_1(G_n,G_0)}\to0,
\end{align*}
where $D_1(G_n,G_0):=\sum_{i\in\mathcal{A}_1}\pin K^n_{i1}\Big(\trone,\frac{\trone}{2},\frac{\trone}{2},\trone,\frac{\trone}{2}\Big)$. 
By expanding the formulations of $S^{n}_{\ell_1,\ell_2}(1)$ and $D_1(G_n,G_0)$, we have that
\begin{align}
    \label{eq:polynomial_equation}
    \dfrac{\sum_{i\in\mathcal{A}_1}\sum_{\alpha\in\mathcal{J}_{\ell_1,\ell_2}}\dfrac{\pin}{(\Gamma^0_1)^{\alpha_1+2\alpha_2}~2^{\alpha_2+\alpha_5}\alpha!}(\dcione)^{\alpha_1}(\dgione)^{\alpha_2}(\daione)^{\alpha_3}(\dbione)^{\alpha_4}(\dvione)^{\alpha_5}}{\sum_{i\in\mathcal{A}_1}\pin\Big[|\dcione|^{\trone}+|\dgione|^{\frac{\trone}{2}}+|\daione|^{\frac{\trone}{2}}+|\dbione|^{\trone}+|\dvione|^{\frac{\trone}{2}}\Big]}\to0.
\end{align}

Next, we define $\overline{M}_n:=\max\{|\dcione|,|\dgione|^{1/2},|\daione|^{1/2},|\dbione|,|\dvione|^{1/2}:i\in\mathcal{A}_1\}$ and $\overline{\pi}_n:=\max_{i\in\mathcal{A}_1}\pin$. For any $i\in\mathcal{A}_1$, since the sequence $(\pin/\overline{\pi}_n)_{i\in\mathcal{A}_1}$ is bounded, we can substitute it with its subsequence that admits a non-negative limit $p_i^2=\lim_{n\to\infty}\pin/\overline{\pi}_n$. 

Additionally, we define $(\dcione)/(\gj\cdot\overline{M}_n)\to q_{1i}$, $(\dgione)/[2(\gj)^2\cdot\overline{M}^2_n]\to q_{2i}$, $(\daione)/\overline{M}^2_n\to q_{3i}$, $(\dbione)/\overline{M}_n\to q_{4i}$ and $(\dvione)/2\overline{M}^2_n\to q_{5i}$. It can be seen from the formulation of $\mathcal{O}_{k,\beta}(\Theta)$ that $\pin\geq \delta$, therefore, $p_i$'s will not vanish and at least one of them is equal to 1. Similarly, at least one of the limits $q_{1i},q_{2i},\ldots,q_{5i}$ will be equal to either $1$ or $-1$.

Since
\begin{align*}
    \dfrac{\sum_{i\in\mathcal{A}_1}\pin\Big[|\dcione|^{\trone}+|\dgione|^{\frac{\trone}{2}}+|\daione|^{\frac{\trone}{2}}+|\dbione|^{\trone}+|\dvione|^{\frac{\trone}{2}}\Big]}{\overline{\pi}_n\overline{M}^{\ell_1+\ell_2}_n}\not\to0,
\end{align*}
for all pairs $(\ell_1,\ell_2)$ such that $\ell_1+\ell_2\in[\trone]$, we can divide both the numerator and the denominator in equation~\eqref{eq:polynomial_equation} by $\overline{\pi}_n\overline{M}^{\ell_1+\ell_2}_n$, and then let $n\to\infty$ to achieve the following system of polynomial equations:
\begin{align*}
    \sum_{i\in\mathcal{A}_1}\sum_{\alpha\in\mathcal{J}_{\ell_1,\ell_2}}\dfrac{p_i^2~q_{1i}^{\alpha_1}~q_{2i}^{\alpha_2}~q_{3i}^{\alpha_3}~q_{4i}^{\alpha_4}~q_{5i}^{\alpha_5}}{\alpha_1!~\alpha_2!~\alpha_3!~\alpha_4!~\alpha_5!}=0,
\end{align*}
for all pairs $(\ell_1,\ell_2)$ such that $0\leq\ell_1+\ell_2\leq \trone$. Nevertheless, according to the definition of $\trone$, the above system cannot admit any non-trivial solutions, which is a contradiction. Thus, case 1 does not hold.

\textbf{Case 2}: $\dfrac{1}{\Dtilde(G_n,G_0)}\cdot\sum_{\substack{j:|\mathcal{A}_j|>1\\ \ktilde+1\leq j\leq k_0}}\sum_{i\in\mathcal{A}_j}\pin K^n_{-3,ij}\Big(\trj,\frac{\trj}{2},\trj,\frac{\trj}{2}\Big)\not\to 0$.


This assumption implies that there exists an index $\ktilde+1\leq j^*\leq k_0:|\mathcal{A}_{j^*}|>1$ such that 
\begin{align}
    \label{eq:mimic_i_tilde}
    \dfrac{1}{\Dtilde(G_n,G_0)}\cdot\sum_{i\in\mathcal{A}_{j^*}}\pin K^n_{-3,ij^*}\Big(\trs,\frac{\trs}{2},\trs,\frac{\trs}{2}\Big)\not\to 0.
\end{align}
By applying similar arguments for equation~\eqref{eq:mimic_i_bar} in the proof of Theorem~\ref{theorem:type_1_setting} to equation~\eqref{eq:mimic_i_tilde}, we are able to point out that equation~\eqref{eq:mimic_i_tilde} cannot happen, which is a contradiction. As a result, case 2 cannot happen either.

Collect the results of the above two scenarios, we realize that the limit in equation~\eqref{eq:contradict_result} does not hold true, which is a contradiction. As a consequence, not all the coefficients of elements in the set $\mathcal{H}$, defined in equation~\eqref{eq:set_H}, in the representations of $A_n/\Dtilde(G_n,G_0)$, $B_n/\Dtilde(G_n,G_0)$ and $E_n/\Dtilde(G_n,G_0)$ go to zero as $n\to\infty$. 

\textbf{Step 3 - Application of Fatou's lemma:} Next, we denote by $m_n$ the maximum of the absolute values of those coefficients, which means that
\begin{align*}
    m_n:=\max\left\{\max_{\substack{(j,\ell_1,\ell_2)\in\mathcal{Q}}}\dfrac{|Q^{n}_{\ell_1,\ell_2}(j)|}{\Dtilde(G_n,G_0)},~\max_{\substack{ (j,\alpha_3,\tau_1,\tau_2)\in\mathcal{T}}}\dfrac{|T^{n}_{\alpha_3,\tau_1,\tau_2}(j)|}{\Dtilde(G_n,G_0)}\right\}.
\end{align*}
In addition, let us define $Q^{n}_{\ell_1,\ell_2}(j)/m_n\to\zeta_{\ell_1,\ell_2}(j)$ for $(j,\ell_1,\ell_2)\in\mathcal{Q}$ and $T^{n}_{\alpha_3,\tau_1,\tau_2}(j)\to\xi_{\alpha_3,\tau_1,\tau_2}(j)$ for $(j,\alpha_3,\tau_1,\tau_2)\in\mathcal{T}$ as $n\to\infty$. As not all the coefficients of elements of $\mathcal{H}$ in the representations of $A_n/\Dtilde(G_n,G_0)$, $B_n/\Dtilde(G_n,G_0)$ and $E_n/\Dtilde(G_n,G_0)$ vanish as $n\to\infty$, at least one among $\zeta_{\ell_1,\ell_2}(j)$ and $\xi_{\alpha_3,\tau_1,\tau_2}(j')$ is different from zero and $m_n\not\to 0$. By invoking the Fatou's lemma, we get that
\begin{align*}
    0=\lim_{n\to\infty}\dfrac{1}{m_n}\cdot\dfrac{2V(p_{G_n},p_{G})}{\Dtilde(G_n,G_0)}\geq\int\liminf_{n\to\infty}\dfrac{1}{m_n}\cdot\dfrac{|p_{G_n}(X,Y)-p_{G}(X,Y)|}{\Dtilde(G_n,G_0)}~\dint(X,Y)\geq 0.
\end{align*}
Furthermore, we have that
\begin{align*}
    \dfrac{1}{m_n}\cdot&\dfrac{p_{G_n}(X,Y)-p_{G}(X,Y)}{\Dtilde(G_n,G_0)}\\
    &\to \sum_{(j,\ell_1,\ell_2)\in\mathcal{Q}}\zeta_{\ell_1,\ell_2}(j)X^{\ell_1}\cdot\dfrac{\partial^{\ell_2} f_{\mathcal{D}}}{\partial h_1^{\ell_2}}(Y| \aj X+\bj,\vj)\cdot f_{\mathcal{L}}(X| \cj,\gj)\\
    &+\sum_{(j,\alpha_3,\tau_1,\tau_2)\in\mathcal{T}}\xi_{\alpha_3,\tau_1,\tau_2}(j)X^{\alpha_3}\cdot\dfrac{\partial^{\tau_1}f_{\mathcal{L}}}{\partial c^{\tau_1}}(X| \cj,\gj)\cdot\dfrac{\partial^{\alpha_3+\tau_2}f_{\mathcal{D}}}{\partial h_1^{\alpha_3+\tau_2}}(Y| \aj X+\bj,\vj).
\end{align*}
Consequently, we achieve that
\begin{align*}
    &\sum_{(j,\ell_1,\ell_2)\in\mathcal{Q}}\zeta_{\ell_1,\ell_2}(j)X^{\ell_1}\cdot\dfrac{\partial^{\ell_2} f_{\mathcal{D}}}{\partial h_1^{\ell_2}}(Y| \aj X+\bj,\vj)\cdot f_{\mathcal{L}}(X| \cj,\gj)\\
    &+\sum_{(j,\alpha_3,\tau_1,\tau_2)\in\mathcal{T}}\xi_{\alpha_3,\tau_1,\tau_2}(j)X^{\alpha_3}\dfrac{\partial^{\tau_1}f_{\mathcal{L}}}{\partial c^{\tau_1}}(X| \cj,\gj)\dfrac{\partial^{\alpha_3+\tau_2}f_{\mathcal{D}}}{\partial h_1^{\alpha_3+\tau_2}}(Y| \aj X+\bj,\vj)=0,
\end{align*}
for almost surely $(X,Y)$. Since elements of the set $\mathcal{H}$ defined in equation~\eqref{eq:set_H} are linearly independent (proof of this claim is deferred to the end of this proof), the above equation indicates that $\zeta_{j,\ell_1,\ell_2}(j)=\xi_{\alpha_3,\tau_1,\tau_2}(j')=0$ for all $(j,\ell_1,\ell_2)\in\mathcal{Q}$ and $(j',\alpha_3,\tau_1,\tau_2)\in\mathcal{T}$, which contradicts the fact that at least one among $\zeta_{j,\ell_1,\ell_2}(j)$, $\xi_{\alpha_3,\tau_1,\tau_2}(j')$ is different from zero. Hence, we reach the conclusion in equation~\eqref{eq:local_version_2}.

\textbf{Linear independence of elements in $\mathcal{H}$}: For completion, we will show that elements of the set $\mathcal{F}$ defined in equation~\eqref{eq:set_H} are linearly independent by definition. In particular, assume that there exist real numbers $\zeta_{\ell_1,\ell_2}(j)$ and $\xi_{\alpha_3,\tau_1,\tau_2}(j')$, where $(j,\ell_1,\ell_2)\in\mathcal{Q}$ and $(j',\alpha_3,\tau_1,\tau_2)\in\mathcal{T}$, such that the following equation holds for almost surely $(X,Y)$:
\begin{align*}
    &\sum_{(j,\ell_1,\ell_2)\in\mathcal{Q}}\zeta_{\ell_1,\ell_2}(j)X^{\ell_1}\cdot\dfrac{\partial^{\ell_2} f_{\mathcal{D}}}{\partial h_1^{\ell_2}}(Y| \aj X+\bj,\vj)\cdot f_{\mathcal{L}}(X| \cj,\gj)\\
    &+\sum_{(j',\alpha_3,\tau_1,\tau_2)\in\mathcal{T}}\xi_{\alpha_3,\tau_1,\tau_2}(j')X^{\alpha_3}\dfrac{\partial^{\tau_1}f_{\mathcal{L}}}{\partial c^{\tau_1}}(X| \cjp,\gjp)\dfrac{\partial^{\alpha_3+\tau_2}f_{\mathcal{D}}}{\partial h_1^{\alpha_3+\tau_2}}(Y| \ajp X+\bjp,\vjp)=0,
\end{align*}
Now, we rewrite the above equation as follows:
\begin{align}
\label{eq:independent_2}   
&\sum_{j=1}^{k_0}\sum_{\omega=0}^{2\trj}\Bigg[\sum_{\alpha_3+\tau_2=\omega}\sum_{\tau_1=0}^{2(\trj-\alpha_3)-\tau_2}\xi_{\alpha_3,\tau_1,\tau_2}(j)X^{\alpha_3}\cdot \dfrac{\partial^{\tau_1}f_{\mathcal{L}}}{\partial c^{\tau_1}}(X| \cj,\gj)\cdot\mathbf{1}_{\{\ktilde+1\leq j\leq k_0\}} \nonumber\\
&+\sum_{\ell_1=0}^{2\trj-\omega}\zeta_{\ell_1,\omega}(j)X^{\ell_1}\cdot f_{\mathcal{L}}(X| \cj,\gj)\cdot\mathbf{1}_{\{j\in[\ktilde]\}}\Bigg]\dfrac{\partial^{\omega}f_{\mathcal{D}}}{\partial h_1^{\omega}}(Y| \aj X+\bj,\vj)=0,
\end{align}
for almost surely $(X,Y)$. As $(\aj,\bj,\vj)$ for $j\in[k_0]$ are $k_0$ distinct tuples, we deduce that $((\aj)^{\top}X+\bj,\vj)$ for $j\in[k_0]$ are also $k_0$ distinct tuples for almost surely $X$. Thus, for almost surely $X$, one has $\dfrac{\partial^{\omega}f_{\mathcal{D}}}{\partial h_1^{\omega}}(Y| (\aj)^{\top}X+\bj,\vj)$ for $j\in[k_0]$ and $0\leq\omega\leq 2\trj$ are linearly independent with respect to $Y$. Given that result, the equation~\eqref{eq:independent_2} indicates that for almost surely $X$,
\begin{align*}
    &\sum_{\alpha_3+\tau_2=\omega}\sum_{\tau_1=0}^{2(\trj-\alpha_3)-\tau_2}\xi_{\alpha_3,\tau_1,\tau_2}(j)X^{\alpha_3}\cdot \dfrac{\partial^{\tau_1}f_{\mathcal{L}}}{\partial c^{\tau_1}}(X| \cj,\gj)\cdot\mathbf{1}_{\{\ktilde+1\leq j\leq k_0\}}\\
    &\hspace{4cm}+\sum_{\ell_1=0}^{2\trj-\omega}\zeta_{\ell_1,\omega}(j)X^{\ell_1}\cdot f_{\mathcal{L}}(X| \cj,\gj)\cdot\mathbf{1}_{\{j\in[\ktilde]\}}=0.
\end{align*}
for all $j\in[k_0]$ and $0\leq\omega\leq 2\trj$. This equation is equivalent to
\begin{align}
    \label{eq:first_equation}
    \sum_{\ell_1=0}^{2\trj-\omega}\zeta_{\ell_1,\omega}(j)X^{\ell_1}\cdot f_{\mathcal{L}}(X| \cj,\gj)&=0,\\
    \label{eq:second_equation}
    \sum_{\alpha_3+\tau_2=\omega'}\sum_{\tau_1=0}^{2(\trjp-\alpha_3)-\tau_2}\xi_{\alpha_3,\tau_1,\tau_2}(j')X^{\alpha_3}\cdot \dfrac{\partial^{\tau_1}f_{\mathcal{L}}}{\partial c^{\tau_1}}(X| \cjp,\gjp)&=0,
\end{align}
for all $j\in[\ktilde]$, $0\leq \omega\leq 2\trj$ and $\ktilde+1\leq j^{'}\leq k_0$, $0\leq \omega\leq 2\trjp$. We can treat the left hand side of equation~\eqref{eq:first_equation} as a polynomial of the random vector $X\in\mathcal{X}$, which is a compact set in $\mathbb{R}$. Meanwhile, the left hand side of equation~\eqref{eq:second_equation} can be viewed as another polynomial of $X$ and $X-\cjp$, where $\cjp\neq 0$. As a result, the above equations hold when $\zeta_{\ell_1,\omega}(j)=0$ for all $j\in[\ktilde]$, $0\leq\omega\leq 2\trj$, $0\leq\ell_1\leq 2\trj-\omega$, and $\xi_{\alpha_3,\tau_1,\tau_2}(j{'})=0$ for all $\ktilde+1\leq j'\leq k_0$, $0\leq\omega'\leq 2\trjp$, $\alpha_3+\tau_2=\omega'$ and $0\leq\tau_1\leq 2(\trj-\alpha_3)-\tau_2$. This result is equivalent to $\zeta_{\ell_1,\ell_2}(j)=0$, for all $(j,\ell_1,\ell_2)\in\mathcal{Q}$ and $\xi_{\alpha_3,\tau_1,\tau_2}(j')=0$ for all $(j',\alpha_3,\tau_1,\tau_2)\in\mathcal{T}$.

Hence, the elements of $\mathcal{H}$ are linearly independent, which completes the proof.

\section{PROOF OF REMAINING RESULTS}\label{sec_appendix_Auxiliary_Results}
In this appendix, we provide proofs for Proposition~\ref{prop_identifiability}, Proposition~\ref{prop_density_rate} and Lemma~\ref{lemma_r_tilde} in that order.
\subsection{Proof of Proposition \ref{prop_identifiability}}
\label{appendix_identifiability}
For any two mixing measures $G=\sum_{i=1}^{k}\pi_i\delta_{(c_i,\Gamma_i,a_i,b_i,\nu_i)}$ and $G'=\sum_{i=1}^{k'}\pi'_i\delta_{(c'_i,\Gamma'_i,a'_i,b'_i,\nu'_i)}$, we assume that $p_{G}(X,Y)=p_{G'}(X,Y)$ holds true for almost surely $(X,Y)\in\mathcal{X}\times\mathcal{Y}$, or equivalently,
\begin{align}
    \label{eq:identifiability_equation}
    \sum_{i=1}^{k}\pi_if_{\mathcal{L}}(X|c_i,\Gamma_i)f_{\mathcal{D}}(Y|(a_i)^{\top}X+b_i,\nu_i)=\sum_{i=1}^{k'}\pi'_if_{\mathcal{L}}(X|c'_i,\Gamma'_i)f_{\mathcal{D}}(Y|(a'_i)^{\top}X+b'_i,\nu'_i).
\end{align}
Recall that if $Y|X\sim\mathcal{N}_{1}(a^{\top}X+b,\nu)$ and $X\sim\mathcal{N}_{d}(c,\Gamma)$, then
\begin{align*}
    \begin{pmatrix}
        X \\ Y
    \end{pmatrix}\sim
    \mathcal{N}_{d+1}\left(\begin{pmatrix}
        c \\ a^{\top}c+b
    \end{pmatrix},
    \begin{pmatrix}
        \Gamma & \Gamma a\\
        a^{\top}\Gamma & a^{\top}\Gamma a+\nu
    \end{pmatrix}\right).
\end{align*}
Let us denote 
\begin{align*}
    \psi_i:=\begin{pmatrix}
    c_i \\ (a_i)^{\top}c_i+b_i
\end{pmatrix}, \quad \Sigma_i:=\begin{pmatrix}
    \Gamma_i & \Gamma_i a_i\\
        (a_i)^{\top}\Gamma_i & (a_i)^{\top}\Gamma_i a_i+\nu_i
\end{pmatrix}, \\
\psi'_i:=\begin{pmatrix}
    c'_i \\ (a'_i)^{\top}c'_i+b'_i
\end{pmatrix}, \quad \Sigma'_i:=\begin{pmatrix}
    \Gamma'_i & \Gamma'_i a'_i\\
        (a'_i)^{\top}\Gamma'_i & (a_i)^{\top}\Gamma'_i a'_i+\nu'_i
\end{pmatrix}.
\end{align*}
Then, equation~\eqref{eq:identifiability_equation} can be rewritten as
\begin{align}
    \label{eq:new_id_equation}
    \sum_{i=1}^{k}\pi_if(X,Y|\psi_i,\Sigma_i) = \sum_{i=1}^{k'}\pi'_if(X,Y|\psi'_i,\Sigma'_i),    
\end{align}
for almost surely $(X,Y)$, where $f$ belongs to the family of $(d+1)$-dimensional Gaussian density functions. Since the location-scale Gaussian mixtures are identifiable, it follows from the above equation that $k=k'$ and $\{\pi_1,\pi_2,\ldots,\pi_k\}\equiv\{\pi'_1,\pi'_2,\ldots,\pi'_k\}$. WLOG, we may assume that $\pi_i=\pi'_i$ for any $i\in[k]$. 

Subsequently, we construct a partition of the set $[k]$, denoted by $P_1,P_2,\ldots,P_m$ that satisfies the following properties: 
\begin{itemize}
    \item[(i)] $\pi_{i}=\pi'_{i}$ for any $i\in P_{\ell}$ and $\ell\in[m]$;
    \item[(ii)] $\pi_{i}\neq \pi'_{j}$ if $i$ and $j$ are not in the same set $P_{\ell}$ for any $\ell\in[m]$.
\end{itemize}
Given this partition, we represent equation~\eqref{eq:new_id_equation} as follows:
\begin{align*}
    \sum_{\ell=1}^{m}\sum_{i\in P_{\ell}}\pi_{i}f(X,Y|\psi_i,\Sigma_i)=\sum_{\ell=1}^{m}\sum_{i\in P_{\ell}}\pi'_{i}f(X,Y|\psi'_i,\Sigma'_i),
\end{align*}
for almost surely $(X,Y)$. Consequently, for each $\ell\in[m]$, we obtain that
\begin{align*}
    \left\{(\psi_i,\Sigma_i):i\in P_{\ell}\right\}\equiv\left\{(\psi'_i,\Sigma'_i):i\in P_{\ell}\right\}.
\end{align*}
WLOG, we may assume that $(\psi_i,\Sigma_i)=(\psi'_i,\Sigma'_i)$ for any $i\in P_{\ell}$. Given this result, by some simple algebraic derivations, we achieve that $(c_i,\Gamma_i,a_i,b_i,\nu_i)=(c'_i,\Gamma'_i,a'_i,b'_i,\nu'_i)$ for any $i\in P_{\ell}$ and $\ell\in[m]$. As a result, it follows that
\begin{align*}
    G=\sum_{\ell=1}^{m}\sum_{i\in P_{\ell}}\pi_i\delta_{(c_i,\Gamma_i,a_i,b_i,\nu_i)}=\sum_{\ell=1}^{m}\sum_{i\in P_{\ell}}\pi'_i\delta_{(c'_i,\Gamma'_i,a'_i,b'_i,\nu'_i)}=G'.
\end{align*}
Hence, the proof is completed.
\subsection{Proof of Proposition \ref{prop_density_rate}}
\label{appendix_density_rate}
Prior to presenting the proof of Proposition~\ref{prop_density_rate}, let us review fundamental background on density estimation for M-estimators, which is covered in \cite{Vandegeer-2000}. First of all, we define $\mathcal{P}_{k,\beta}(\Theta):=\{p_{G}(X,Y):G\in\mathcal{O}_{k,\beta}(\Theta)\}$ as the set of joint densities of all mixing measure in $\mathcal{O}_{k,\beta}(\Theta)$. In addition, we denote
\begin{align*}
    \mathcal{Q}_{k,\beta}(\Theta)&:=\{p_{(G+G_0)/2}(X,Y):G\in\mathcal{O}_{k,\beta}(\Theta)\},\\
    \mathcal{Q}_{k,\beta}^{1/2}(\Theta)&:=\{p_{(G+G_0)/2}^{1/2}(X,Y):G\in\mathcal{O}_{k,\beta}(\Theta)\}.
\end{align*}
Subsequently, for any $\delta>0$, the Hellinger ball centered around the density $p_{G_0}(X,Y)$ and intersected with the set $\mathcal{Q}_{k,\beta}^{1/2}(\Theta)$ is defined as
\begin{align*}
    \mathcal{Q}_{k,\beta}^{1/2}(\Theta,\delta):=\{g^{1/2}\in\mathcal{Q}_{k,\beta}^{1/2}(\Theta):h(g,p_{G_0})\leq\delta\}.
\end{align*}
Additionally, Geer et al. \cite{Vandegeer-2000} introduce the following quantity to capture the size of the above Hellinger ball:
\begin{align}
    \label{eq:JB}
    \mathcal{J}_B(\delta,\mathcal{Q}_{k,\beta}^{1/2}(\Theta)):=\int_{\delta^2/2^{13}}^{\delta} H_B^{1/2}\Big(u,\mathcal{Q}_{k,\beta}^{1/2}(\Theta,u),\|\cdot\|\Big)\dint u\vee \delta,
\end{align}
where $H_B^{1/2}\Big(u,\mathcal{Q}_{k,\beta}^{1/2}(\Theta,u),\|\cdot\|\Big)$ denotes the bracketing entropy of $\mathcal{Q}_{k,\beta}^{1/2}(\Theta,u)$ under the Euclidean distance, and $u\vee\delta:=\max\{u,\delta\}$. Given these notations, let us state the result regarding the joint density estimation rate presented in Theorem 7.4 in \cite{Vandegeer-2000}.
\begin{lemma}[Theorem 7.4, \cite{Vandegeer-2000}]
    \label{theorem_74}
    Take $\Psi(\delta)\geq \mathcal{J}_B(\delta,\mathcal{Q}_{k,\beta}^{1/2}(\Theta))$ such that $\Psi(\delta)/\delta^2$ is a non-increasing function of $\delta$. Then, for a universal constant $c$ and a sequence $(\delta_n)$ that satisfies $\sqrt{n}\delta_n^2\geq c\Psi(\delta_n)$, we obtain that
    \begin{align*}
        \mathbb{P}\Big(h(p_{\widehat{G}_n},p_{G_0})>\delta\Big)\leq c\exp\left(-\frac{n\delta^2}{c^2}\right),
    \end{align*}
    for any $\delta\geq\delta_n$.
\end{lemma}
Proof of Lemma~\ref{theorem_74} is provided in \cite{Vandegeer-2000}. Next, we introduce the upper bounds of the covering number (under the sup norm) $N(\varepsilon,\mathcal{P}_{k,\beta}(\Theta),\|\cdot\|_{\infty})$, and the bracketing entropy (under the Hellinger distance) $H_B(\varepsilon,\mathcal{P}_{k,\beta}(\Theta),h)$ of the metric space $\mathcal{P}_{k,\beta}(\Theta)$. For further detail about the definitions of these terms, readers are referred to \cite{Vandegeer-2000}. 
\begin{lemma}
    \label{lemma:bracketing_entropy_bounds}
    Given a bounded set $\Theta$, we have for any $\varepsilon\in[0,1/2]$ that
    \begin{itemize}
        \item[(i)] $\log N(\varepsilon,\mathcal{P}_{k,\beta}(\Theta),\|\cdot\|_{\infty})\lesssim \log(1/\varepsilon)$;
        \item[(ii)] $H_B(\varepsilon,\mathcal{P}_{k,\beta}(\Theta),h)\lesssim \log(1/\varepsilon)$. 
    \end{itemize}
\end{lemma}
Proof of Lemma~\ref{lemma:bracketing_entropy_bounds} is relegated to Appendix~\ref{appendix:lemma:bracketing_entropy_bounds}. Now, we already have all necessary ingredients to provide the proof for Proposition~\ref{prop_density_rate} in Appendix~\ref{appendix:main_proof}
\subsubsection{Proof of Proposition~\ref{prop_density_rate}}
\label{appendix:main_proof}
    Note that for any $u>0$, we have
    \begin{align*}
        H_B(u,\mathcal{Q}_{k,\beta}^{1/2}(\Theta),\|\cdot\|)\leq H_B(u,\mathcal{P}_{k,\beta}(\Theta),h)\leq \log(1/u),
    \end{align*}
    where the second inequality is induced by part (ii) of Lemma~\ref{lemma:bracketing_entropy_bounds}. Then, it follows from equation~\eqref{eq:JB} that
    \begin{align}
        \label{eq:JB_bound}
        \mathcal{J}_B(\delta,\mathcal{Q}_{k,\beta}^{1/2}(\Theta))
        \leq \int_{\delta^2/2^{13}}^{\delta}\log(1/u)\dint u\vee\delta.
    \end{align}
    By choosing $\Psi(\delta):=\delta\cdot[\log(1/\delta)]^{1/2}$, we get that $\Psi(\delta)/\delta^2$ is a non-increasing function of $\delta$ and $\Psi(\delta)\geq\mathcal{J}_B(\delta,\mathcal{Q}_{k,\beta}^{1/2}(\Theta))$ from equation~\eqref{eq:JB_bound}. Let $\delta_n:=\sqrt{\log(n)/n}$, we achieve that  $\sqrt{n}\delta_n^2\geq c\Psi(\delta_n)$ for some universal constant $c$. As a result, Lemma~\ref{theorem_74} gives us that
    \begin{align*}
        \mathbb{P}(h(p_{\widehat{G}_n},p_{G_0})>C_1\sqrt{\log(n)/n})\lesssim \exp(-C_2\log(n))=n^{-C_2},
    \end{align*}
    where $C_1$ and $C_2$ are some universal constants. Finally, since the Total Variation is upper bounded by the Hellinger distance, we obtain the desired conclusion.

\subsubsection{Proof of Lemma~\ref{lemma:bracketing_entropy_bounds}}
\label{appendix:lemma:bracketing_entropy_bounds}
    \textbf{Part (i).} Given some $\varepsilon>0$, since $\Theta$ is a compact set, we can find an $\varepsilon$-cover of $\Theta$, denoted by $\Theta_{\varepsilon}$. Additionally, let $\Delta_{\varepsilon}$ be an $\varepsilon$-cover of an $(k-1)$-dimensional simplex. Assume that $|\Theta_{\varepsilon}|=T$ and $|\Delta_{\varepsilon}|=S$. Note that $\Theta\subset\mathbb{R}^d\times\mathcal{S}^+_d\times\mathbb{R}^d\times\mathbb{R}\times\mathbb{R}_+$ is a subspace of $\mathbb{R}^{d^2+4d}$, then it can be checked that $T=\mathcal{O}(\varepsilon^{-(d^2+4d)k})$ and $S=\mathcal{O}(\varepsilon^{-(k-1)})$. Next, we define
\begin{align*}
    \mathcal{G}:=\left\{p_{G}\in\mathcal{P}_{k,\beta}(\Theta):(\pi_1,\pi_2,\ldots,\pi_{k})\in\Delta_{\varepsilon}, (c_i,\Gamma_i,a_i,b_i,\nu_i)\in\Theta_{\varepsilon}\right\}.
\end{align*}
Given some mixing measure $G=\sum_{i=1}^{k'}\pi_i\delta_{\theta_i}\in\mathcal{O}_{k,\beta}(\Theta)$ with $k'\leq k$ and $\theta_i:=(c_i,\Gamma_i,a_i,b_i,\nu_i)\in\Theta$, let us consider $\overline{G}=\sum_{i=1}^{k'}\pi_i\delta_{\widetilde{\theta}_i}$ where $\widetilde{\theta}_i:=(\widetilde{c}_i,\widetilde{\Gamma}_i,\widetilde{a}_i,\widetilde{b}_i,\widetilde{\nu}_i)\in\Theta_{\varepsilon}$ such that $\|\widetilde{\theta}_i-\theta_i\|\leq\varepsilon$ for any $i\in[k']$. In addition, we also take into account another mixing measure $\widetilde{G}=\sum_{i=1}^{k'}\widetilde{\pi}_i\delta_{\widetilde{\theta}_i}$ where $(\widetilde{\pi}_1,\widetilde{\pi}_2,\ldots,\widetilde{\pi}_{k'},0,\ldots,0)\in\Delta_{\varepsilon}$ such that $\|(\widetilde{\pi}_i)_{i=1}^{k'}-(\pi_i)_{i=1}^{k'}\|\leq\varepsilon$. From the definition of $\mathcal{G}$, we get that $p_{\widetilde{G}}\in\mathcal{G}$. Since $\|(\widetilde{\pi}_i)_{i=1}^{k'}-(\pi_i)_{i=1}^{k'}\|\leq\varepsilon$, we can deduce that
\begin{align*}
    \|p_{\overline{G}}-p_{\widetilde{G}}\|_{\infty}\leq \sum_{i=1}^{k'}|\widetilde{\pi}_i-\pi_i|\cdot\|f_{\mathcal{L}}(X|\widetilde{c}_i,\widetilde{\Gamma}_i)f_{\mathcal{D}}(Y|(\widetilde{a}_i)^{\top}X+\widetilde{b}_i,\widetilde{\nu}_i)\|_{\infty}\lesssim\varepsilon.
\end{align*}
Next, we consider
\begin{align*}
    \|p_{G}-p_{\overline{G}}\|_{\infty}\leq \sum_{i=1}^{k'}\pi_i\|F(\theta_i|X,Y)-F(\widetilde{\theta}_i|X,Y)\|_{\infty},
\end{align*}
where we denote $F(\theta|X,Y):=f_{\mathcal{L}}(X|c,\Gamma)f_{\mathcal{D}}(Y|a^{\top}X+b,\nu)$. As $F$ is twice differentiable with respect to $\theta$ and $\mathcal{X}$ is a bounded set, we achieve the following inequality:
\begin{align*}
    \sum_{i=1}^{k'}\pi_i\|F(\theta_i|X,Y)-F(\widetilde{\theta}_i|X,Y)\|_{\infty}\leq \sum_{i=1}^{k'}\pi\|\widetilde{\theta}_i-\theta_i\|\lesssim\varepsilon,
\end{align*}
which leads to $\|p_{G}-p_{\overline{G}}\|_{\infty}\leq\varepsilon$. As a consequence, by the triangle inequality, we have
\begin{align*}
    \|p_{G}-p_{\widetilde{G}}\|_{\infty}\leq \|p_{G}-p_{\overline{G}}\|_{\infty}+\|p_{\overline{G}}-p_{\widetilde{G}}\|_{\infty}\lesssim\varepsilon.
\end{align*}
Given this result, it follows that $\mathcal{G}$ is an $\varepsilon$-cover of $\mathcal{P}_{k,\beta}(\Theta)$, therefore,
\begin{align*}
    N\Big(\varepsilon,\mathcal{P}_{k,\beta}(\Theta),\|\cdot\|_{\infty}\Big)\leq |\mathcal{G}|=S\times T =\mathcal{O}(\varepsilon^{-(d^2+4d)k})\times\mathcal{O}(\varepsilon^{-(k-1)})=\mathcal{O}(\varepsilon^{-(d^2+4d+1)k+1}),
\end{align*}
which implies that $\log N\Big(\varepsilon,\mathcal{P}_{k,\beta}(\Theta),\|\cdot\|_{\infty}\Big)\lesssim \log (1/\varepsilon)$.

\textbf{Part (ii).} We begin with finding an upper bound for the density $f_{\mathcal{L}}(X|c,\Gamma)f_{\mathcal{D}}(Y|a^{\top}X+b,\nu)$. Since $\mathcal{X}$ and $\Theta$ are bounded sets, we can find positive constants $u,u_1,u_2,u_3,l_1,l_3$ such that $\|c\|\leq u$, $l_1\leq \lambda_{\min}(\Gamma)\leq \lambda_{\max}(\Gamma)\leq u_1$, $-u_2\leq a^{\top}X+b\leq u_2$ and $l_3\leq \nu\leq u_3$, where $\lambda_{\min}(\Gamma)$ and $\lambda_{\max}(\Gamma)$ are the smallest and the largest eigenvalues of $\Gamma$, respectively. Firstly, it is clear that
\begin{align*}
    f_{\mathcal{L}}(X|c,\Gamma)=\frac{1}{\sqrt{(2\pi)^{d}\det(\Gamma)}}\exp\Big(-\frac{1}{2}(x-c)^{\top}\Gamma^{-1}(x-c)\Big)\leq\frac{1}{(2\pi l_1)^{d/2}}.
\end{align*}
Additionally, note that 
\begin{align*}
    (X-c)^{\top}\Gamma^{-1}(x-c)\geq \lambda_{\min}(\Gamma^{-1})\|X-c\|^2=\frac{1}{\lambda_{\max}(\Gamma)}\|X-c\|^2.
\end{align*}
Moreover, for any $\|X\|\geq 2u$, by the Cauchy-Schwartz inequality, we get 
\begin{align*}
    4\|X-c\|^2-\|X\|^2=3\|X\|^2-8X^{\top}c + 4\|c\|^2\geq 3\|X\|^2-8\|X\|\cdot\|c\| + 4\|c\|^2\geq 0,
\end{align*}
which implies that $(X-c)^{\top}\Gamma^{-1}(x-c)\geq\frac{1}{4u_1}\|X\|^2$. As a result,
\begin{align*}
    f_{\mathcal{L}}(X|c,\Gamma)=\frac{1}{\sqrt{(2\pi)^{d}\det(\Gamma)}}\exp\Big(-\frac{1}{2}(x-c)^{\top}\Gamma^{-1}(x-c)\Big)\leq\frac{1}{(2\pi l_1)^{d/2}}\exp\left(-\frac{\|X\|^2}{8u_1}\right),
\end{align*}
for any $\|X\|\geq 2u$. Combine this result with the previous bound, we obtain that $f_{\mathcal{L}}(X|c,\Gamma)\leq G_1(X)$, where
\begin{align*}
    G_1(X):=\begin{cases}
       \dfrac{1}{(2\pi l_1)^{d/2}}\exp\left(-\dfrac{\|X\|^2}{8u_1}\right), \quad \|X\|\geq 2u,\\
        \textbf{}\\
       \dfrac{1}{(2\pi l_1)^{d/2}}, \hspace{2.5cm} \|X\|<2u.
    \end{cases}
\end{align*}
By arguing in a similar fashion, we also have $f_{\mathcal{D}}(Y|a^{\top}X+b,\nu)\leq G_2(X,Y)$ where
\begin{align*}
    G_2(X,Y):=\begin{cases}
        \dfrac{1}{\sqrt{2\pi l_3}}\exp\left(-\dfrac{Y^2}{8u_3}\right), \quad |Y|\geq 2u_2\\
        \textbf{}\\
        \dfrac{1}{\sqrt{2\pi l_3}},\hspace{2.4cm} |Y|<2u_2.
    \end{cases}
\end{align*}
Consequently, we achieve that $f_{\mathcal{L}}(X|c,\Gamma)f_{\mathcal{D}}(Y|a^{\top}X+b,\nu)\leq G(X,Y):=G_1(X)G_2(X,Y)$.

Next, given some $\eta>0$ that we will choose later, we consider an $\eta$-cover of $\mathcal{P}_{k,\beta}(\Theta)$ which is assumed to have $N$ elements denoted by $f_1,f_2,\ldots,f_N$. For any $i\in[N]$, we define 
\begin{align*}
    L_i(X,Y):=\max\{f_i(X,Y)-\eta,0\}, \quad U_i(X,Y):=\{f_i(X,Y)+\eta,G(X,Y)\}.
\end{align*}
Then, we can validate that $\mathcal{P}_{k,\beta}(\Theta)\subset \cup_{i=1}^{N}[L_i(X,Y),U_i(X,Y)]$ and $U_i(X,Y)-L_i(X,Y)\leq \min\{2\eta,G(X,Y)\}$. Furthermore, we also deduce that
\begin{align*}
    &\|U_i-L_i\|_1=\int(U_i(X,Y)-L_i(X,Y))\dint (X,Y)\\
    &=\int_{|Y|<2u_2}(U_i(X,Y)-L_i(X,Y))\dint (X,Y) + \int_{|Y|\geq 2u_2}(U_i(X,Y)-L_i(X,Y))\dint (X,Y)\\
    &\leq c_1\eta+\exp(-c_1^2/(2u_3))\leq c_2\eta,
\end{align*}
where $c_1=\max\{2u_2,\sqrt{8}u_3\}\log(1/\eta)$ and $c_2>0$ is some universal constant. This means that each bracket $[L_i(X,Y),U_i(X,Y)]$ is of size $c_2\eta$. Recall that the bracketing entropy is the logarithm of the smallest number of brackets to cover $\mathcal{P}_{k,\beta}(\Theta)$, it follows that
\begin{align*}
    H_B(c_2\eta,\mathcal{P}_{k,\beta}(\Theta),\|\cdot\|_1)&\leq \log N(\eta,\mathcal{P}_{k,\beta}(\Theta),\|\cdot\|_1)\leq\log N(\eta,\mathcal{P}_{k,\beta}(\Theta),\|\cdot\|_{\infty})\lesssim\log(1/\eta), 
\end{align*}
where the second inequality occurs since $\|\cdot\|_{\infty}\leq \|\cdot\|_1$, while the last inequality is due to the result in part (i). Moreover, as the Hellinger distance is upper bounded by the $L1$-norm $\|\cdot\|_1$, we get that 
\begin{align*}
    H_B(c_2\eta,\mathcal{P}_{k,\beta}(\Theta),h)\leq H_B(c_2\eta,\mathcal{P}_{k,\beta}(\Theta),\|\cdot\|_1)\lesssim\log(1/\eta).
\end{align*}
Here, if we choose $\eta=\varepsilon/c_2$, we can conclude that $H_B(\varepsilon,\mathcal{P}_{k,\beta}(\Theta),h)\leq \log(1/\varepsilon)$.

\subsection{Proof of Lemma \ref{lemma_r_tilde}}
\label{appendix_r_tilde}
We begin with recalling the system of interest here:
\begin{align}
\label{eq:original_system}
\sum_{l=1}^{m}\sum_{\alpha\in\mathcal{J}_{\ell_1,\ell_2}}\dfrac{p_l^2~q_{1l}^{\alpha_1}~q_{2l}^{\alpha_2}~q_{3l}^{\alpha_3}~q_{4l}^{\alpha_4}~q_{5l}^{\alpha_5}}{\alpha_1!~\alpha_2!~\alpha_3!~\alpha_4!~\alpha_5!}=0,
\end{align}
with unknown variables $\{(p_{l},q_{1l},q_{2l},q_{3l},q_{4l},q_{5l})\}_{l=1}^{m}\subset\mathbb{R}^5$ for all $\ell_1\geq 0$ and $\ell_2\geq 0$ that satisfy $1\leq \ell_1+\ell_2\leq r$, where
\begin{align*}
    \mathcal{J}_{\ell_1,\ell_2}:=\{\alpha=(\alpha_i)_{i=1}^5\in\mathbb{N}^5:\alpha_1+2\alpha_2+\alpha_3=\ell_1,\ \alpha_3+\alpha_4+2\alpha_5=\ell_2\}.
\end{align*}
Let us consider only a part of the above system when $\ell_1=0$ as follows:
\begin{align}
\label{eq:reduced_system}
\sum_{l=1}^{m}\sum_{\substack{\alpha_4,\alpha_5\in\mathbb{N}\\ \alpha_4+2\alpha_5=\ell_2}}\dfrac{p_{l}^2~q_{4l}^{\alpha_4}~q_{5l}^{\alpha_5}}{\alpha_4!~\alpha_5!}=0,
\end{align}
for all $1\leq \ell_2\leq r$, which takes the same form as the system in equation~\eqref{eq:system_r_bar}. Thus, it follows from Lemma~\ref{lemma_r_bar} that the smallest positive integer $r$ such that the system~\eqref{eq:reduced_system} does not admit any non-trivial solutions is $\Bar{r}(m)$. Therefore, we obtain that $\widetilde{r}(m)\leq \Bar{r}(m)$. 

Next, we will respectively show that $\widetilde{r}(2)=4$ and $\widetilde{r}(3)=6$. 

\textbf{When $m=2$:} In this case, it follows from the above result that $\widetilde{r}(m)\leq \Bar{r}(m)=4$. Thus, it is sufficient to demonstrate $\widetilde{r}(m)>3$, i.e. pointing out a non-trivial solution for the system~\eqref{eq:original_system} when $r=3$, which is given by
\begin{align}
    \label{eq:system_3}
    \sum_{l=1}^{m}p_{l}^2q_{1l}=0,\quad \sum_{l=1}^{m}p_{l}^2q_{4l}=0,\nonumber\\
    \sum_{l=1}^{m}p_{l}^2\Big(\frac{1}{2!}q_{1l}^{2}+q_{2l}\Big)=0,\quad \sum_{l=1}^{m}p_{l}^2\Big(q_{1l}q_{4l}+q_{3l}\Big)=0,\quad \sum_{l=1}^{m}p_{l}^2\Big(\frac{1}{2!}q_{4l}^{2}+q_{5l}\Big)=0,\nonumber\\
    \sum_{l=1}^{m}p_{l}^2\Big(\frac{1}{3!}q_{1l}^{3}+q_{1l}q_{2l}\Big)=0,\quad \sum_{l=1}^{m}p_{l}^2\Big(\frac{1}{2!}q_{1l}^{2}q_{4l}+q_{1l}q_{3l}+q_{2l}q_{4l}\Big)=0,\nonumber\\
    \sum_{l=1}^{m}p_{l}^2\Big(\frac{1}{2!}q_{1l}q_{4l}^{2}+q_{1l}q_{5l}+q_{3l}q_{4l}\Big)=0,\quad \sum_{l=1}^{m}p_{l}^2\Big(\frac{1}{3!}q_{4l}^{3}+q_{4l}q_{5l}\Big)=0.
\end{align}
We can check that the following is a non-trivial solution of the system~\eqref{eq:system_3}:  
\begin{align*}
    p_{l}=0, \quad q_{1l}=q_{2l}=q_{3l}=0, \quad \forall l\in[m],\\
    q_{41}=1,q_{42}=-1, \quad q_{51}=q_{52}=-\frac{1}{2}.
\end{align*}
Hence, we conclude that $\widetilde{r}(m)=4$.

\textbf{When $m=3$:} Again, according to Lemma~\ref{lemma_r_bar}, we have $\widetilde{r}(m)\leq \Bar{r}(m)=6$. Therefore, it suffices to show a non-trivial solution of the system~\eqref{eq:original_system} for $r=5$, which is a combination of the system~\eqref{eq:system_3} and the following system:
\begin{align*}
    \sum_{l=1}^{m}p_{l}^2\Big(\frac{1}{4!}q_{1l}^{4}+\frac{1}{2!}q_{1l}^{2}q_{2l}+\frac{1}{2!}q_{2l}^{2}\Big)=0, \quad \sum_{l=1}^{m}p_{l}^2\Big(\frac{1}{4!}q_{4l}^{2}+\frac{1}{2!}q_{4l}^{2}q_{5l}+\frac{1}{2!}q_{5l}^{2}\Big)=0,\\
    \sum_{l=1}^{m}p_{l}^2\Big(\frac{1}{3!}q_{1l}^{3}q_{4l}+\frac{1}{2!}q_{1l}q_{2l}^{2}+\frac{1}{2!}q_{1l}^{2}q_{2l}+q_{2l}q_{3l}\Big)=0,\\
    \sum_{l=1}^{m}p_{l}^2\Big(\frac{1}{3!}q_{1l}q_{4l}^{3}+\frac{1}{2!}q_{1l}q_{4l}q_{5l}^{2}+\frac{1}{2!}q_{3l}q_{4l}^{2}+q_{3l}q_{5l}\Big)=0,\\
    \sum_{l=1}^{m}p_{l}^2\Big(\frac{1}{2!2!}q_{1l}^{2}q_{4l}^{2}+\frac{1}{2!}q_{2l}q_{4l}^{2}+\frac{1}{2!}q_{1l}^{2}q_{5l}+q_{2l}q_{5l}+q_{1l}q_{3l}q_{4l}+\frac{1}{2!}q_{3l}^{2}\Big)=0,\\
    \sum_{l=1}^{m}p_{l}^2\Big(\frac{1}{5!}q_{1l}^{5}+\frac{1}{3!}q_{1l}^{3}q_{2l}+\frac{1}{2!}q_{1l}q_{2l}^{2}\Big)=0,\quad \sum_{l=1}^{m}p_{l}^2\Big(\frac{1}{5!}q_{4l}^{5}+\frac{1}{3!}q_{4l}^{3}q_{5l}+\frac{1}{2!}q_{4l}q_{5l}^{2}\Big)=0,\\
    \sum_{l=1}^{m}p_{l}^2\Big(\frac{1}{4!}q_{1l}^{4}q_{4l}+\frac{1}{2!}q_{1l}^{2}q_{2l}+\frac{1}{2!}q_{2l}^{2}q_{4l}+\frac{1}{3!}q_{1l}^{3}q_{3l}+q_{1l}q_{2l}q_{3l}\Big)=0,\\
    \sum_{l=1}^{m}p_{l}^2\Big(\frac{1}{4!}q_{1l}q_{4l}^{4}+\frac{1}{2!}q_{1l}q_{4l}^{2}q_{5l}+\frac{1}{2!}q_{1l}q_{5l}^{2}+\frac{1}{3!}q_{3l}q_{4l}^{3}+q_{3l}q_{4l}q_{5l}\Big)=0,\\
    \sum_{l=1}^{m}p_{l}^2\Big(\frac{1}{3!2!}q_{1l}^{3}q_{4l}^{2}+\frac{1}{3!}q_{1l}^{3}q_{5l}+\frac{1}{2!}q_{1l}q_{2l}q_{4l}^{2}+q_{1l}q_{2l}q_{4l}+\frac{1}{2!}q_{1l}^{2}q_{3l}q_{4l}+q_{2l}q_{3l}q_{4l}+\frac{1}{2!}q_{1l}q_{3l}^{2}\Big)=0,\\
    \sum_{l=1}^{m}p_{l}^2\Big(\frac{1}{2!3!}q_{1l}^{2}q_{4l}^{3}+q_{1l}^{2}q_{4l}q_{5l}+\frac{1}{3!}q_{2l}q_{3l}^{3}+q_{2l}q_{4l}q_{5l}+\frac{1}{2!}q_{1l}q_{3l}q_{4l}^{2}+q_{1l}q_{3l}q_{5l}+\frac{1}{2!}q_{3l}^{2}q_{4l}\Big)=0.
\end{align*}
It can be verified that the following is a non-trivial of this system:
\begin{align*}
    p_{l}=0, \quad q_{1l}=q_{2l}=q_{3l}=0, \quad \forall l\in[m],\\
    q_{41}=\frac{\sqrt{3}}{3},q_{42}=-\frac{\sqrt{3}}{3},q_{43}=0, \quad q_{51}=q_{52}=-\frac{1}{6},q_{53}=0.
\end{align*}
As a consequence, we obtain that $\widetilde{r}(m)>5$, which implies the desired conclusion that $\widetilde{r}(m)=6$.

\end{document}